\documentclass{article}

\usepackage[preprint]{neurips_2026}
\usepackage{alphalph}

\usepackage[utf8]{inputenc}
\usepackage[T1]{fontenc}
\usepackage{hyperref}
\usepackage{url}
\usepackage{booktabs}
\usepackage{longtable}
\usepackage{amsfonts}
\usepackage{amsmath}
\usepackage{amssymb}
\usepackage{nicefrac}
\usepackage{microtype}
\usepackage{xcolor}
\usepackage{graphicx}
\usepackage{multirow}
\usepackage{array}
\usepackage{tabularx}
\usepackage{placeins}
\usepackage{float}
\usepackage{subcaption}

\newcommand{\poolbench}{\textsc{PoolBench}}

\title{\poolbench{}: A Benchmark for Pooling Strategies in\
Concept Representation Evaluation for Decoder-Only LLMs}

\author{Ayushi Agarwal\\
\small Independent Researcher}

\begin{document}

\maketitle

\begin{abstract}
Pooling is a consequential but under-examined design choice in decoder-only concept
representation work: to build a passage-level vector, practitioners must collapse
token-level hidden states into a single representation, yet no shared protocol
exists for comparing that choice across concepts, models, and downstream tasks.
The result is a scattered literature where reported gains are confounded by
simultaneous changes in dataset, layer, construction method, and pooling rule,
making it impossible for practitioners to make principled decisions.
We introduce \poolbench{}, a benchmark that isolates pooling as the experimental
variable under a fixed evaluation protocol.
\poolbench{} covers 17 concepts, 19 pooling strategies, and 3 open-weight
decoder-only models (Llama-3.1-8B, Gemma-2-9B, Mistral-7B), all evaluated on a
single audited corpus of 37{,}693 real-text passages.
The primary evaluation axis is linear separability (D1/AUROC); steered concept
prevalence (D2/SCP) and output-level disentanglement (D3) serve as additional
diagnostic axes.
The primary leaderboard finding is decisive: the best strategy,
\texttt{W4\_hierarchical}, reaches a cross-model mean AUROC of 0.7799, while the
widely adopted \texttt{P1\_last\_token} baseline reaches only 0.7640 and is
statistically significantly worse under Friedman+Nemenyi testing
($p = 2.0 \times 10^{-36}$; 77 significant pairs among 18 effective strategies).
Rankings are stable across candidate layers ($\rho=0.961$--$0.990$).
A key negative result is that strong detection does not imply strong steering:
D2 and D3 are substantially weaker than D1 for most concepts, and the gap persists
across strategies, indicating a fundamental representational limit rather than a pooling failure.
On mid-difficulty concepts, the leading strategy (\texttt{W4\_hierarchical}) outperforms the
commonly adopted default (\texttt{P1\_last\_token}) by 0.042--0.113 AUROC;
construction method choice (DiffMean vs.~REPE) has a larger effect ($\Delta$AUROC 0.15) than
pooling choice ($\Delta$AUROC 0.016), establishing the correct practical hierarchy.
We release the corpus, pre-extracted activations, scorer models, steering vectors,
and evaluation code to provide the community with a structured, methodologically
consistent evaluation reference and reusable protocol for pooling research.
\end{abstract}

\section{Introduction}
Activation-based concept work in decoder-only language models requires collapsing
token-level hidden states into a single passage representation before fitting a
linear probe or constructing a steering direction.
That collapsing step---pooling---is rarely treated as an explicit experimental
variable. Published activation-steering
work~\citep{turner2023activation,rimsky2023steering,zou2023representation} typically
fixes a single pooling rule (most often the final token or mean of all tokens)
without evaluating alternatives, making it impossible to attribute reported gains to
the pooling choice rather than to simultaneous differences in dataset, construction
method, or layer selection.

The absence of a shared protocol has practical consequences.
When building a concept direction for \emph{coding}, \emph{hedging}, or
\emph{mathematical precision} in a decoder-only model, practitioners have no empirical basis for
choosing among pooling rules.
We introduce \poolbench{}, a benchmark that isolates pooling as the sole
experimental variable under a fixed protocol: construction (DiffMean), corpus, and
per-model layer selection are held constant so that AUROC differences across the
19~evaluated strategies can be attributed to the pooling function alone.

The benchmark spans 17 concepts and 3 decoder-only models (Llama-3.1-8B,
Gemma-2-9B, Mistral-7B) on an audited corpus of 37{,}693 real-text passages,
evaluated along three axes: linear separability of positive and negative passages
(D1/AUROC), steered concept prevalence in generated text (D2/SCP), and output-level
concept disentanglement (D3).
The D1 leaderboard yields a statistically significant strategy ordering
($p = 2.0 \times 10^{-36}$; 77 Nemenyi-significant pairs among 18 effective strategies) that is stable across
candidate layers ($\rho = 0.961$--$0.990$).
\texttt{W4\_hierarchical} achieves the highest cross-model mean AUROC (0.7799);
the widely adopted \texttt{P1\_last\_token} default reaches only 0.7640.
A secondary finding is that strong linear detectability does not transfer to strong
steered concept prevalence: strategies that rank highest on D1 cluster near zero on
D2~SCP, while uniform-coverage strategies show more consistent behaviour across
both axes.
Full corpus specifications, strategy definitions, released artifacts, and
reproducibility details are provided in Appendix~\ref{app:artifacts}.

\section{Related Work}
\label{sec:related}
Activation steering constructs concept directions by pooling hidden states from
contrasting passage sets, then injecting the resulting direction into the residual
stream at inference time.
Turner et al.~\citeyear{turner2023activation} and Rimsky et al.~\citeyear{rimsky2023steering}
established DiffMean (Contrastive Activation Addition) as the dominant construction
method; Zou et al.~\citeyear{zou2023representation} proposed the related RepE
framework with per-pair normalisation.
Li et al.~\citeyear{li2023iti} introduced Inference-Time Intervention (ITI), which
selects the top attention heads by probe accuracy before pooling; this is the
nearest prior evidence that the choice of which tokens to aggregate materially
alters steering vector quality.
Burns et al.~\citeyear{burns2023discovering} and Marks and
Tegmark~\citeyear{marks2023geometry} showed that a wide range of propositional
concepts form linearly separable directions in the residual stream.
Zou et al.~\citeyear{zou2025axbench} subsequently reported that steering vectors
frequently underperform prompting on instruction-following tasks; we do not revisit
that comparison, as \poolbench{} addresses the prior question of which pooling rule
yields the best concept direction given the decision to use activation steering.
Engels et al.~\citeyear{engels2024not} identify conditions under which
representational structure resists linear decomposition, motivating our
pre-registered linearity check before admitting concepts to the Nemenyi ranking
analysis.

The question of how to aggregate token-level representations into a passage vector
predates decoder-only models.
Devlin et al.~\citeyear{devlin2019bert} established the \texttt{[CLS]} token as the
default for BERT-style architectures, where masked-LM pre-training explicitly
trains that position for sequence-level classification.
For dense retrieval and similarity tasks, Arora et al.~\citeyear{arora2017simple}
proposed Smooth Inverse Frequency (SIF) weighting and Izacard
et al.~\citeyear{izacard2022contriever} studied length-normalised mean pooling.
Neither was evaluated in the context of concept-direction construction for
decoder-only models, where the causal-LM objective provides no analogous pooling
inductive bias.
Sparse autoencoders~\citep{gao2023scaling} provide a complementary lens on
residual-stream directions; we include SAE-based diagnostics using GemmaScope,
LlamaScope, and the community \texttt{mistral-7b-res-wg} SAELens release as
supplementary analyses.
The benchmarking conventions used in \poolbench{}---controlled variables,
pre-committed minimum detectable effect, and Friedman+Nemenyi ranking
analysis---follow the statistical framework of
Demsar~\citeyear{demsar2006statistical} and the reproducibility conventions of
HELM~\citep{liang2022holistic}.

\section{Benchmark Design}
\label{sec:design}
\poolbench{} takes as given that the practitioner has decided to use
activation-based concept representations in a decoder-only transformer, and
examines which pooling rule produces the best result under a fixed protocol.
The benchmark does not evaluate whether activation steering outperforms
prompting~\citep{zou2025axbench}, nor does it cover encoder architectures for which
the \texttt{[CLS]} token convention is established~\citep{devlin2019bert}.
The central design principle is isolation of the pooling variable: construction
method (DiffMean), corpus, and layer selection are held constant, so that observed
differences in AUROC can be attributed to the pooling function alone.

\paragraph{Corpus and concept coverage.}
\label{sec:corpus}
The benchmark corpus contains 37{,}693 passages drawn from 20 open public datasets
(see Appendix~\ref{app:sources} for the complete source map with HuggingFace links)
and organised into 17 concepts spanning lexical, syntactic, register, and
semantic-abstract discourse phenomena.
The concept set ranges from register tasks where signal is broadly distributed
(\texttt{academic\_tone}, \texttt{bureaucratic}, \texttt{code\_docs}) to
syntactically grounded tasks where even an attribution-based reference stays near
chance (\texttt{causation}, \texttt{contrast}, \texttt{hedging},
\texttt{negation\_density}).
Family labels are descriptive appendix metadata only
(Appendix~\ref{app:families}); they carry no routing implications.

\paragraph{Passage length and class balance.}
Every passage is windowed to 300--500 tokens as measured by the Llama-3.1-8B
tokeniser.
The 300-token minimum ensures that linguistically filtered strategies (L-family,
which rely on part-of-speech or dependency parsers) have enough informative tokens
to average over; the 500-token ceiling holds the positional regime constant so that
last-token and window strategies behave consistently across the corpus.
Two documented exceptions apply: \texttt{toxicity} uses a 15--30-token floor,
reflecting the short-form nature of online toxic language, and \texttt{deference}
uses an 8--128 token range to accommodate the Intel Polite-Guard
source~\citep{intel2023politeguard}.
Class balance is enforced at exactly 700 positive / 700 negative training examples
and 300 positive / 300 negative test examples per concept, giving equal empirical
risk during probe training.
With $n = 700$ training examples and 5-fold cross-validation, each test fold
contains $\approx 140$ examples per class.
The pre-committed 95\% CI half-width is $\pm 0.0276$ per fold, setting the minimum
detectable effect (MDE) at 0.05 AUROC; the cross-concept mean CI half-width is
$\pm 0.0097$.
This pre-commitment is stored in \texttt{results/power\_analysis.json} before the
full sweep began (Appendix~\ref{app:power}).

\paragraph{Construction modes and domain stratification.}
For the seven sparse-lexical and syntactic concepts where signal is carried by
specific surface markers
(\texttt{hedging}, \texttt{causation}, \texttt{contrast}, \texttt{conditionality},
\texttt{legal\_formality}, \texttt{frustration}, \texttt{negation\_density}),
negatives are constructed by rule-based rewriting of the same passage with the
markers removed or replaced.
This eliminates topic and vocabulary confounds: each matched positive--negative pair
is drawn from the same source document, differing only in the target linguistic
phenomenon.
Paired passages must differ by no more than $\pm 25$ tokens after tokenisation to
prevent pooling strategies from exploiting length as a proxy signal.
For the remaining ten concepts (dense-lexical, register, semantic-abstract),
negatives are independently sampled from domain-disjoint sources.
Every concept must span at least three source domains, preventing strategies from
learning domain lexis as a proxy for the concept.

An April~2026 integrity audit applied MD5-based deduplication across all passage
texts; full deduplication counts, contamination ablation results, and per-concept
released counts are in Appendix~\ref{app:corpusqa}.
All 17 concepts pass the pre-committed power criterion
(Appendix~\ref{app:power}).

\paragraph{Strategy space.}
The benchmark evaluates 19 pooling strategies falling into five groups:
position-based, uniform averaging, window-based, saliency-guided, and
linguistically filtered.
This is intentionally broader than a simple last-token-versus-mean comparison:
it includes strong defaults that practitioners already use, lightweight ablations,
and more structured strategies that attempt to isolate concept-bearing positions.

The evaluated strategies are:
\texttt{P1\_last\_token}, \texttt{P2\_first\_token}, \texttt{P3\_CLS},
\texttt{A1\_mean}, \texttt{A2\_max}, \texttt{A3\_random}, \texttt{A4\_norm},
\texttt{W1\_mean\_last\_4}, \texttt{W2\_mean\_last\_8},
\texttt{W3\_mean\_last\_16}, \texttt{W4\_hierarchical},
\texttt{S1\_attention\_weighted}, \texttt{S2\_SIF}, \texttt{S3\_ITI\_exact},
\texttt{L1\_POS\_filtered}, \texttt{L2\_dependency\_rel},
\texttt{L3\_named\_entity}, \texttt{L4\_subword\_root}, and \texttt{L5\_SVO}.
Of the 19, 18 are effective pooling strategies.
\texttt{L2\_dependency\_rel} is included for completeness but collapses to
mean pooling on 100\% of passages across all three models due to a parser
implementation issue (zero dependency arcs matched); it is therefore tied
exactly to \texttt{A1\_mean} and is not a distinct strategy in this release
(see Section~\ref{sec:d1} and Appendix~\ref{app:fallback}).
An Input$\times$Gradient attribution method is reported separately as a
non-ranked upper-bound reference and is not part of the leaderboard.

\paragraph{D1: concept detection AUROC.}
The primary leaderboard metric is AUROC~\citep{hanley1982roc} from a 5-fold
out-of-fold logistic probe trained on the 700-per-class training split.
Formally, for a pooled representation $\phi_s(x) \in \mathbb{R}^d$ produced by
strategy $s$ and a trained linear probe $f_\theta:\mathbb{R}^d \to [0,1]$,
\begin{equation}
  \text{D1}(c,s) = P\!\left(f_\theta(\phi_s(x^+)) > f_\theta(\phi_s(x^-))\right)
  \label{eq:d1}
\end{equation}
where $x^+$, $x^-$ are randomly drawn positive and negative passages for concept $c$.
AUROC is threshold-independent and equals the probability that the probe assigns a
higher score to a positive than to a negative example, making it robust to the
balanced-class setting \citep{hanley1982roc,kohavi1995cv}.
The pre-committed MDE is 0.05 AUROC; the 95\% CI half-width per fold is
$\pm 0.0276$.

\paragraph{D2: steered concept prevalence (SCP).}
D2 measures whether steering the model along the concept direction causes it to
generate more of the target concept in free-form text~\citep{turner2023activation,
rimsky2023steering}.
For steering coefficient $\alpha$ the metric is
\begin{equation}
  \mathrm{SCP}_c(\alpha) = \frac{N_c^\mathrm{steered}(\alpha) -
  N_c^\mathrm{baseline}}{T}
  \label{eq:scp}
\end{equation}
where $N_c^\mathrm{steered}$ is the count of target-concept tokens in the steered
output scored by a fine-tuned BERT classifier, and $T$ is total output tokens.
An ideal direction produces monotonically increasing SCP with $\alpha$.
In practice we observe three regimes: monotonic growth (easy register concepts),
early saturation, and non-monotonic or fragile scaling (hard syntactic concepts).
D2 is a downstream behavioral diagnostic rather than the primary leaderboard
metric: it reveals representational limits (why D1 separability does not transfer
to steering) rather than discriminating among strategies for steering use-cases,
where inter-strategy SCP differences of 0.005--0.015 are at the boundary of
distinguishability from generation noise.
D2 conceptually extends the REPE/CAA activation-addition
framework~\citep{zou2023representation,rimsky2023steering} to a standardised
cross-strategy measurement.

\paragraph{D3: output-level disentanglement.}
D3 measures whether steering target concept $A$ bleeds into a semantically adjacent
neighbour concept $B$ at the output level:
\begin{equation}
  \mathrm{D3}_{c} = \frac{\Delta_B}{\Delta_A},\quad
  \Delta_A = \mathrm{SCP}_A(\alpha^*),\quad \Delta_B = \mathrm{SCP}_B(\alpha^*)
  \label{eq:d3}
\end{equation}
A ratio below 1 is desirable.
Because the ratio depends on the D2 denominator, D3 is only interpreted when
$\lvert\Delta_A\rvert > 0.01$; under that gate it does not alter the main findings.

The three axes mirror the standard practitioner workflow: verify linear detectability (D1), confirm steered generation (D2)~\citep{zou2023representation,burns2023discovering}, and check specificity (D3). Published steering work typically reports only the D2 analogue~\citep{turner2023activation,rimsky2023steering}, making it impossible to distinguish a pooling failure from a representational limit. Separating D1 from D2 is therefore the key diagnostic contribution of \poolbench{}: D2 and D3 are explicitly framed as diagnostic axes that reveal the boundaries of concept representation, not as axes that rank pooling strategies for steering tasks. D1 is the primary leaderboard metric; D2 SCP values are small in absolute magnitude (most cross-model means within $\pm 0.05$ of zero), and D3 is interpreted only when $|\Delta_A| > 0.01$.

\section{Experimental Setup}
\label{sec:setup}

\paragraph{Models and activation extraction.}
The leaderboard covers three open-weight decoder-only models:
Llama-3.1-8B~\citep{dubey2024llama3}, Gemma-2-9B~\citep{mesnard2024gemma},
and Mistral-7B~\citep{jiang2023mistral}.
For each model, hidden-state activations are extracted at three candidate layers
simultaneously via forward-pass hook registration---a single forward pass captures
all three layers---yielding per-token residual-stream vectors at the selected
depths.
Candidate layers are L\{16, 24, 31\} for Llama-3.1-8B, L\{14, 28, 41\} for
Gemma-2-9B, and L\{8, 16, 24\} for Mistral-7B.

\paragraph{Best-layer selection and stability.}
The released best layer for each model (L31, L28, L16 respectively) is chosen by
maximising the cross-concept mean AUROC of \texttt{A1\_mean} (uniform mean
pooling) on the training fold.
Using the simplest possible pooling baseline as the selection criterion avoids
circularity: the selected layer is not the one that makes the winning strategy
look best, but the one where the model's average token representation is most
discriminative under any linear probe.
Layer-rank stability is verified post-hoc: Spearman's $\rho$ across all
$17 \times 19$ concept--strategy cells ranges from 0.961 to 0.990 across every
within-model candidate-layer pair (Section~\ref{sec:stability}), confirming that
the leaderboard conclusions are invariant to this choice.

\paragraph{Linearity check.}
Before the main sweep we verify that D1 AUROC can be attributed to the linear
probe for each concept.
Using mean-pooled activations at the best layer, we compute the gap
$\Delta_\text{lin}(c) = \text{AUROC}_\text{MLP}(c) - \text{AUROC}_\text{linear}(c)$
with a two-layer MLP probe on the same 5-fold split.
Concepts with $\Delta_\text{lin}(c) \geq 0.03$ (pre-registered as 60\% of the
MDE) are excluded from the Nemenyi statistical analysis but still reported in D1
tables.
Six concepts are flagged as non-linear on Llama and Gemma
(\texttt{legal\_formality}, \texttt{frustration}, \texttt{imdb\_sentiment},
\texttt{toxicity}, \texttt{deference}, \texttt{negation\_density}); five on Mistral
(\texttt{imdb\_sentiment} passes at gap $= +0.027$).
Notably, all five hard syntactic concepts
(\texttt{causation}, \texttt{contrast}, \texttt{legal\_formality}, \texttt{hedging},
\texttt{negation\_density}) show a \emph{negative} MLP--linear gap ($-0.008$ to
$-0.18$): both probes are near chance and the MLP overfits, confirming that these
concepts are representation-limited rather than non-linear in any exploitable sense.
The linearity gap by concept is visualised in Appendix~\ref{app:concepts}.

\paragraph{Reliability.}
Benchmark reliability is measured via ICC(2,1)~\citep{shrout1979icc}, which treats
the three candidate layers as raters and concept--strategy cells as subjects,
asking whether AUROC rankings produced at the selected layer agree with rankings at
the other candidate layers.
Cross-model mean ICC is 0.8773 (Llama), 0.8586 (Gemma), and 0.9217 (Mistral),
in the good-to-excellent range under standard interpretive thresholds~\citep{koo2016icc};
however, this masks concept-specific fragility: \texttt{contrast} and
\texttt{causation} fall below the 0.75 acceptable threshold on Llama (ICC~0.34)
and Gemma (ICC~0.40--0.57), consistent with both concepts being near-chance and
therefore offering no stable ranking signal; \texttt{toxicity} is the weakest on
Mistral.
Per-concept ICC analysis is in Appendix~\ref{app:icc}.

\paragraph{Construction method.}
The primary leaderboard fixes concept direction construction to C1 DiffMean:
$\vec{v}_c = \bar{\phi}(X^+) - \bar{\phi}(X^-)$,
the mean-activated difference between positive and negative training passages.
This is held fixed so that pooling remains the sole experimental variable.
We additionally evaluate C2 PCA (top principal component of the activation
difference distribution), C3 logistic-regression weights, and
C4 REPE~\citep{zou2023representation} as a robustness check
(Section~\ref{sec:interaction}; extended per-concept and optimal-layer results
are in Appendices~\ref{app:construction}--\ref{app:construction_ext}).

\paragraph{SAE interpretability.}
As a supplementary diagnostic we encode the DiffMean concept direction through a
Sparse Autoencoder (SAE)~\citep{gao2023scaling} to identify which model features
activate most strongly in the concept direction, per concept and per strategy.
For Mistral-7B we use the publicly available \texttt{mistral-7b-res-wg} SAELens
release (\texttt{hook\_resid\_pre}), which differs from the
\texttt{hook\_resid\_post} hook-point used for Llama and Gemma; the residual stream
is approximately preserved across one block, so feature attributions are
qualitatively comparable.
Full results and sparsity analyses are in Appendix~\ref{app:sae}.

\paragraph{Statistics.}
Strategy ranking uses Friedman's test followed by Nemenyi post-hoc pairwise
testing~\citep{demsar2006statistical,friedman1940rankings}.
The Friedman test operates on the 11 linearity-passing concepts
(\mbox{Effective $N = 12.106$}) and yields $p = 2.0 \times 10^{-36}$;
the critical distance is 8.069 rank positions; 77 strategy pairs are significantly
different (using $k = 19$ strategies for comparability; the degenerate
\texttt{L2\_dependency\_rel} participates as an effective duplicate of
\texttt{A1\_mean} and does not alter any significant-pair conclusion).
The Input$\times$Gradient attribution reference~\citep{simonyan2014saliency} is
reported as a non-ranked ceiling: it uses token-level saliency unavailable to any
of the 19 pooling strategies, providing a diagnostic bound on representational
signal available at the selected layer (Appendix~\ref{app:oracle}).

\paragraph{Compute.}
Full hardware specifications and GPU-hour breakdowns are in
Appendix~\ref{app:compute}.

\section{Results}
\subsection{Main D1 leaderboard}
\label{sec:d1}
\begin{figure}[t]
  \centering
  \begin{subfigure}[b]{0.47\linewidth}
    \centering
    \includegraphics[width=\linewidth]{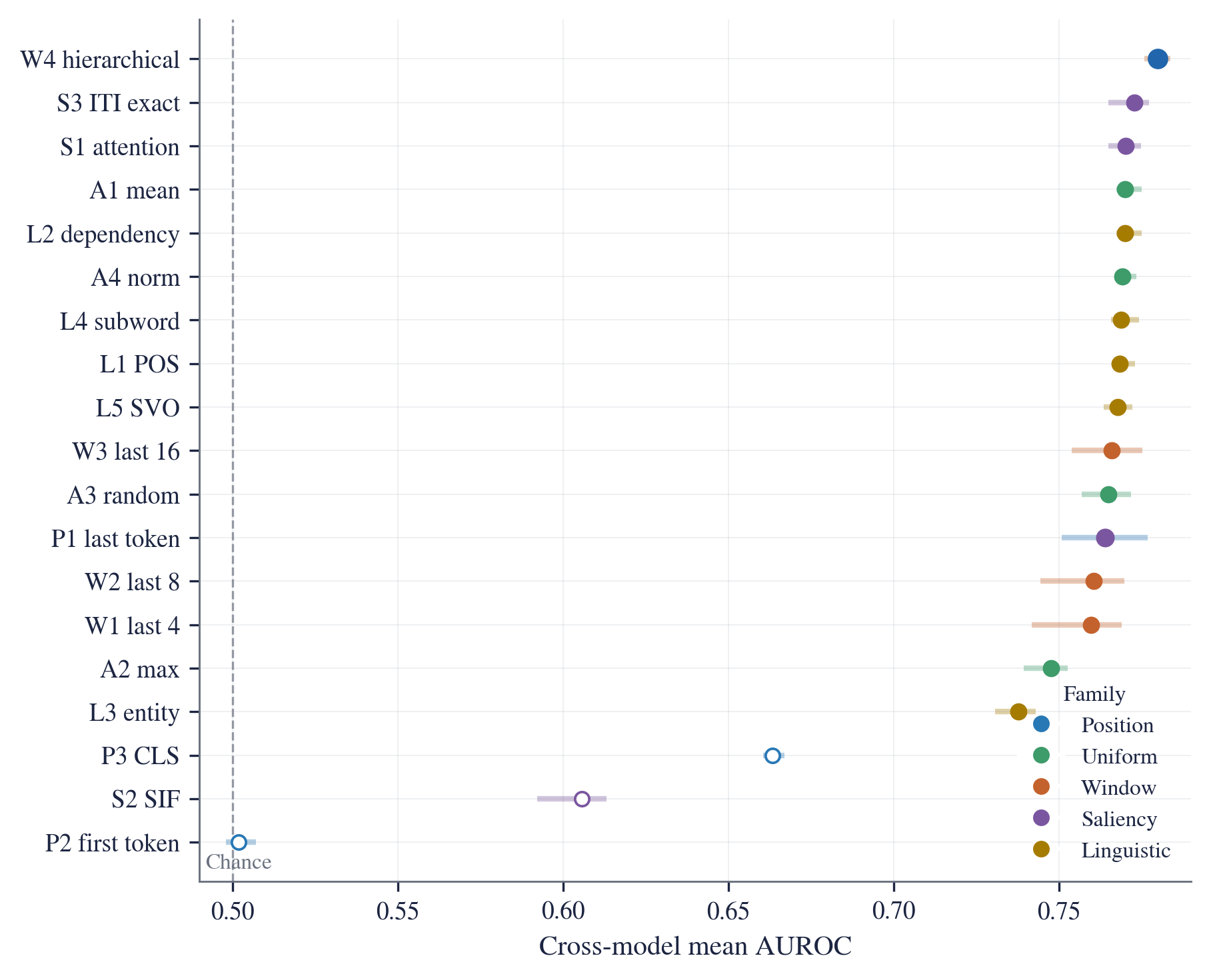}
    \caption{Primary D1 leaderboard: cross-model mean AUROC with min--max range.
    \texttt{W4\_hierarchical} combines the best mean with the tightest range;
    \texttt{P1\_last\_token} appears competitive by mean alone but is materially
    less stable.}
    \label{fig:leaderboard}
  \end{subfigure}
  \hfill
  \begin{subfigure}[b]{0.47\linewidth}
    \centering
    \includegraphics[width=\linewidth]{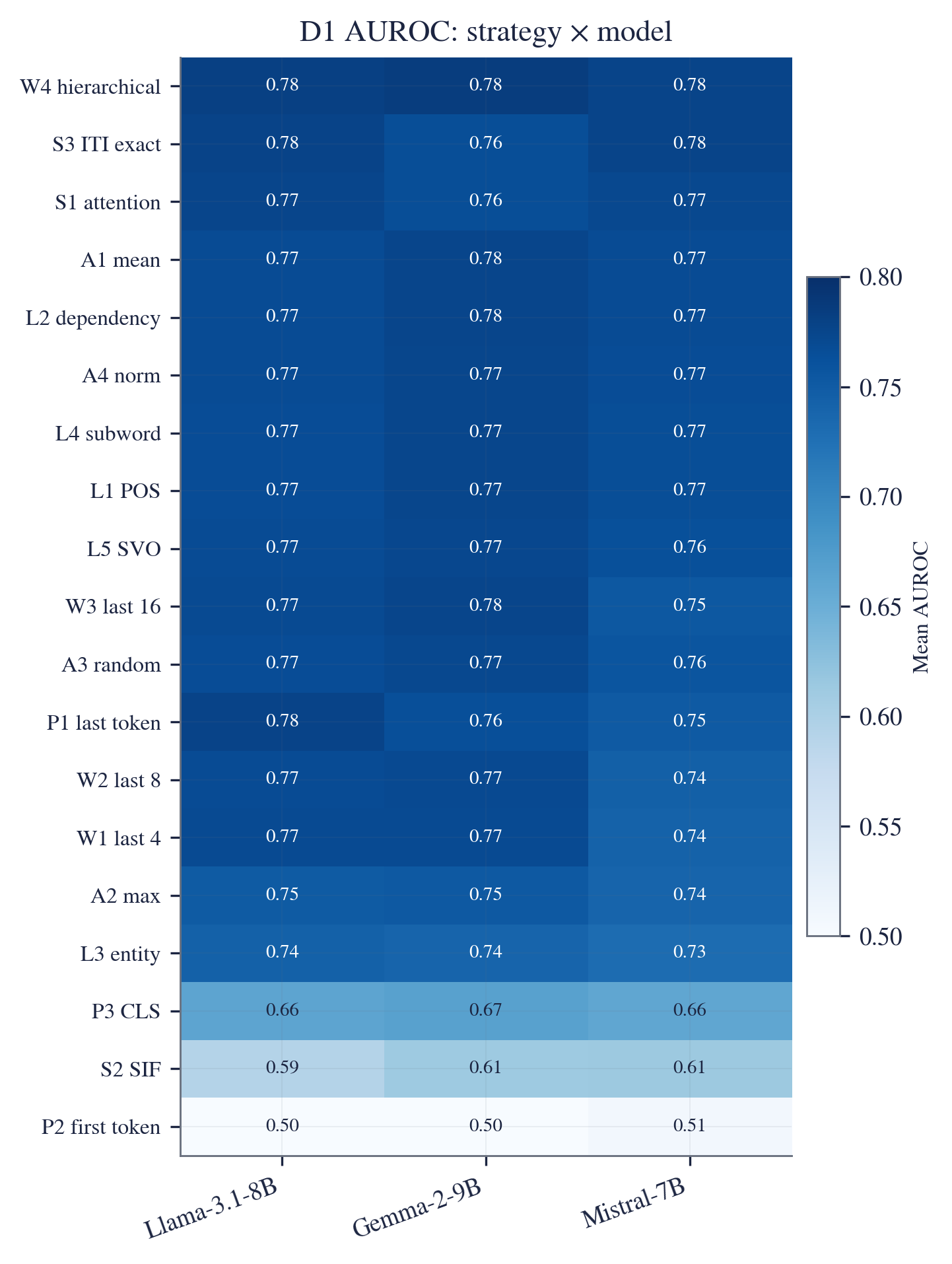}
    \caption{D1 AUROC for every strategy--model pair.
    The top tier (\texttt{W4}, \texttt{S3}, \texttt{S1}, \texttt{A1}) is consistent
    across all three models.
    \texttt{P1\_last\_token} drops sharply on Mistral.
    The bottom tier (\texttt{P3\_CLS}, \texttt{S2\_SIF}, \texttt{P2\_first\_token})
    is uniformly weak.}
    \label{fig:matrix}
  \end{subfigure}
  \caption{Left: D1 leaderboard ranked by cross-model mean AUROC.
  Right: per-strategy per-model AUROC matrix exposing stability differences.}
  \label{fig:leaderboard_matrix}
\end{figure}

Figure~\ref{fig:leaderboard} and Table~\ref{tab:d1} present the primary D1 results.
Pooling strategy choice materially affects concept representation quality at a scale that is both statistically significant and practically meaningful.

\begin{table}[t]
\centering
\caption{Primary D1 leaderboard at the released best layer per model.
Cross-model mean is averaged over the three model columns.
$\dagger$\,\texttt{L2\_dependency\_rel} collapses to mean pooling on all passages
(parser implementation issue); it is not a distinct strategy in this release and
is excluded from effective strategy counts.
All other strategies are independent pooling implementations.}
\label{tab:d1}
\small
\resizebox{\linewidth}{!}{%
\begin{tabular}{lcccc}
\toprule
Strategy & Llama-3.1-8B & Gemma-2-9B & Mistral-7B & Cross-model mean \\
\midrule
\texttt{W4\_hierarchical} & 0.7802 & 0.7836 & 0.7758 & \textbf{0.7799} \\
\texttt{S3\_ITI\_exact} & 0.7772 & 0.7650 & 0.7760 & 0.7727 \\
\texttt{S1\_attention\_weighted} & 0.7747 & 0.7650 & 0.7710 & 0.7702 \\
\texttt{A1\_mean} & 0.7673 & 0.7751 & 0.7676 & 0.7700 \\
\texttt{L2\_dependency\_rel}$^\dagger$ & 0.7673 & 0.7751 & 0.7676 & 0.7700 \\
\texttt{A4\_norm} & 0.7675 & 0.7733 & 0.7669 & 0.7692 \\
\texttt{L4\_subword\_root} & 0.7665 & 0.7742 & 0.7657 & 0.7688 \\
\texttt{L1\_POS\_filtered} & 0.7661 & 0.7731 & 0.7660 & 0.7684 \\
\texttt{L5\_SVO} & 0.7674 & 0.7723 & 0.7636 & 0.7677 \\
\texttt{W3\_mean\_last\_16} & 0.7688 & 0.7752 & 0.7538 & 0.7659 \\
\texttt{A3\_random} & 0.7661 & 0.7718 & 0.7568 & 0.7649 \\
\texttt{P1\_last\_token} & 0.7768 & 0.7642 & 0.7509 & 0.7640 \\
\texttt{W2\_mean\_last\_8} & 0.7675 & 0.7697 & 0.7443 & 0.7605 \\
\texttt{W1\_mean\_last\_4} & 0.7685 & 0.7690 & 0.7417 & 0.7597 \\
\texttt{A2\_max} & 0.7506 & 0.7527 & 0.7393 & 0.7475 \\
\texttt{L3\_named\_entity} & 0.7430 & 0.7395 & 0.7307 & 0.7377 \\
\texttt{P3\_CLS} & 0.6624 & 0.6669 & 0.6605 & 0.6633 \\
\texttt{S2\_SIF} & 0.5921 & 0.6118 & 0.6132 & 0.6057 \\
\texttt{P2\_first\_token} & 0.5003 & 0.4981 & 0.5071 & 0.5018 \\
\bottomrule
\end{tabular}%
}
\end{table}

The key finding extends beyond the identity of the top-ranked strategy: some strategies are both strong and stable while others perform deceptively well in aggregate but are inconsistent across model--concept cells.
\texttt{W4\_hierarchical} achieves the strongest overall score at 0.7799 and varies
by only 0.0078 AUROC across models.
The same stability characterizes \texttt{A1\_mean}, \texttt{A4\_norm}, and
\texttt{L1\_POS\_filtered}, each varying by fewer than 0.009 AUROC across the three
models.
By contrast, \texttt{W3\_mean\_last\_16} spans 0.0214 AUROC across models,
\texttt{P1\_last\_token} spans 0.0260, and the shorter trailing windows span
about 0.025--0.027.

The rank-based analysis resolves this distinction.
\texttt{W3\_mean\_last\_16} and \texttt{P1\_last\_token} look close to the top tier
by cross-model mean, but their average Friedman ranks are 12.30 and 13.12
respectively---meaning they lose too many individual concept-model cells to count
as robust defaults.
The Nemenyi test elevates \texttt{W4\_hierarchical} because it combines high mean
performance with consistent cell-wise ranking across all three models
(Figure~\ref{fig:nemenyi_cd}; Appendix~\ref{app:nemenyi_stats}).
It should be noted that many of the 77 significant pairs involve the clearly
weak bottom tier (\texttt{S2\_SIF}, \texttt{P3\_CLS}, \texttt{P2\_first\_token})
versus the competitive strategies; the number of significant pairs
\emph{within} the top-10 strategies is smaller and is detailed in
Appendix~\ref{app:nemenyi_stats}.

\paragraph{Tractable vs.~hard concepts.}
The cross-concept mean AUROC figures in Table~\ref{tab:d1} are numerically dominated by the six near-ceiling concepts where all strategies exceed 0.90 AUROC
(\texttt{academic\_tone}, \texttt{bureaucratic}, \texttt{code\_docs},
\texttt{depression}, \texttt{narrative}, \texttt{planning}).
The practical pooling question lives in the six concepts with cross-model mean
AUROC between 0.70 and 0.90 (\texttt{toxicity}, \texttt{deference},
\texttt{frustration}, \texttt{imdb\_sentiment}, \texttt{numerical\_precision},
\texttt{conditionality}), where genuine headroom exists.
In this tractable-but-not-saturated regime, \texttt{W4\_hierarchical} outperforms
\texttt{P1\_last\_token} by 0.042--0.113 AUROC per concept per model, confirming
that the leaderboard gap reflects genuine pooling advantage rather than noise
accumulation in the easy regime.
At the other extreme, the five hardest concepts (\texttt{causation},
\texttt{contrast}, \texttt{hedging}, \texttt{negation\_density},
\texttt{legal\_formality}) sit near chance across all strategies \emph{and} under
the IxG attribution reference: these are representation-limited at the selected
layer, and no pooling innovation will materially improve D1.
The apparent ``best strategy'' on those cells is determined by sampling noise
rather than superiority; the average-rank view in Appendix~\ref{app:rankings} is
more informative than the cell-wise maximum for near-chance cells.

\subsection{Cross-model robustness and construction sensitivity}
\label{sec:interaction}
Figure~\ref{fig:matrix} makes the cross-model pattern explicit.
Rows are strategies sorted by cross-model mean AUROC; columns are the three models.

The matrix shows two additional regularities. First, the overall leader is consistent but not identical across models:
\texttt{W4\_hierarchical} is best on Llama and Gemma, while Mistral very slightly
prefers \texttt{S3\_ITI\_exact} (0.7760 vs.\ 0.7758).
That difference is small rather than a substantive rank inversion.
Second, family membership alone is not a reliable proxy for strategy quality.
The window family has the highest family-level mean, but the uniform and linguistic
families sit just below it, while the position family collapses because
\texttt{P2\_first\_token} and \texttt{P3\_CLS} are so weak.
The benchmark therefore argues against naive positional defaults more clearly than
it anoints any single design family.

Two structural failure modes are also visible. \texttt{S2\_SIF} performs poorly (0.6057), indicating that down-weighting high-frequency tokens---designed for generic sentence similarity---is actively harmful for the discourse-style concepts in this corpus. \texttt{P2\_first\_token} collapses to chance (0.5018) because the BOS token carries no passage-specific content. Construction-method robustness is addressed below.

Users may ask whether the pooling conclusion is an artefact of fixing construction to DiffMean.

\begin{figure}[t]
  \centering
  \begin{subfigure}[b]{0.47\linewidth}
    \centering
    \includegraphics[width=\linewidth]{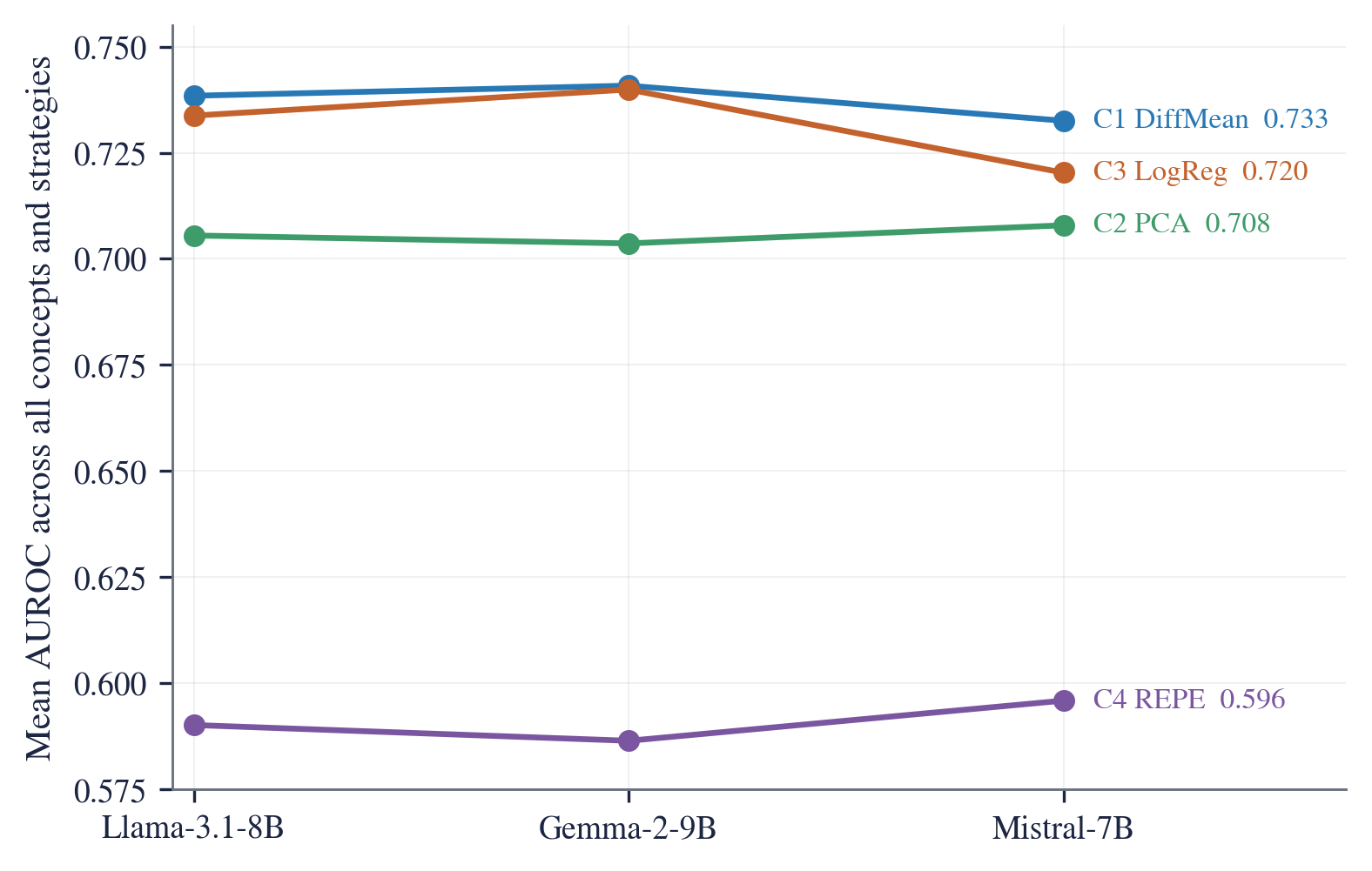}
    \caption{Construction-method comparison across the three benchmark models.
    DiffMean and logistic-regression directions are close on Llama and Gemma, but
    only DiffMean remains strong on Mistral.
    REPE is uniformly weak; the dominant construction effect is larger in absolute
    terms than the pooling gap the leaderboard measures.}
    \label{fig:construction}
  \end{subfigure}
  \hfill
  \begin{subfigure}[b]{0.47\linewidth}
    \centering
    \includegraphics[width=\linewidth]{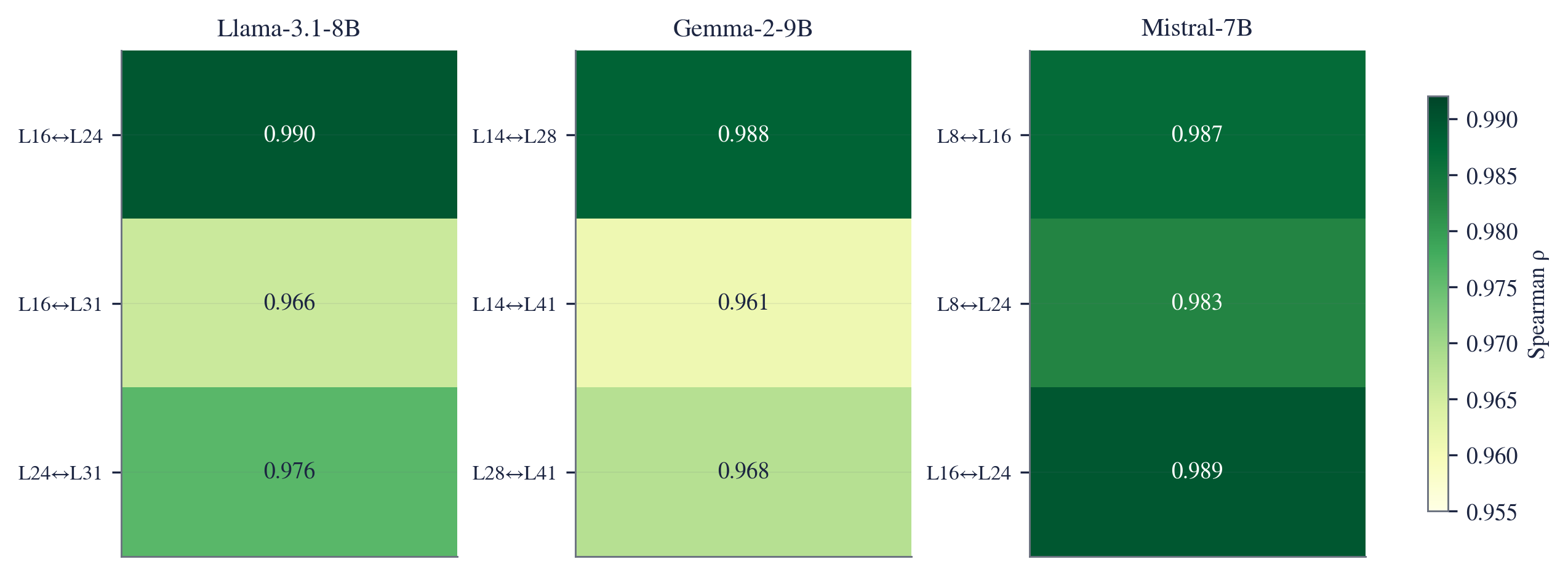}
    \caption{Layer-rank stability across candidate layers ($\rho$ range 0.96--0.99).
    Every within-model comparison remains above $\rho=0.96$, ruling out the
    leaderboard as an artefact of the released layer choice.}
    \label{fig:layercorr}
  \end{subfigure}
  \caption{Left: construction-method AUROC comparison. Right: Spearman layer-rank
  correlation confirming leaderboard stability across candidate layers.}
  \label{fig:construction_layercorr}
\end{figure}

\begin{table}[t]
\centering
\caption{Construction-method comparison, averaged across all 17 concepts and 19
strategies at the best layer for each method-model pair.}
\label{tab:construction}
\small
\begin{tabular}{lcccc}
\toprule
Method & Llama-3.1-8B & Gemma-2-9B & Mistral-7B & Cross-model mean \\
\midrule
\texttt{C1\_DiffMean} & 0.7384 & 0.7408 & 0.7325 & \textbf{0.7373} \\
\texttt{C3\_LogReg} & 0.7338 & 0.7399 & 0.7203 & 0.7313 \\
\texttt{C2\_PCA} & 0.7055 & 0.7036 & 0.7079 & 0.7057 \\
\texttt{C4\_REPE} & 0.5901 & 0.5864 & 0.5959 & 0.5908 \\
\bottomrule
\end{tabular}
\end{table}

The simplest construction method, DiffMean, is the strongest overall.
DiffMean and logistic regression are nearly tied on Gemma (0.7408 vs.\ 0.7399),
but logistic regression drops sharply on Mistral (0.7203) compared to DiffMean
(0.7325).
PCA is consistently lower but relatively flat across models.
REPE remains low everywhere with no compensating advantage.
The cross-model construction gap between DiffMean (0.7373) and REPE (0.5908) is
0.1465 AUROC—roughly nine times larger than the pooling gap between
\texttt{W4\_hierarchical} and \texttt{P1\_last\_token} (0.0159).
That asymmetry explains the primary benchmark design: construction must be fixed if
pooling is to be isolated rather than buried under larger upstream variance.
A concept-level breakdown of C1--C4 is in Appendix~\ref{app:construction};
extended views of pairwise method correlations, optimal layer selection, and
fallback diagnostics are in Appendix~\ref{app:construction_ext}.

\subsection{Stability, concept difficulty, and D1--D2 dissociation}
\label{sec:stability}
Figure~\ref{fig:layercorr} reports layer-wise rank correlation across all
$17 \times 19$ concept--strategy cells: Spearman's $\rho$ ranges from 0.961 to
0.990 across every within-model candidate-layer pair, and mean ICC is
0.8773 (Llama), 0.8586 (Gemma), and 0.9217 (Mistral), indicating that the
leaderboard ordering is invariant to the released layer choice.
Residual fragility is concept-specific: \texttt{contrast} and \texttt{causation}
are the weakest-ICC concepts on Llama and Gemma; \texttt{toxicity} is weakest on
Mistral.

\begin{figure}[t]
  \centering
  \begin{subfigure}[b]{0.47\linewidth}
    \centering
    \includegraphics[width=\linewidth]{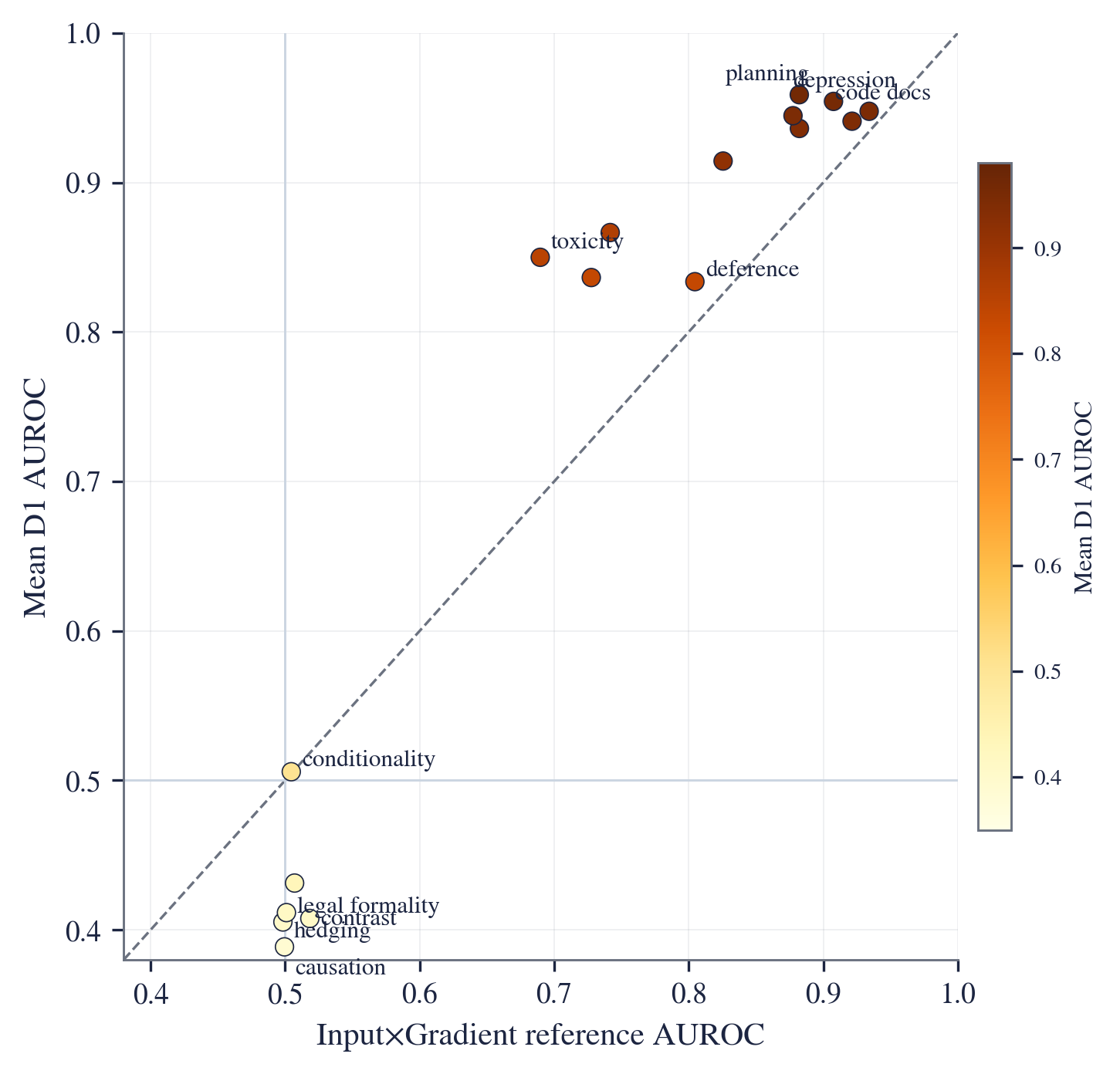}
    \caption{Mean D1 AUROC versus the Input$\times$Gradient reference.
    Hard concepts cluster near chance; mid-difficulty concepts (upper-left quadrant)
    are where pooling strategy choice carries practical weight.}
    \label{fig:concepts}
  \end{subfigure}
  \hfill
  \begin{subfigure}[b]{0.47\linewidth}
    \centering
    \includegraphics[width=\linewidth]{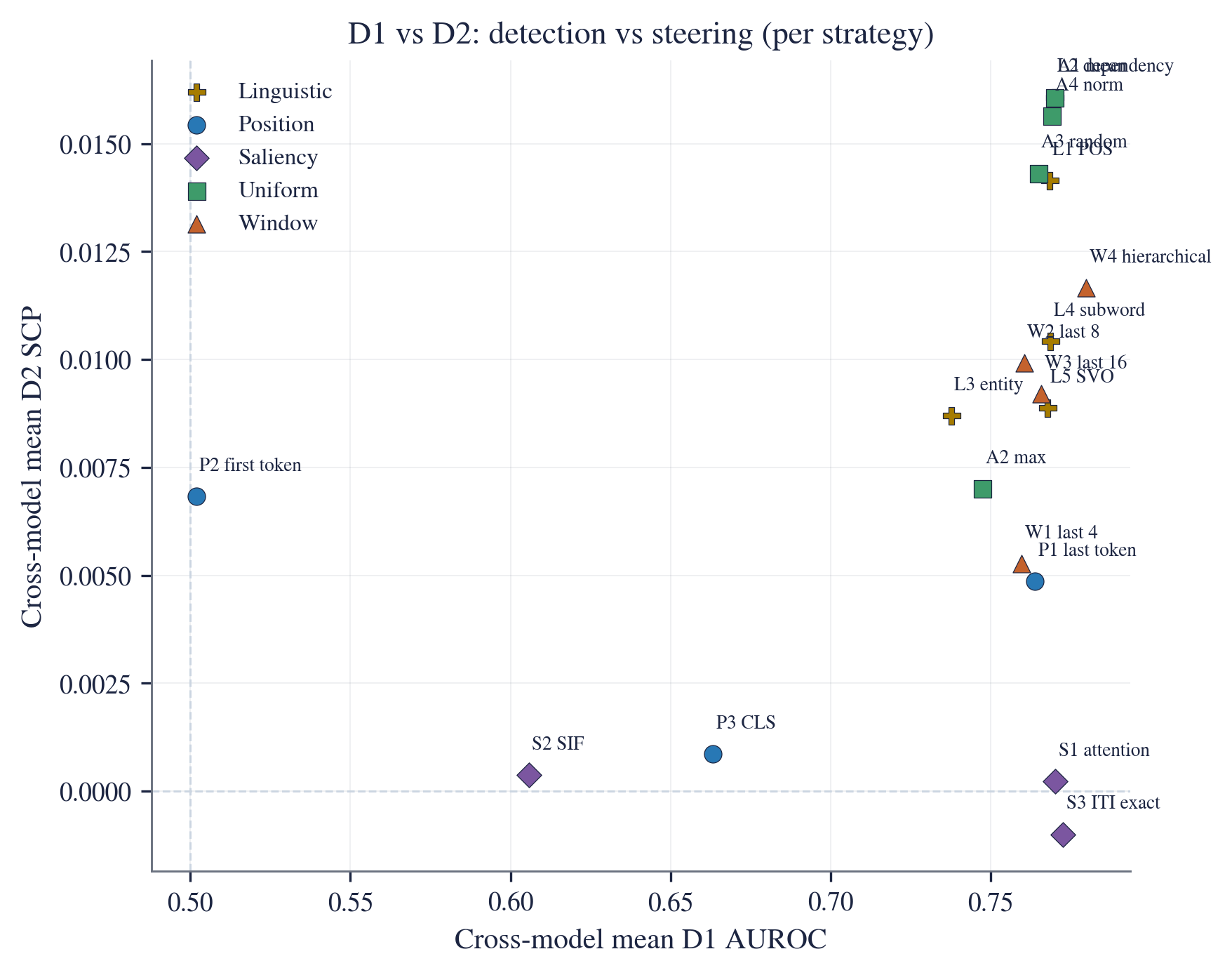}
    \caption{D1 AUROC vs.\ D2 SCP per strategy.
    Shape encodes strategy family.
    Saliency strategies (\texttt{S1}, \texttt{S3}) are the strongest D1 detectors
    but cluster near zero D2; uniform strategies are most consistent across both axes.}
    \label{fig:d1vsd2}
  \end{subfigure}
\end{figure}

Figure~\ref{fig:concepts} separates representation-limited concepts from
pooling-sensitive ones by comparing each concept's mean D1 AUROC against its
Input$\times$Gradient reference score.
The easiest concepts reach cross-model mean AUROC above 0.94
(\texttt{planning}: 0.9587; \texttt{depression}: 0.9542; \texttt{code\_docs}:
0.9476; \texttt{bureaucratic}: 0.9447).
The five hardest---\texttt{causation} (0.3887), \texttt{hedging} (0.4053),
\texttt{contrast} (0.4077), \texttt{legal\_formality} (0.4116), and
\texttt{negation\_density}---remain near chance under both the probe and the
IxG oracle, confirming a representational limit at the selected layer.
Practically meaningful pooling comparisons reside in the seven mid-difficulty
concepts with cross-model AUROC 0.70--0.90, where \texttt{W4\_hierarchical}
outperforms \texttt{P1\_last\_token} by 0.042--0.113~AUROC per concept per model.
Full oracle gap and ICC analyses are in Appendices~\ref{app:oracle}--\ref{app:icc}.

Figure~\ref{fig:d1vsd2} plots each strategy's cross-model mean D1 AUROC against
its D2 SCP, revealing that high linear separability does not uniformly transfer to
large steering effects.
\texttt{S3\_ITI\_exact} ranks second in D1 but records D2~SCP near or below zero
on Llama and Gemma; \texttt{S1\_attention\_weighted} follows the same pattern.
Uniform strategies (\texttt{A1\_mean}, \texttt{A3\_random}, \texttt{A4\_norm})
are the most consistent across both axes, with \texttt{W4\_hierarchical}
showing slightly better D2 than \texttt{P1\_last\_token} on Gemma and Mistral
but near-equivalent on Llama ($+$0.0004).
D3 is heterogeneous and is only meaningfully interpreted when $|\Delta_A| > 0.01$.
Most concept--strategy cells do not clear this gate, reflecting the small D2
SCP denominators.
Of the four concepts that do pass the gate (\texttt{narrative}, \texttt{planning},
\texttt{bureaucratic}, \texttt{deference}), D3 ratios are generally close to or
below~1: the target concept prevalence increases more than the neighbour, indicating
reasonably specific steering.
Saliency strategies (\texttt{S1}, \texttt{S3}) exhibit higher D3 ratios on Llama,
suggesting a tradeoff between detection specificity in hidden-state space and
output-level concept specificity.
Uniform strategies (\texttt{A1\_mean}, \texttt{A4\_norm}) rank near the top on D3,
consistent with their D2 performance.
The D1 vs.~D3 scatter (Appendix~\ref{app:d3}) confirms that high D1 does not
predict low D3: detection accuracy and steering specificity are empirically
independent.
Extended D2 views, alpha-response curves, and full D3 results are in
Appendices~\ref{app:d2_extended}--\ref{app:d3}.

\section{Discussion}
Pooling choice is a statistically consequential variable ($p = 2.0 \times 10^{-36}$;
77 Nemenyi-significant pairs): \texttt{W4\_hierarchical} (cross-model mean AUROC
0.7799, range 0.0078) is the most reliable general-purpose unsupervised strategy,
and \texttt{P1\_last\_token} is significantly outperformed despite its ubiquity.
For steering-focused use-cases, uniform strategies (\texttt{A1\_mean},
\texttt{A4\_norm}) trade a small D1 deficit for more consistent D2 SCP behaviour.

The D1--D2 dissociation is the benchmark's key methodological finding: high linear
separability does not imply effective steering, and \poolbench{} provides the first
controlled measurement of this gap across 19~strategies---separating pooling
failures from fundamental representational limits that no pooling innovation can
resolve.
The current scope is intentionally focused: three 7--9B English base models, one
construction method (DiffMean), and real-text passages.
Instruction-tuned variants, larger checkpoints, multilingual corpora, and
additional construction methods are natural extensions; the released corpus,
activations, and evaluation protocol are designed for exactly this kind of
incremental expansion (Appendix~\ref{app:artifacts}).

\section{Conclusion}
Every activation-steering pipeline makes an implicit bet on a pooling rule.
We provide evidence that pooling choice is statistically significant,
the default is not the best option, and the performance gap is largest
in the concepts where it matters most.
We establishes that the ceiling on steering quality
is set by representation, not by pooling: D1 and D2 dissociate systematically, and
that dissociation is invisible without a controlled multi-strategy protocol.
Knowing \emph{why} a direction fails to steer---whether the pooling rule discards
concept signal or whether the concept simply lacks a clean linear subspace at the
probed layer---is the diagnostic gap this benchmark fills.
We release the corpus, activations, and evaluation protocol so that future work can
ask harder versions of this question across models, languages, and concepts that
do not yet have an answer.

\begin{ack}
The author thanks the open-source community for tools and datasets that made this benchmark possible.
\end{ack}

\clearpage
\section*{References}
\bibliographystyle{plainnat}

\clearpage
\appendix
\renewcommand{\thesection}{\AlphAlph{\value{section}}}


\clearpage
\section*{Appendix}
\addcontentsline{toc}{section}{Appendix}

This appendix provides complete supplementary materials for all benchmark sections.
Each section follows the evaluation pipeline in execution order: corpus construction,
power analysis, activation extraction and layer selection, concept linearity,
D1 detection leaderboard, construction-method comparison, D2 activation steering,
D3 concept disentanglement, oracle and attribution reference, interpretability
diagnostics, cross-metric synthesis, concept family reference, and artifact release.

\clearpage
\section{Corpus Construction: Sources and Negative Construction}
\label{app:sources}

Before the benchmark sweep, each of the 17 concepts was assigned a construction mode
(paired rewrite or independent sampling) based on whether the concept's signal is
carried by local surface markers or by distributed content.
Paired-rewrite concepts use the same source passage for both the positive and the
negative, with the target markers removed or replaced; this design eliminates
vocabulary and topic confounds entirely.
Independent-sampling concepts draw positives and negatives from distinct, domain-disjoint
source corpora to prevent the probe from learning domain membership as a proxy.
Tables~\ref{tab:sources_a} and \ref{tab:sources_b} record the complete source map.
Every concept must span at least three source domains (Appendix~\ref{app:corpusqa});
\texttt{deference} has a documented two-domain exception because the Intel Polite-Guard
dataset is the only sizable English-language deference corpus currently available.

\begin{table}[h]
\centering
\caption{Released source map for matched-pair and rewrite-based concepts.}
\label{tab:sources_a}
\scriptsize
\resizebox{\linewidth}{!}{%
\begin{tabular}{lp{1.7cm}p{4.2cm}p{4.8cm}p{1.2cm}}
\toprule
Concept & Mode & Positive source(s) & Negative source(s) or construction & Domains \\
\midrule
\texttt{hedging} & Paired rewrite & \href{https://huggingface.co/datasets/gfissore/arxiv-abstracts-2021}{\texttt{gfissore/arxiv-abstracts-2021}}, \href{https://huggingface.co/datasets/cc\_news}{\texttt{cc\_news}}, \href{https://huggingface.co/datasets/FiscalNote/billsum}{\texttt{FiscalNote/billsum}}~\citep{kornilova2019billsum} & Same-source passages with hedge markers removed or replaced by assertive equivalents & 3 \\
\texttt{legal\_formality} & Paired rewrite & \href{https://huggingface.co/datasets/coastalcph/lex_glue}{\texttt{lex\_glue/scotus}}~\citep{chalkidis2021lexglue}, \href{https://huggingface.co/datasets/coastalcph/lex_glue}{\texttt{lex\_glue/eurlex}}, \href{https://huggingface.co/datasets/FiscalNote/billsum}{\texttt{FiscalNote/billsum}} & Same-source passages with legal markers removed or paraphrased into plain language & 3 \\
\texttt{frustration} & Paired rewrite & \href{https://huggingface.co/datasets/Yelp/yelp\_review\_full}{\texttt{Yelp/yelp\_review\_full}}~\citep{zhang2015text}, \href{https://huggingface.co/datasets/sentence-transformers/reddit}{\texttt{sentence-transformers/reddit}}~\citep{baumgartner2020pushshift}, \href{https://huggingface.co/datasets/cc\_news}{\texttt{cc\_news}} & Same-source passages with frustration markers softened or removed & 3 \\
\texttt{causation} & Paired rewrite & \href{https://huggingface.co/datasets/gfissore/arxiv-abstracts-2021}{\texttt{gfissore/arxiv-abstracts-2021}}, \href{https://huggingface.co/datasets/cc\_news}{\texttt{cc\_news}}, \href{https://huggingface.co/datasets/FiscalNote/billsum}{\texttt{FiscalNote/billsum}} & Same-source passages with explicit causal connectives removed & 3 \\
\texttt{contrast} & Paired rewrite & \href{https://huggingface.co/datasets/gfissore/arxiv-abstracts-2021}{\texttt{gfissore/arxiv-abstracts-2021}}, \href{https://huggingface.co/datasets/cc\_news}{\texttt{cc\_news}}, \href{https://huggingface.co/datasets/FiscalNote/billsum}{\texttt{FiscalNote/billsum}} & Same-source passages with adversative connectives replaced by additive or neutral links & 3 \\
\texttt{conditionality} & Paired rewrite & \href{https://huggingface.co/datasets/gfissore/arxiv-abstracts-2021}{\texttt{gfissore/arxiv-abstracts-2021}}, \href{https://huggingface.co/datasets/cc\_news}{\texttt{cc\_news}}, \href{https://huggingface.co/datasets/FiscalNote/billsum}{\texttt{FiscalNote/billsum}} & Same-source passages with conditional markers removed and clauses rewritten declaratively & 3 \\
\texttt{negation\_density} & Lexical filter & \href{https://huggingface.co/datasets/gfissore/arxiv-abstracts-2021}{\texttt{gfissore/arxiv-abstracts-2021}} & Same-source passages with negation operators removed and clauses rewritten affirmatively & 1 \\
\bottomrule
\end{tabular}%
}
\end{table}

\begin{table}[h]
\centering
\caption{Released source map for independently sampled concepts.}
\label{tab:sources_b}
\scriptsize
\resizebox{\linewidth}{!}{%
\begin{tabular}{lp{1.8cm}p{4.3cm}p{4.6cm}p{1.2cm}}
\toprule
Concept & Mode & Positive source(s) & Negative source(s) or construction & Domains \\
\midrule
\texttt{imdb\_sentiment} & Independent & \href{https://huggingface.co/datasets/stanfordnlp/imdb}{\texttt{stanfordnlp/imdb}}~\citep{maas2011learning} (positive) & \href{https://huggingface.co/datasets/stanfordnlp/imdb}{\texttt{stanfordnlp/imdb}} (negative) & 1 \\
\texttt{toxicity} & Independent & \href{https://huggingface.co/datasets/tdavidson/hate\_speech\_offensive}{\texttt{tdavidson/hate\_speech\_offensive}}~\citep{davidson2017hate}, \href{https://huggingface.co/datasets/google/civil\_comments}{\texttt{google/civil\_comments}}~\citep{borkan2019nuanced}, Surge-AI CSV & Non-toxic labels from the same sources under stricter hostile-text exclusion rules & 3 \\
\texttt{depression} & Independent & \href{https://huggingface.co/datasets/mrjunos/depression-reddit-cleaned}{\texttt{mrjunos/depression-reddit-cleaned}} & \href{https://huggingface.co/datasets/dlb/mentalreddit}{\texttt{dlb/mentalreddit}} with depression markers removed & 1 \\
\texttt{academic\_tone} & Domain filter & \href{https://huggingface.co/datasets/gfissore/arxiv-abstracts-2021}{\texttt{gfissore/arxiv-abstracts-2021}} & \href{https://huggingface.co/datasets/sentence-transformers/reddit}{\texttt{sentence-transformers/reddit}}~\citep{baumgartner2020pushshift} & 2 \\
\texttt{code\_docs} & Domain filter & \href{https://huggingface.co/datasets/Nan-Do/code-search-net-python}{\texttt{Nan-Do/code-search-net-python}}~\citep{husain2019codesearchnet} & \href{https://huggingface.co/datasets/sentence-transformers/reddit}{\texttt{sentence-transformers/reddit}} & 2 \\
\texttt{bureaucratic} & Domain filter & \href{https://huggingface.co/datasets/FiscalNote/billsum}{\texttt{FiscalNote/billsum}}~\citep{kornilova2019billsum} & \href{https://huggingface.co/datasets/Yelp/yelp\_review\_full}{\texttt{Yelp/yelp\_review\_full}}~\citep{zhang2015text} & 2 \\
\texttt{narrative} & Domain filter & \href{https://huggingface.co/datasets/euclaise/writingprompts}{\texttt{euclaise/writingprompts}} & \href{https://huggingface.co/datasets/wikimedia/wikipedia}{\texttt{wikimedia/wikipedia 20231101.en}} & 2 \\
\texttt{deference} & Label filter & \href{https://huggingface.co/datasets/Intel/polite-guard}{\texttt{Intel/polite-guard}}~\citep{intel2023politeguard} (\textit{polite} + \textit{somewhat polite}) & \href{https://huggingface.co/datasets/Intel/polite-guard}{\texttt{Intel/polite-guard}} (\textit{neutral} + \textit{impolite}) & 1 \\
\texttt{planning} & Text/source filter & \href{https://huggingface.co/datasets/gursi26/wikihow-cleaned}{\texttt{gursi26/wikihow-cleaned}}, \href{https://huggingface.co/datasets/sentence-transformers/reddit}{\texttt{sentence-transformers/reddit}}, \href{https://huggingface.co/datasets/Yelp/yelp\_review\_full}{\texttt{Yelp/yelp\_review\_full}} & Same 3 sources, filtered to exclude planning-keyword passages & 3 \\
\texttt{numerical\_precision} & Lexical filter & \href{https://huggingface.co/datasets/gfissore/arxiv-abstracts-2021}{\texttt{gfissore/arxiv-abstracts-2021}} & \href{https://huggingface.co/datasets/cc\_news}{\texttt{cc\_news}} passages with zero digits and vague quantifiers & 2 \\
\bottomrule
\end{tabular}%
}
\end{table}

\section{Corpus Quality: Released Counts and Integrity Audit}
\label{app:corpusqa}

After source collection and construction, the corpus underwent a three-stage
integrity audit before the benchmark sweep.
First, MD5-based within-class deduplication removed 329 duplicate passages that had
been ingested from overlapping source files.
Second, train/test deduplication removed 87 passages whose exact text appeared in
both the training and test splits, eliminating leakage that would inflate test AUROC.
Third, cross-label deduplication removed 5 passages from \texttt{depression} that
also appeared as negatives in an adjacent concept, preventing the probe from learning
cross-concept signal.
Seed-word contamination is enforced at build time: negative passages are rejected if
they contain any of the concept's defining seed markers.
Table~\ref{tab:corpusqa} records the final released counts after all deduplication.
Figure~\ref{fig:corpussize} shows the same information visually.
All 17 concepts pass the pre-committed power criterion; the contamination ablation
analysis confirming that surviving passages do not carry incidental confound signal
is in Appendix~\ref{app:ablation}.

\begin{table}[t]
\centering
\caption{Released corpus QA summary. The benchmark targets 700/300 train/test per
class; all released concepts remain adequately powered under the stored power
analysis even when counts fall slightly below target.}
\label{tab:corpusqa}
\small
\resizebox{\linewidth}{!}{%
\begin{tabular}{lccccp{4.7cm}}
\toprule
Concept & Family & Train per class & Test per class & Domains & Notes \\
\midrule
\texttt{academic\_tone} & register & 700 & 295--300 & 6 & Near target after deduplication. \\
\texttt{bureaucratic} & register & 700 & 298--300 & 6 & Expanded marker list after audit. \\
\texttt{causation} & syntactic & 699 & 295--298 & 3 & Hard concept; low ICC and low IxG reference score. \\
\texttt{code\_docs} & register & 700 & 294--300 & 3 & High-signal register concept. \\
\texttt{conditionality} & syntactic & 700 & 273 & 3 & Adequately powered but lower released test count. \\
\texttt{contrast} & syntactic & 698 & 299 & 3 & Hard concept; weakest ICC across models. \\
\texttt{deference} & semantic-abstract & 700 & 300 & 2 & Documented domain exception. \\
\texttt{depression} & dense-lexical & 562 & 241 & 1 & Source-limited but still adequately powered (CI $\pm 0.043$). \\
\texttt{frustration} & sparse-lexical & 700 & 234--298 & 3 & Regex bug fixed during audit. \\
\texttt{hedging} & sparse-lexical & 697 & 295--298 & 3 & Hard concept; low IxG reference score. \\
\texttt{imdb\_sentiment} & dense-lexical & 700 & 300 & 1 & Single-domain by design. \\
\texttt{legal\_formality} & sparse-lexical & 699 & 296--299 & 3 & Hard concept; low IxG reference score. \\
\texttt{narrative} & semantic-abstract & 700 & 276--300 & 3 & High-signal narrative register concept. \\
\texttt{negation\_density} & syntactic & 699 & 299--300 & 3 & Hard concept; IxG reference near chance. \\
\texttt{numerical\_precision} & sparse-lexical & 700 & 300 & 5 & Strong signal and strong IxG reference score. \\
\texttt{planning} & semantic-abstract & 700 & 267--300 & 3 & Easiest released concept overall. \\
\texttt{toxicity} & dense-lexical & 700 & 300 & 3 & Relaxed short-text token window by design. \\
\bottomrule
\end{tabular}%
}
\end{table}

\section{Corpus Size Distribution}
\label{app:corpus_size}

\begin{figure}[H]
  \centering
  \includegraphics[width=0.80\linewidth]{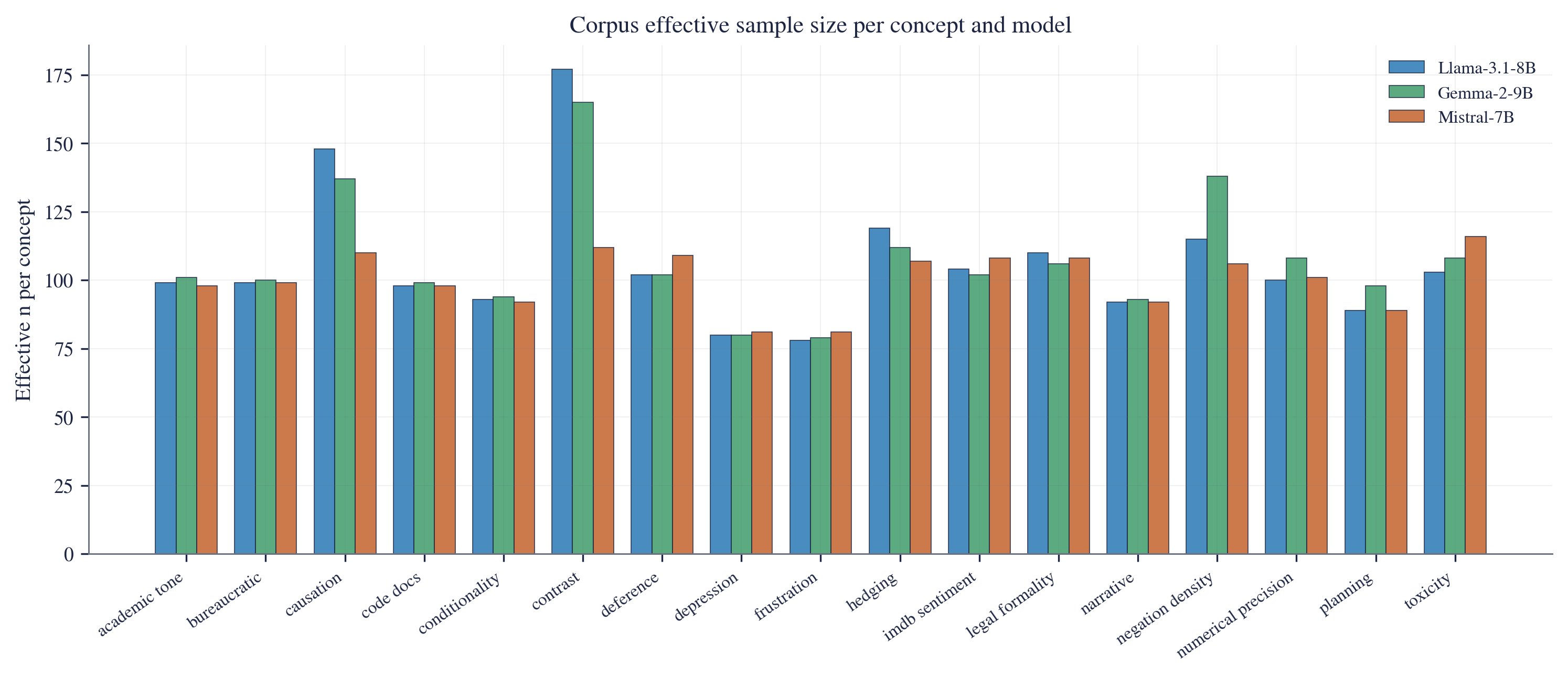}
  \caption{Released passage counts per concept (train positive, train negative, test
  positive, test negative) sorted alphabetically.
  Dashed reference lines mark the benchmark targets of 700 train and 300 test passages
  per class.
  \texttt{depression} is visibly below target on both axes due to source exhaustion;
  the pre-committed power analysis confirms it remains adequately powered for the stated
  0.05 AUROC MDE.
  \texttt{conditionality} and \texttt{frustration} have modestly reduced test counts
  after deduplication but still clear the power threshold.
  All other concepts meet or closely approach both targets.}
  \label{fig:corpussize}
\end{figure}

Figure~\ref{fig:corpussize} illustrates the released passage counts per concept
alongside the benchmark targets, complementing the numerical summary in
Table~\ref{tab:corpusqa}.

\section{Contamination and Ablation Analysis}
\label{app:ablation}

\begin{figure}[H]
  \centering
  \includegraphics[width=0.72\linewidth]{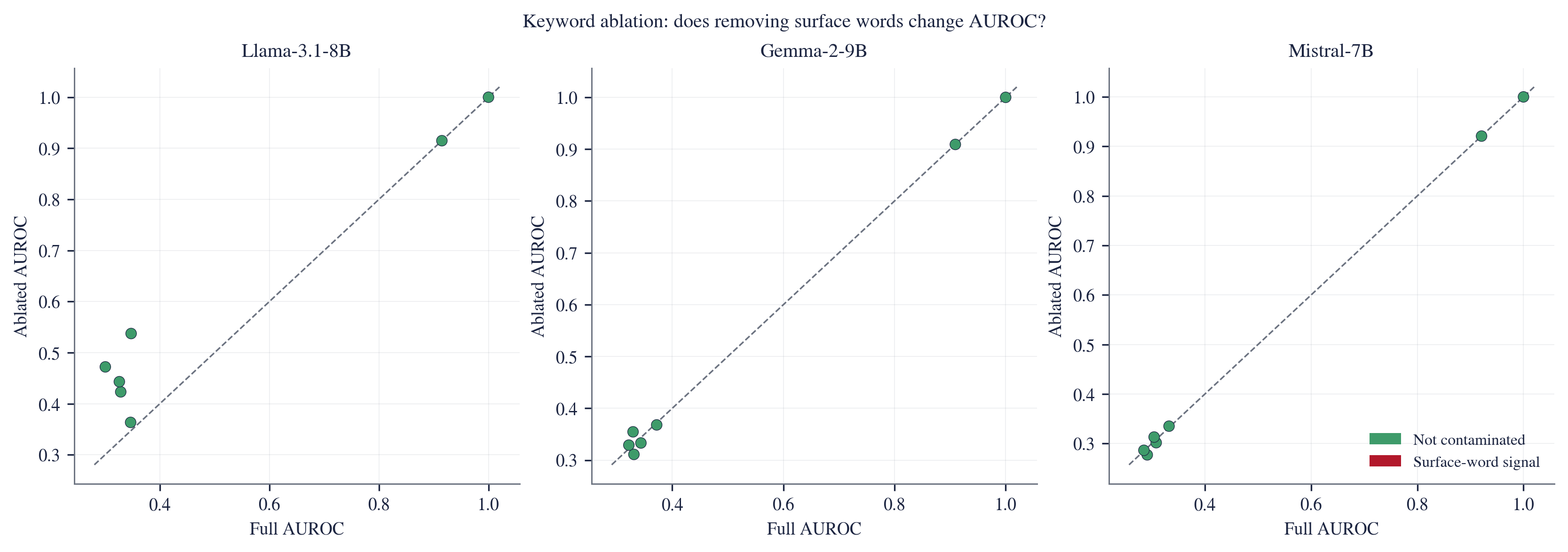}
  \caption{Per-concept probe AUROC before (x-axis) vs.\ after (y-axis) ablation of the
  strongest confound-concept signal.
  Points on the diagonal indicate no sensitivity to ablation.
  Points above the diagonal indicate that removing confound signal improves AUROC
  (the probe was exploiting spurious features that hurt generalisation).
  Points below the diagonal indicate a small cost of masking.
  The majority of concepts cluster tightly near the diagonal, confirming that the corpus
  is not dominated by incidental confound signal.}
  \label{fig:ablationscatter}
\end{figure}

\begin{figure}[H]
  \centering
  \includegraphics[width=0.72\linewidth]{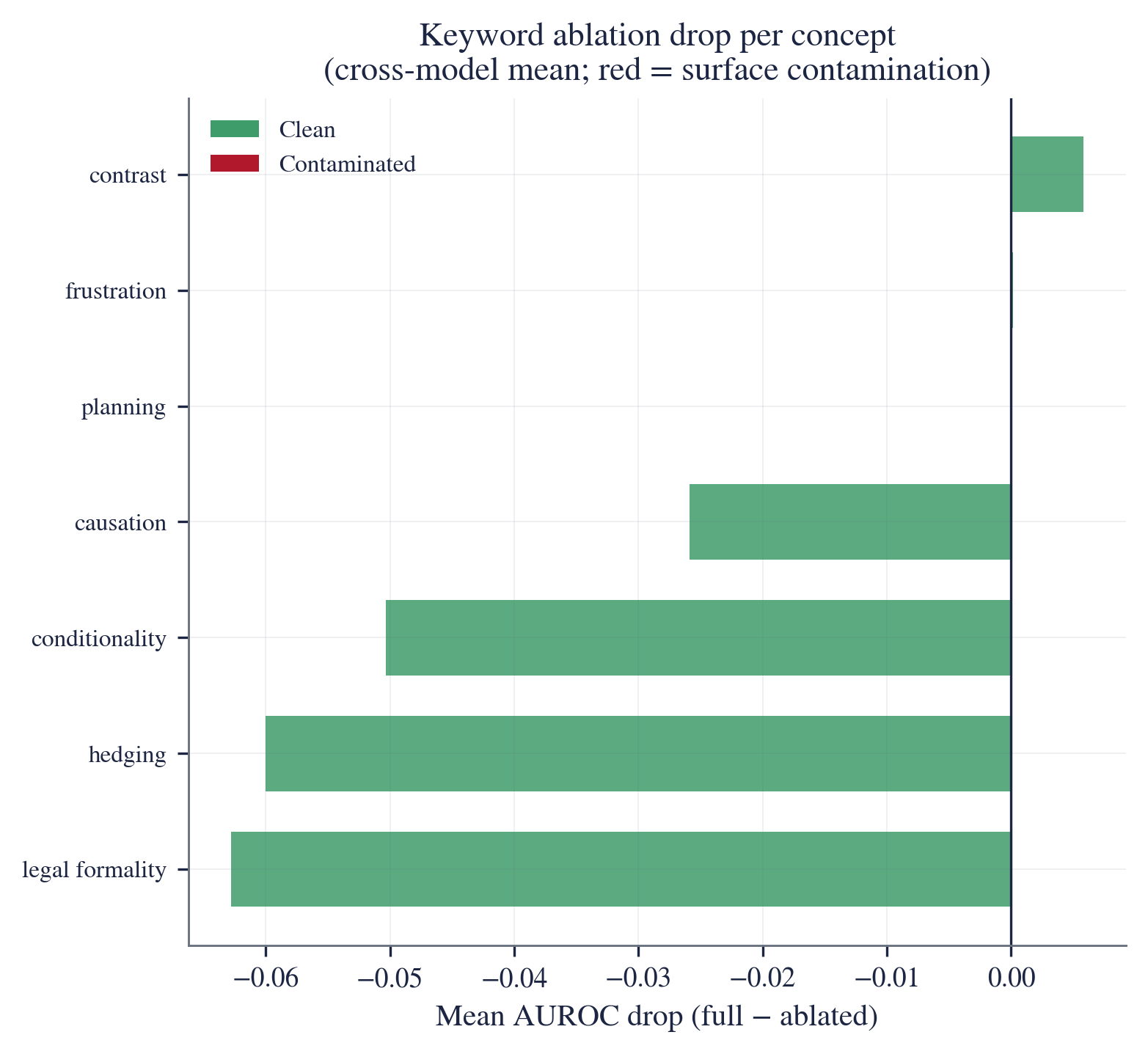}
  \caption{Per-concept AUROC change after ablation, split by model.
  The hard syntactic concepts (\texttt{causation}, \texttt{contrast}, \texttt{hedging})
  show the largest cross-model variance, consistent with their overall representational
  difficulty rather than a corpus contamination problem.
  No concept shows a systematic large degradation.}
  \label{fig:ablationbar}
\end{figure}

\begin{figure}[H]
  \centering
  \includegraphics[width=0.72\linewidth]{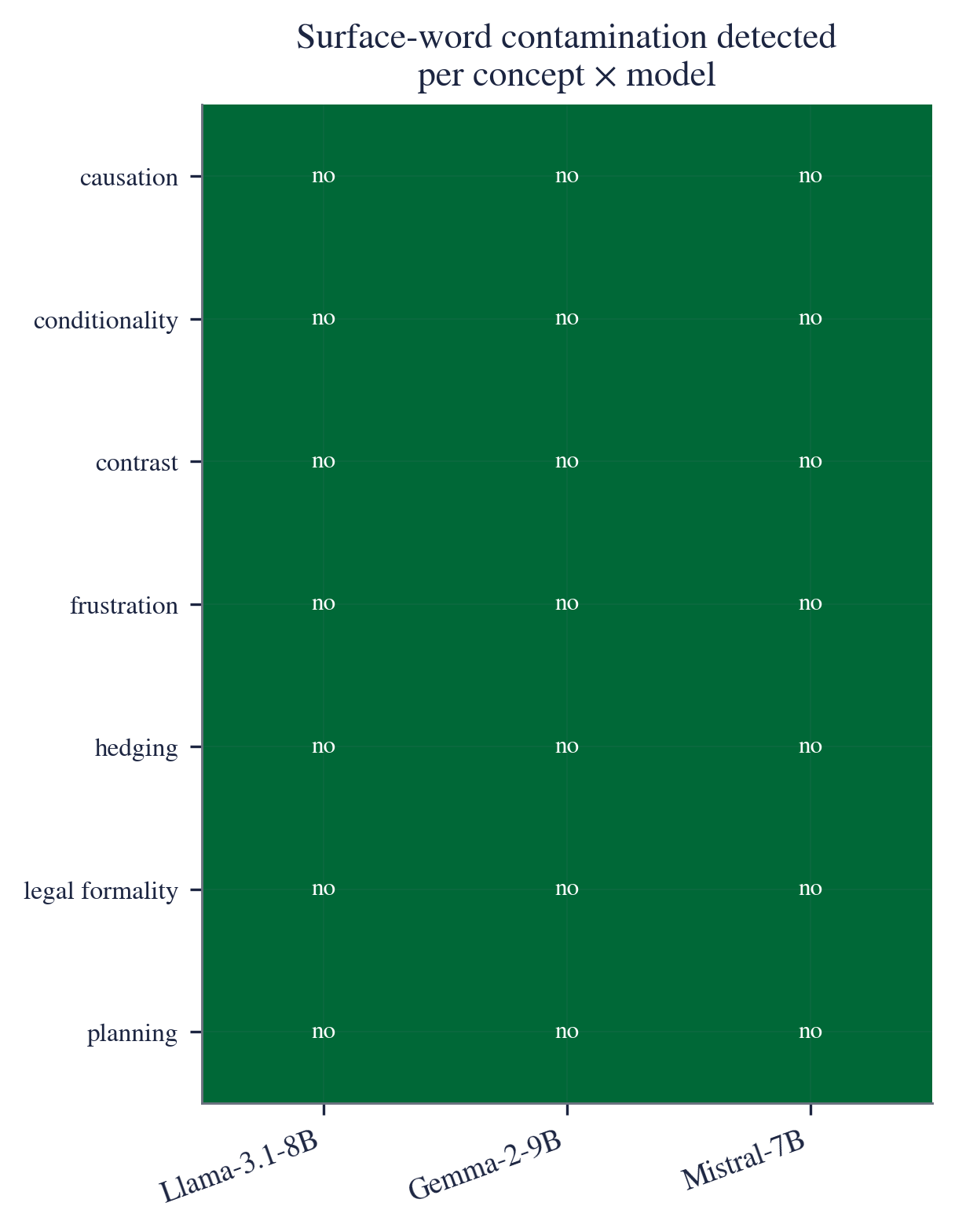}
  \caption{Boolean contamination heatmap.
  Each row is a target concept; each column is a potential confound concept.
  A cell is shaded if the target concept's probe shows substantial sensitivity ($>2\%$
  AUROC shift) to ablation of confound tokens from the corresponding column concept
  in at least one model.
  The pattern is sparse.
  The denser sub-block in the upper-left corresponds to syntactically related concepts
  (\texttt{causation} / \texttt{contrast} / \texttt{conditionality}), where some
  token-level overlap is expected by design and is kept as genuine shared signal.}
  \label{fig:ablationcontam}
\end{figure}

A potential threat to the benchmark's validity is that some positive examples carry
unintended signal from a confound concept---particularly for semantically adjacent pairs
such as \texttt{hedging}/\texttt{contrast} or \texttt{planning}/\texttt{conditionality}.
To test this, we ran an ablation in which the tokens most strongly associated with
each confound concept were masked and the probe was re-evaluated.
If a probe's AUROC drops substantially after masking, it was exploiting the confound
rather than the target concept's own signal.
Figures~\ref{fig:ablationscatter}--\ref{fig:ablationcontam} show that contamination
is sparse and does not materially threaten the benchmark's validity.

\section{Power Analysis and Evaluation Pre-commitments}
\label{app:power}

All evaluation thresholds and statistical criteria were committed to
\texttt{results/power\_analysis.json} before the strategy sweep began.
This pre-registration prevents post-hoc adjustment of the significance threshold or
minimum detectable effect to suit observed results.
Table~\ref{tab:power} records the concrete evaluation geometry that makes benchmark
results interpretable.
The cross-concept CI half-width of $\pm 0.0097$ means that reported differences in
cross-concept mean AUROC greater than $\sim 0.015$ are reliably non-zero at the
benchmark's planned sample size; all headline comparisons in the main paper exceed
this threshold by a wide margin.

\begin{table}[t]
\centering
\caption{Released power-analysis and evaluation-design summary. All values are
taken from \texttt{results/power\_analysis.json}.}
\label{tab:power}
\small
\begin{tabularx}{\linewidth}{@{}l>{\raggedright\arraybackslash}X@{}}
\toprule
Quantity & Released value \\
\midrule
Primary D1 protocol & 5-fold out-of-fold logistic probing on the 700-per-class training split \\
Held-out split usage & 300 passages per class reserved for auxiliary classifier-quality checks \\
Per-fold evaluation size & 140 passages per class-equivalent test fold \\
Bootstrap replicates & 5000 \\
Pre-committed minimum detectable effect & 0.05 AUROC \\
95\% CI half-width, per fold & $\pm 0.0276$ \\
95\% CI half-width, cross-concept mean over 17 concepts & $\pm 0.0097$ \\
Planned Nemenyi critical distance & 3.9 rank positions under the pre-registered effective-$N$ approximation \\
Released corpus size & 37{,}693 records across 17 concepts \\
Overall power status before sweep & PASS \\
\bottomrule
\end{tabularx}
\end{table}

\section{Activation Extraction and Layer Selection}
\label{app:diagnostics}

For each of the three benchmark models, three candidate layers were identified
a priori by dividing the total depth into early, middle, and late thirds.
All three candidate layers were registered simultaneously in a single forward pass
per passage using PyTorch hooks (commit \texttt{065a72c}), which ensures that each
layer sees identical input and that no additional inference cost is incurred over
single-layer extraction.
The best layer was selected as the one that maximises cross-concept mean D1 AUROC
under the \texttt{A1\_mean} strategy on the training fold; using the mean of a
strategy that appears in the benchmark avoids a circular best-layer selection
criterion.
As a post-hoc validity check, Spearman rank correlations of per-concept AUROC
between all candidate-layer pairs confirmed that the strategy ranking is stable
regardless of which candidate layer is released.


\section{Pooling Strategy Definitions}
\label{app:strategies}

Table~\ref{tab:strategy_defs} gives the formal definition of all 19 evaluated
pooling strategies.
Let $h = (h_1, \ldots, h_T) \in \mathbb{R}^{T \times d}$ denote the per-token
residual-stream vectors for a passage of $T$ tokens at the selected layer.
The pooled representation $\phi \in \mathbb{R}^d$ is used for probe training and
steering vector construction.

\begin{table}[H]
\centering
\caption{Formal definitions of all 19 pooling strategies evaluated in \poolbench{}.
$h_i \in \mathbb{R}^d$ is the hidden state at token position $i$; $T$ is total
tokens; $w_i$ are scalar weights; $\lceil\cdot\rceil$ is ceiling division.
$\dagger$~\texttt{L2} collapses to \texttt{A1\_mean} on all passages due to a
parser implementation issue (zero dependency arcs matched).}
\label{tab:strategy_defs}
\small
\begin{tabularx}{\linewidth}{@{}l l X@{}}
\toprule
ID & Family & Definition \\
\midrule
\texttt{P1\_last\_token} & Position & $\phi = h_T$ \\
\texttt{P2\_first\_token} & Position & $\phi = h_1$ \\
\texttt{P3\_CLS} & Position & $\phi = h_1$ (BOS token, analogous to \texttt{[CLS]}) \\
\addlinespace
\texttt{A1\_mean} & Uniform & $\phi = \frac{1}{T}\sum_{i=1}^{T} h_i$ \\
\texttt{A2\_max} & Uniform & $\phi_j = \max_i h_{ij}$ (dimension-wise maximum) \\
\texttt{A3\_random} & Uniform & $\phi = h_k$, $k \sim \mathcal{U}\{1, \ldots, T\}$ \\
\texttt{A4\_norm} & Uniform & $\phi = \frac{1}{T}\sum_{i=1}^{T} \frac{h_i}{\|h_i\|_2 + \epsilon}$ \\
\addlinespace
\texttt{W1\_mean\_last\_4} & Window & $\phi = \frac{1}{4}\sum_{i=T-3}^{T} h_i$ \\
\texttt{W2\_mean\_last\_8} & Window & $\phi = \frac{1}{8}\sum_{i=T-7}^{T} h_i$ \\
\texttt{W3\_mean\_last\_16} & Window & $\phi = \frac{1}{16}\sum_{i=T-15}^{T} h_i$ \\
\texttt{W4\_hierarchical} & Window & Divide $h$ into $K=4$ equal contiguous chunks
  $C_1, \ldots, C_K$ where $|C_k| = \lceil T/K \rceil$.
  Compute the mean of each chunk: $\bar{c}_k = \frac{1}{|C_k|}\sum_{h_i \in C_k} h_i$.
  Pool chunk means: $\phi = \frac{1}{K}\sum_{k=1}^{K} \bar{c}_k$. \\
\addlinespace
\texttt{S1\_attention\_weighted} & Saliency & $\phi = \sum_{i=1}^{T} w_i h_i$,
  $w_i = \bar{a}_i / \sum_j \bar{a}_j$, where $\bar{a}_i$ is the mean inflow
  attention weight to token $i$ averaged across all heads at the selected layer. \\
\texttt{S2\_SIF} & Saliency & $\phi = \sum_{i=1}^{T} w_i h_i$,
  $w_i = \alpha / (\alpha + p_i)$ (Smooth Inverse Frequency~\citep{arora2017simple}
  with $\alpha = 10^{-3}$), projected away from the first principal component. \\
\texttt{S3\_ITI\_exact} & Saliency & Mean of $h_i$ restricted to the top-$K$
  attention heads ranked by per-head linear probe accuracy~\citep{li2023iti}
  ($K=10$, selected per concept on the training fold). \\
\addlinespace
\texttt{L1\_POS\_filtered} & Linguistic & Mean of $h_i$ for tokens whose
  part-of-speech tag is in \{NOUN, VERB, ADJ, ADV\} (content words only). \\
\texttt{L2\_dependency\_rel}$^\dagger$ & Linguistic & Mean of $h_i$ for tokens
  that are the head or dependent in a dependency arc. Falls back to
  \texttt{A1\_mean} when no arcs are matched. \\
\texttt{L3\_named\_entity} & Linguistic & Mean of $h_i$ for tokens inside a
  named-entity span (any NE type). Falls back to \texttt{A1\_mean}. \\
\texttt{L4\_subword\_root} & Linguistic & Mean of $h_i$ for the final subword
  token of each whitespace-delimited word (root subword pooling). \\
\texttt{L5\_SVO} & Linguistic & Mean of $h_i$ for tokens in subject--verb--object
  triples extracted by the dependency parser. Falls back to \texttt{A1\_mean}. \\
\bottomrule
\end{tabularx}
\end{table}

\texttt{W4\_hierarchical} is effectively a two-level mean pooling: rather than
averaging all tokens at once (as in \texttt{A1\_mean}), it first averages within
four contiguous segments, then averages those four segment means.
This down-weights extreme-position tokens implicitly (a single token in a small
edge segment contributes $1/4 \times 1/|C_k|$ rather than $1/T$) without the
sharp positional cutoff of the trailing-window strategies (\texttt{W1}--\texttt{W3}).


\section{Concept Linearity Analysis}
\label{app:concepts}

\begin{figure}[H]
  \centering
  \includegraphics[width=0.68\linewidth]{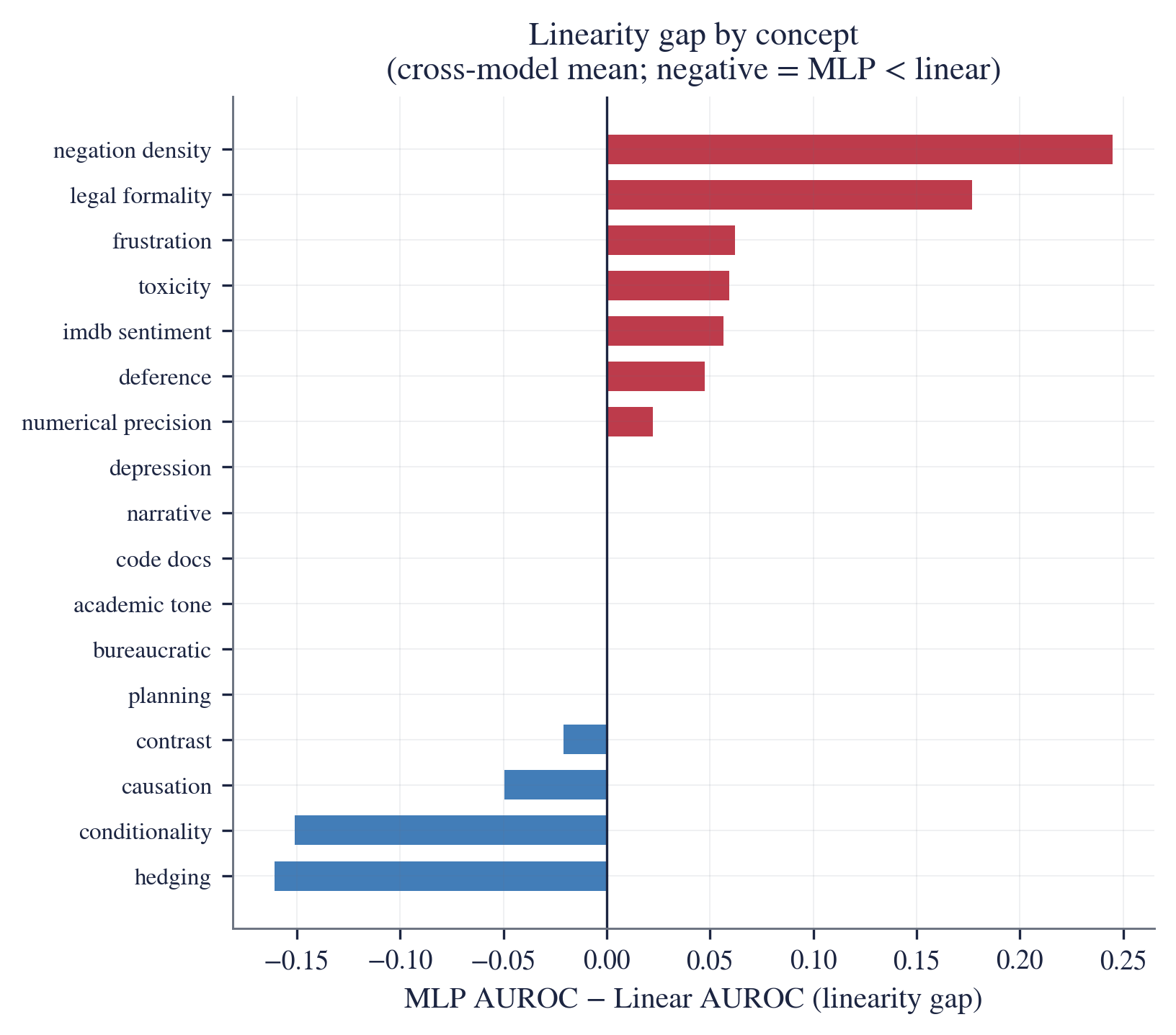}
  \caption{Linearity gap by concept (MLP AUROC $-$ linear AUROC, cross-model mean).
  Negative bars (grey) mean the MLP performs no better than linear, consistent with a
  cleanly linearly separable representation.
  Concepts with large positive gaps may rely on non-linear structure and their D1
  scores should be interpreted with that caveat.
  The pre-registered exclusion threshold ($0.03$ AUROC) is shown as a dashed line.}
  \label{fig:linearity}
\end{figure}

The benchmark's primary D1 probe is logistic regression---a linear classifier.
Before sweeping strategies, we test whether the linear assumption actually holds for
each concept by comparing logistic regression AUROC to a shallow MLP probe.
The linearity gap is defined as
$\Delta_{\mathrm{lin}}(c) = \mathrm{AUROC}_{\mathrm{MLP}} - \mathrm{AUROC}_{\mathrm{linear}}$
under the \texttt{A1\_mean} strategy at the best layer.
A gap greater than the pre-registered threshold of $0.03$ AUROC (60\% of the MDE)
indicates that the representation has exploitable non-linear structure and the concept
is flagged as non-linear.
Six concepts fail on Llama and Gemma; Mistral passes all six of those and one extra
(\texttt{imdb\_sentiment} at $+0.027$).
Flagged concepts are not removed from the leaderboard---all 17~concepts are reported
for completeness---but the non-linear flag is recorded so readers can interpret their
D1 scores accordingly.
Figure~\ref{fig:linearity} shows the gap by concept.
For the hardest syntactic concepts, both the linear probe and the IxG reference are
near chance (AUROC $\approx 0.40$--$0.45$), confirming that these are
representation-limited at the chosen layer and model family rather than
pooling-limited.

\section{Linearity: Per-Model and Pass/Fail Views}
\label{app:linearity}

Figure~\ref{fig:linearity} in Appendix~\ref{app:concepts} shows the cross-model
mean linearity gap.
This section provides the model-specific three-panel scatter and a boolean pass/fail
heatmap that surface model-to-model variation in the linearity structure.

\begin{figure}[H]
  \centering
  \includegraphics[width=\linewidth]{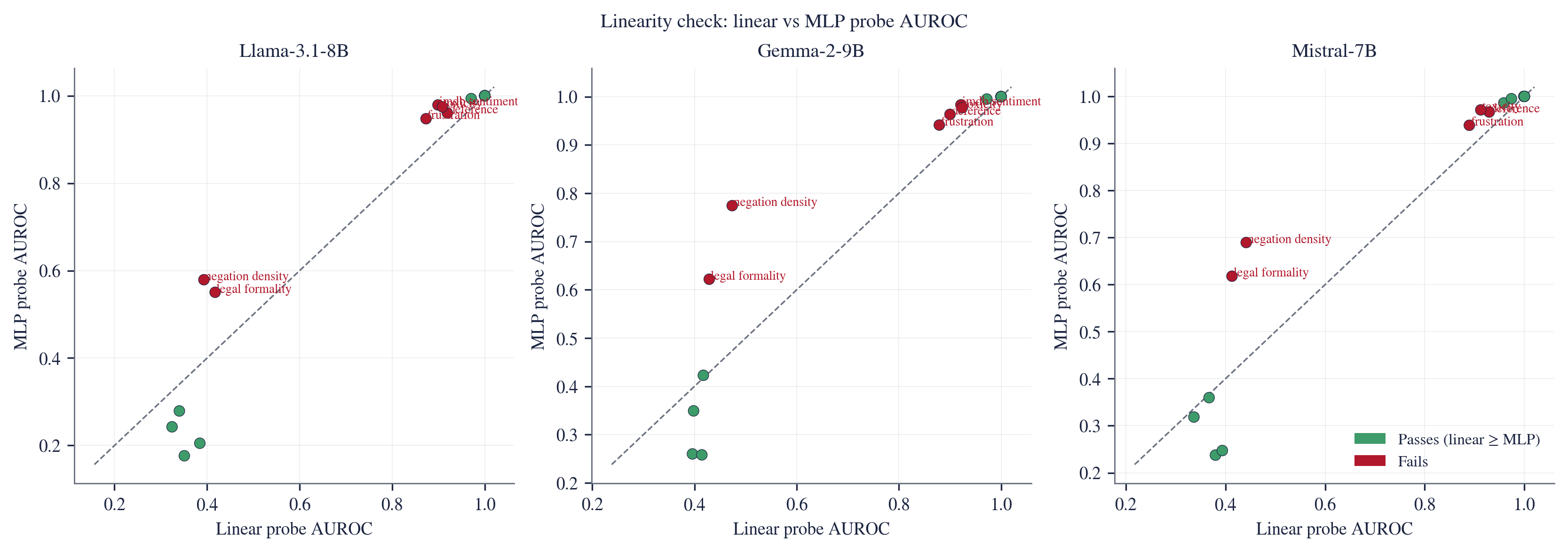}
  \caption{Three-panel linearity scatter: per-concept MLP AUROC (y-axis) vs.\ linear
  AUROC (x-axis), separately for Llama-3.1-8B (left), Gemma-2-9B (centre), and
  Mistral-7B (right).
  Points below the diagonal indicate that the linear probe \emph{outperforms} the MLP,
  consistent with clean linear separability.
  The pattern is consistent across models: easy concepts cluster near the perfect-score
  corner regardless of probe type; hard concepts cluster near the chance corner regardless
  of probe type.
  This rules out the interpretation that hard concepts are merely difficult for linear
  probes but would be detected by a more expressive model.}
  \label{fig:linearityscatter}
\end{figure}

\begin{figure}[H]
  \centering
  \includegraphics[width=0.68\linewidth]{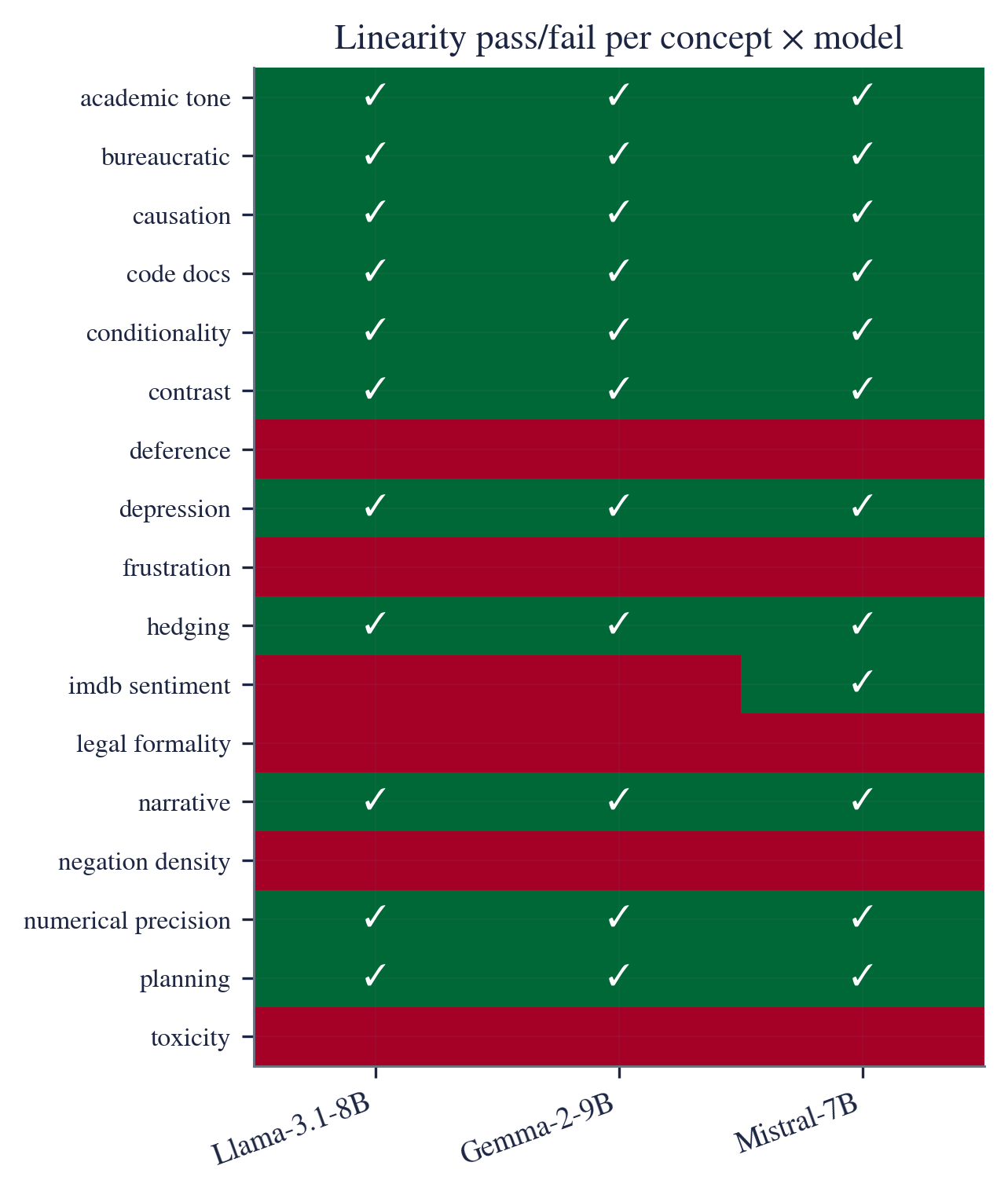}
  \caption{Linearity pass/fail heatmap.
  A cell is green (pass) if the linear probe achieves AUROC within $0.02$ of the MLP
  probe for that concept-model pair.
  Easy register-type and dense-lexical concepts pass consistently across all models;
  hard syntactic concepts fail on at least one model.
  Even for failing concepts, both probes are near chance---the non-linearity flag does
  not imply that a non-linear probe would recover high AUROC.}
  \label{fig:linearityheatmap}
\end{figure}


\section{D1 Detection: Extended Strategy Rankings}
\label{app:rankings}

\begin{figure}[H]
  \centering
  \includegraphics[width=0.72\linewidth]{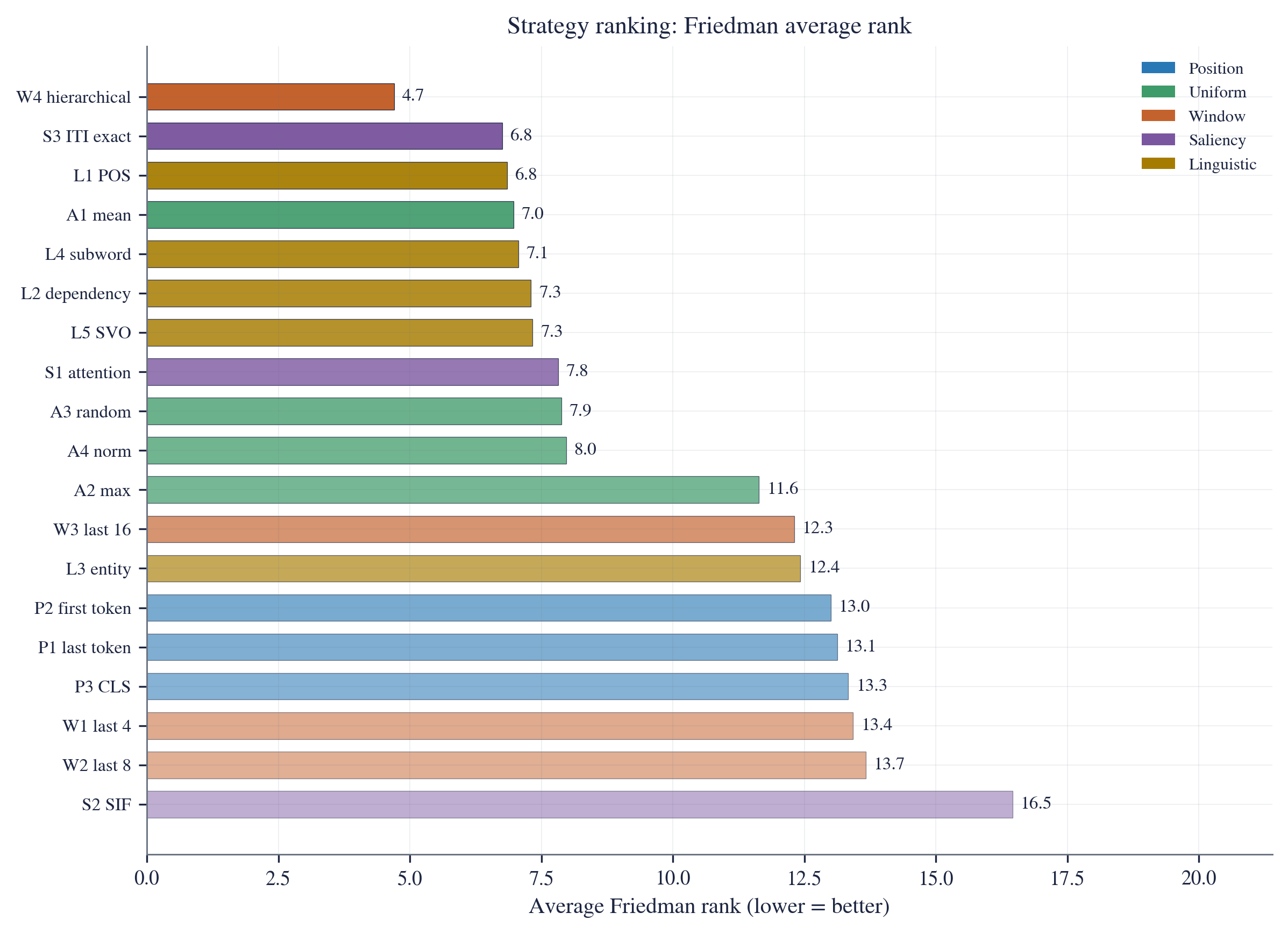}
  \caption{Average Friedman rank per strategy (lower is better).
  The top tier (\texttt{W4\_hierarchical}, \texttt{A1\_mean}, \texttt{S3\_ITI\_exact},
  \texttt{L1\_POS\_filtered}, \texttt{L4\_subword\_root}) has clearly lower ranks than
  the middle group even where AUROC differences are small.
  \texttt{S2\_SIF} and \texttt{P2\_first\_token} are statistical outliers at the bottom.}
  \label{fig:nemenyi}
\end{figure}

The main paper (Table~1) reports point-estimate D1 AUROC per strategy.
This section adds the Friedman average-rank view, which separates overall
AUROC level from rank consistency across benchmark cells.
A strategy that wins many individual cells (low average rank) is a more reliable
default than one that occasional wins at the expense of frequent near-bottom finishes.
Figure~\ref{fig:nemenyi} plots average ranks; Table~\ref{tab:rankings} records both
cross-model mean AUROC and average rank together.
The Critical Difference diagram and full pairwise $p$-value matrix are in
Appendix~\ref{app:nemenyi_stats}; D1 strategy geometry (PCA, clustering, bootstrapped
confidence intervals, and winner maps) is in Appendix~\ref{app:d1_geometry}.

\begin{table}[h]
\centering
\caption{Cross-model strategy ranking summary from the released Nemenyi analysis.
Average rank is the mean Friedman rank across all 17 concept $\times$ 3 model cells.
$\dagger$\,\texttt{L2\_dependency\_rel} is degenerate (see main paper Table~1 note)
and is included here for completeness only.}
\label{tab:rankings}
\small
\begin{tabular}{llcc}
\toprule
Strategy & Family & Cross-model mean AUROC & Average rank \\
\midrule
\texttt{W4\_hierarchical} & window & 0.7799 & 4.70 \\
\texttt{S3\_ITI\_exact} & saliency & 0.7727 & 6.76 \\
\texttt{S1\_attention\_weighted} & saliency & 0.7702 & 7.82 \\
\texttt{A1\_mean} & uniform & 0.7700 & 6.97 \\
\texttt{L2\_dependency\_rel}$^\dagger$ & linguistic & 0.7700 & 7.30 \\
\texttt{A4\_norm} & uniform & 0.7692 & 7.97 \\
\texttt{L4\_subword\_root} & linguistic & 0.7688 & 7.06 \\
\texttt{L1\_POS\_filtered} & linguistic & 0.7684 & 6.85 \\
\texttt{L5\_SVO} & linguistic & 0.7677 & 7.33 \\
\texttt{W3\_mean\_last\_16} & window & 0.7659 & 12.30 \\
\texttt{A3\_random} & uniform & 0.7649 & 7.88 \\
\texttt{P1\_last\_token} & position & 0.7640 & 13.12 \\
\texttt{W2\_mean\_last\_8} & window & 0.7605 & 13.67 \\
\texttt{W1\_mean\_last\_4} & window & 0.7597 & 13.42 \\
\texttt{A2\_max} & uniform & 0.7475 & 11.64 \\
\texttt{L3\_named\_entity} & linguistic & 0.7377 & 12.42 \\
\texttt{P3\_CLS} & position & 0.6633 & 13.33 \\
\texttt{S2\_SIF} & saliency & 0.6057 & 16.45 \\
\texttt{P2\_first\_token} & position & 0.5018 & 13.00 \\
\bottomrule
\end{tabular}
\end{table}

\section{Nemenyi Statistical Test: Full Details}
\label{app:nemenyi_stats}

\begin{figure}[H]
  \centering
  \includegraphics[width=0.80\linewidth]{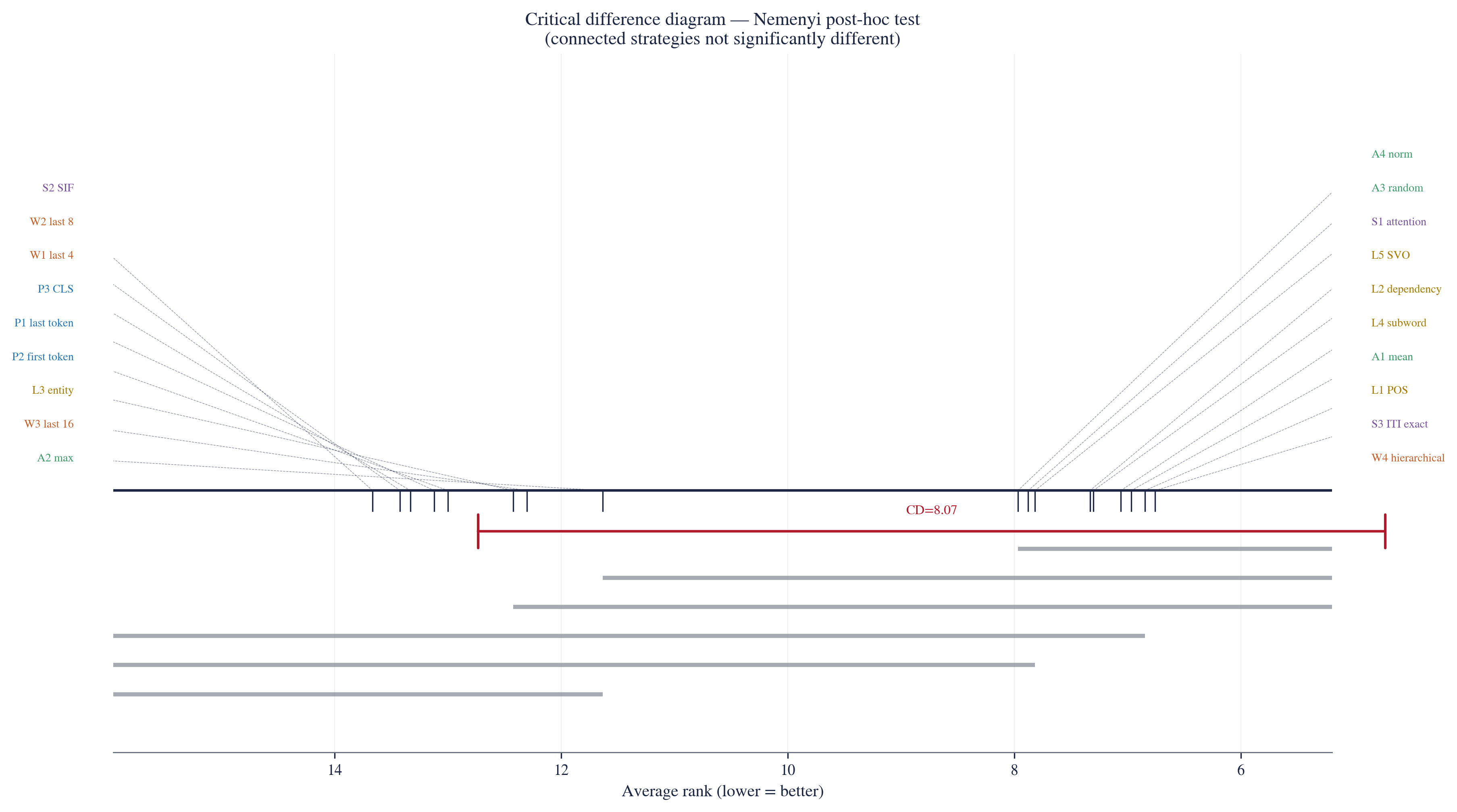}
  \caption{Critical Difference (CD) diagram (\citealt{demsar2006statistical}).
  Strategies are positioned by average Friedman rank (lower = better, plotted left).
  A horizontal bar connects strategies whose pairwise difference does not reach
  significance.
  The top tier (\texttt{W4\_hierarchical}, \texttt{A1\_mean}, \texttt{S3\_ITI\_exact},
  and the leading linguistic strategies) forms a non-significantly-separated cluster,
  but that cluster as a whole is significantly separated from the position-based defaults
  and the clear outliers (\texttt{S2\_SIF}, \texttt{P2\_first\_token}).}
  \label{fig:nemenyi_cd}
\end{figure}

\begin{figure}[H]
  \centering
  \includegraphics[width=0.78\linewidth]{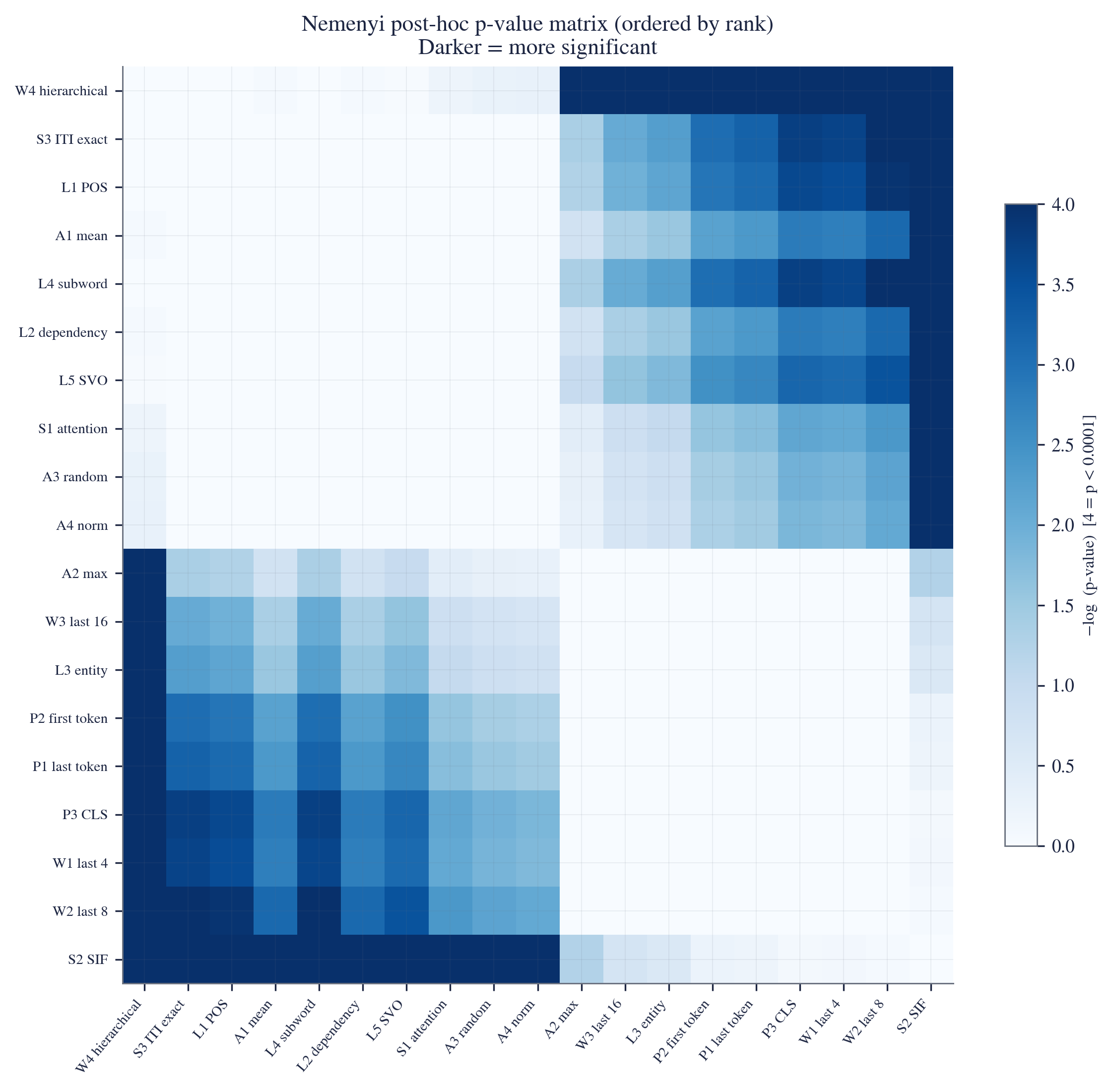}
  \caption{Full Nemenyi $p$-value matrix for all $19\times19$ strategy pairs,
  shaded by $-\log_{10}(p)$; white cells indicate $p > 0.05$.
  The top-left block (best strategies) is largely white, confirming that top-tier
  strategies are pairwise comparable.
  The \texttt{S2\_SIF} and \texttt{P2\_first\_token} rows/columns are universally
  dark: both are highly significantly worse than nearly every other strategy.
  \texttt{P1\_last\_token}'s off-diagonal entries reveal it is significantly worse
  than more strategies than its cross-model mean AUROC alone suggests.}
  \label{fig:nemenyi_pmat}
\end{figure}

\begin{figure}[H]
  \centering
  \includegraphics[width=0.65\linewidth]{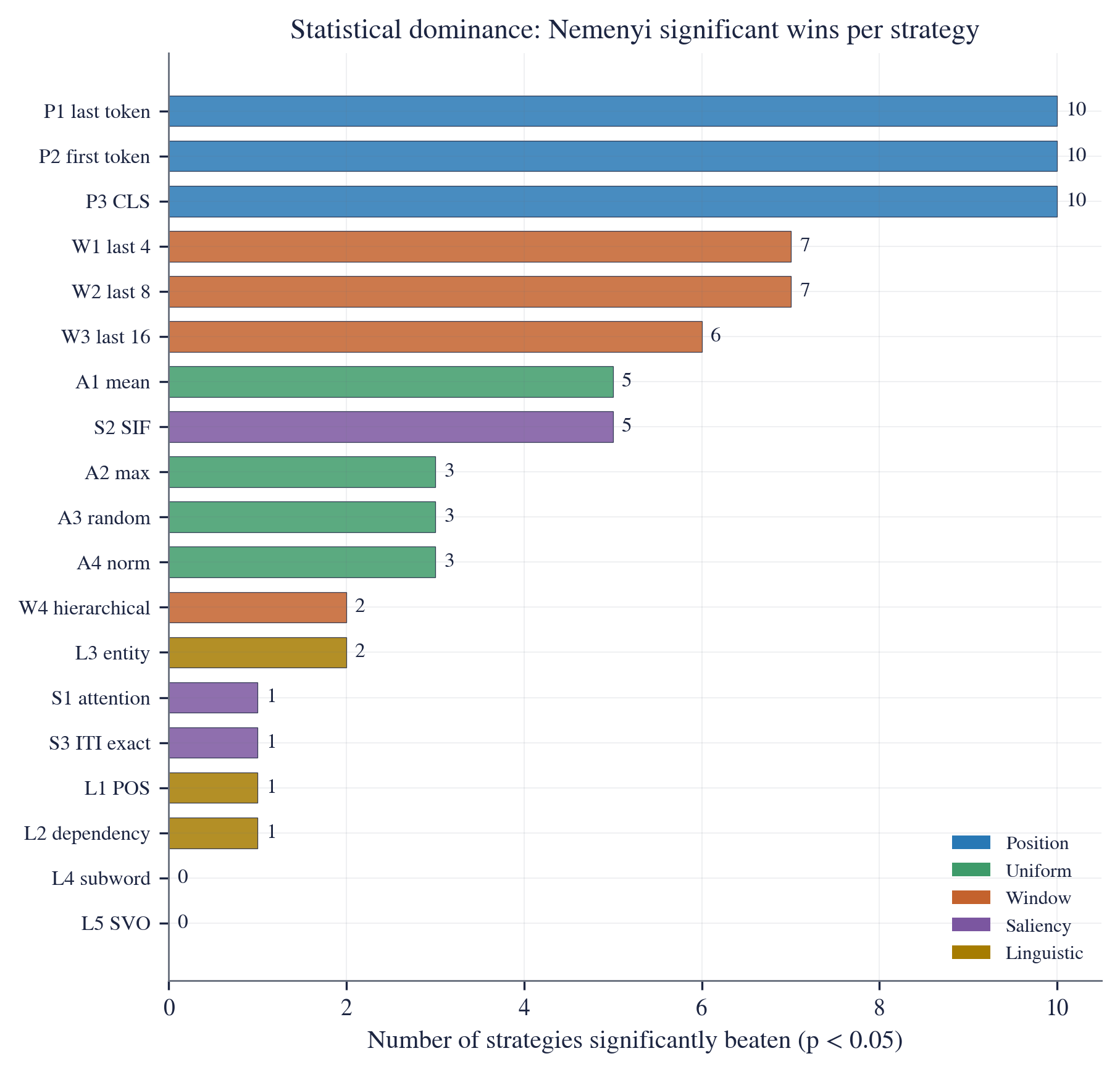}
  \caption{Dominance count: number of other strategies each strategy significantly
  outperforms under Nemenyi ($p < 0.05$), sorted descending.
  \texttt{W4\_hierarchical} dominates the most strategies.
  \texttt{P2\_first\_token} and \texttt{S2\_SIF} dominate zero strategies.}
  \label{fig:nemenyi_dom}
\end{figure}

The Friedman test (Section~3 of the main paper) uses the 11~linear-passing concepts
as benchmark cells, giving an effective $N$ of 12.106 after the Iman--Davenport
correction for ties.
The Nemenyi post-hoc test evaluates all ${19 \choose 2} = 171$ strategy pairs at
$\alpha = 0.05$; 77 pairs reach significance.
Figures~\ref{fig:nemenyi_cd}--\ref{fig:nemenyi_dom} provide the three standard
Nemenyi visualisations: the Critical Difference diagram, the full $p$-value matrix,
and the dominance count.

\section{Strategy Stability Across Models}
\label{app:stability}

\begin{figure}[H]
  \centering
  \includegraphics[width=0.72\linewidth]{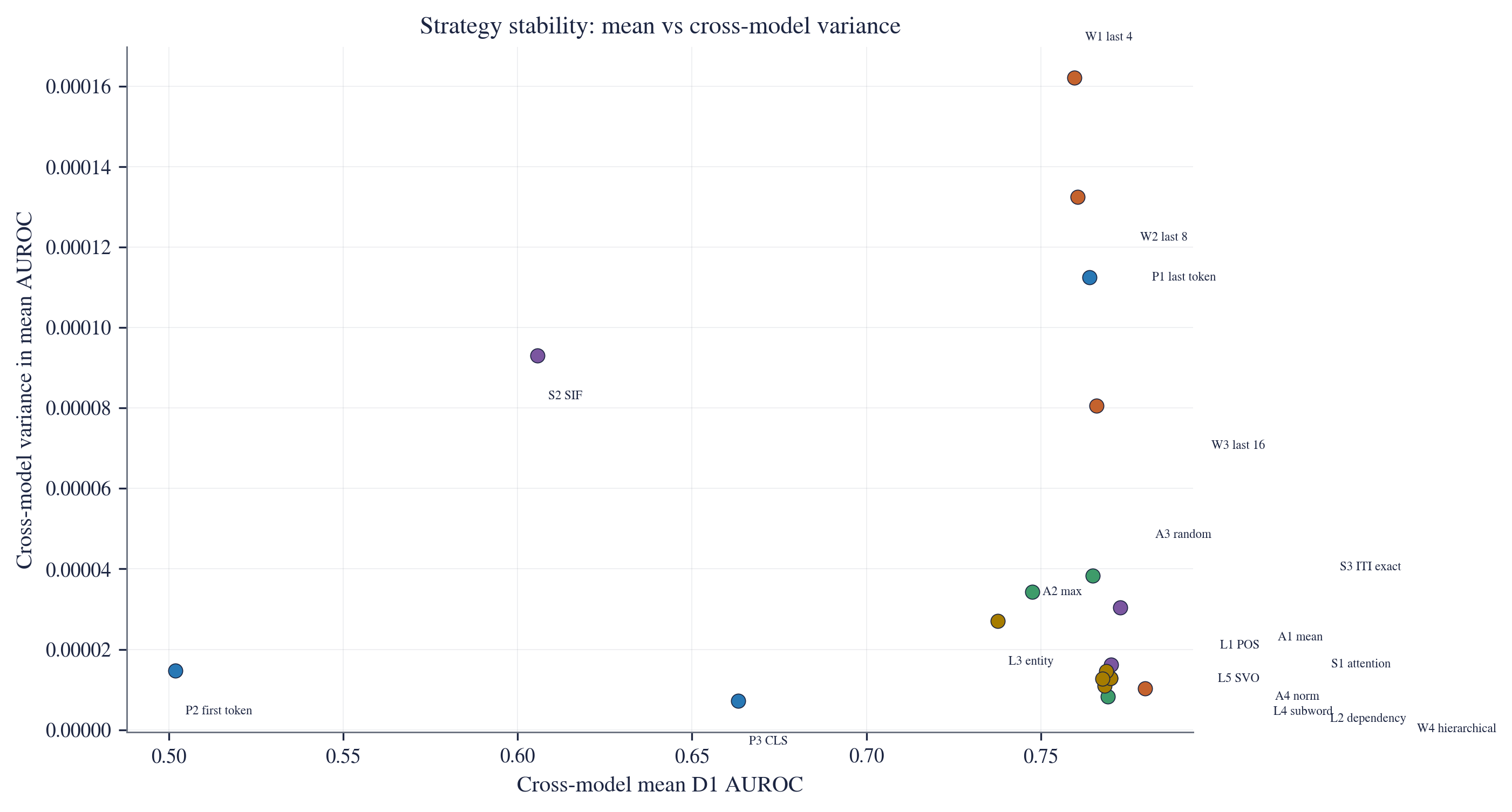}
  \caption{Strategy stability: cross-model mean AUROC vs.\ cross-model variance.
  Strategies in the upper-left quadrant are high-performing and consistent.
  The top-tier strategies cluster at low variance, confirming that their advantage is
  not architecture-specific.
  \texttt{W3\_mean\_last\_16} and \texttt{P1\_last\_token} achieve competitive means
  but with noticeably higher variance, consistent with their poor average rank despite
  their acceptable cross-model mean AUROC.}
  \label{fig:stability}
\end{figure}

A strategy that achieves a high cross-model mean AUROC but with high cross-model
variance is architecture-sensitive and risky as a default.
Figure~\ref{fig:stability} plots mean AUROC against variance for each strategy, making
this tradeoff explicit.
The fallback behaviour of linguistic strategies---which affects how distinctly they
differ from mean pooling---is documented in Appendix~\ref{app:fallback}.

\section{D1 Strategy Geometry and Confidence Intervals}
\label{app:d1_geometry}

The main leaderboard summarises D1 AUROC as point estimates.
This section provides six geometric views that reveal structure hidden behind those
estimates: family-level distributions, bootstrapped confidence intervals, concept-model
winner maps, and low-dimensional projections of the full strategy space.
Together these views confirm the main paper's claim that the top-tier strategies are
robust, that the hard-concept cells are genuinely unresolved, and that the worst
strategies occupy a distinct corner of strategy space.

\begin{figure}[H]
  \centering
  \includegraphics[width=0.78\linewidth]{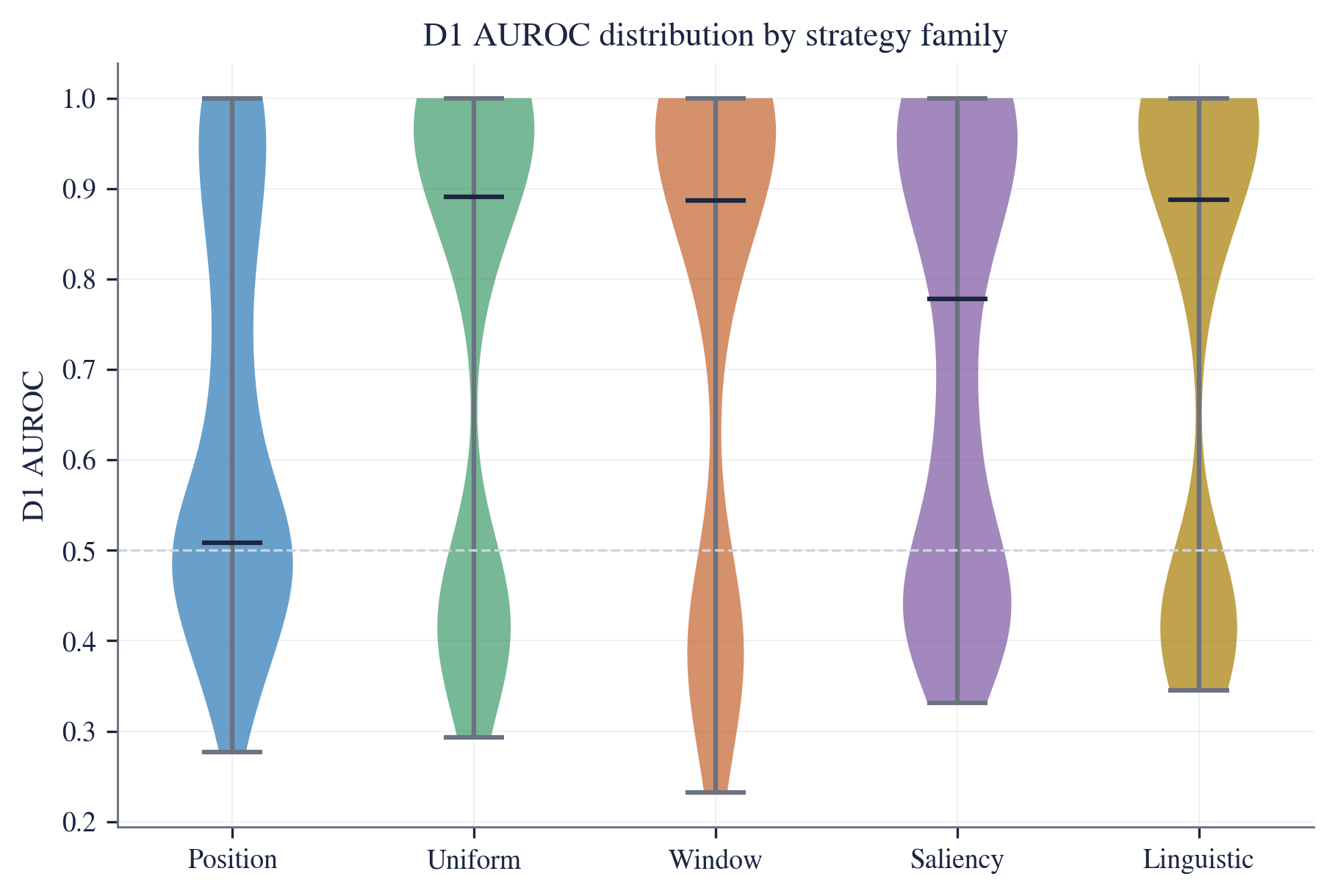}
  \caption{D1 AUROC distribution by strategy family (violin plot, cross-model and
  cross-concept).
  The window family achieves the highest and most consistent median.
  Position strategies are bimodal: \texttt{P1} and \texttt{P3} sit in the mid-range
  while \texttt{P2\_first\_token} collapses to near-chance.
  Uniform strategies show the tightest inter-quartile range, making them the most
  predictable default.
  Linguistic strategies span a wide range that mirrors the variation in parser coverage
  across concepts.}
  \label{fig:d1violin}
\end{figure}

\begin{figure}[H]
  \centering
  \includegraphics[width=0.72\linewidth]{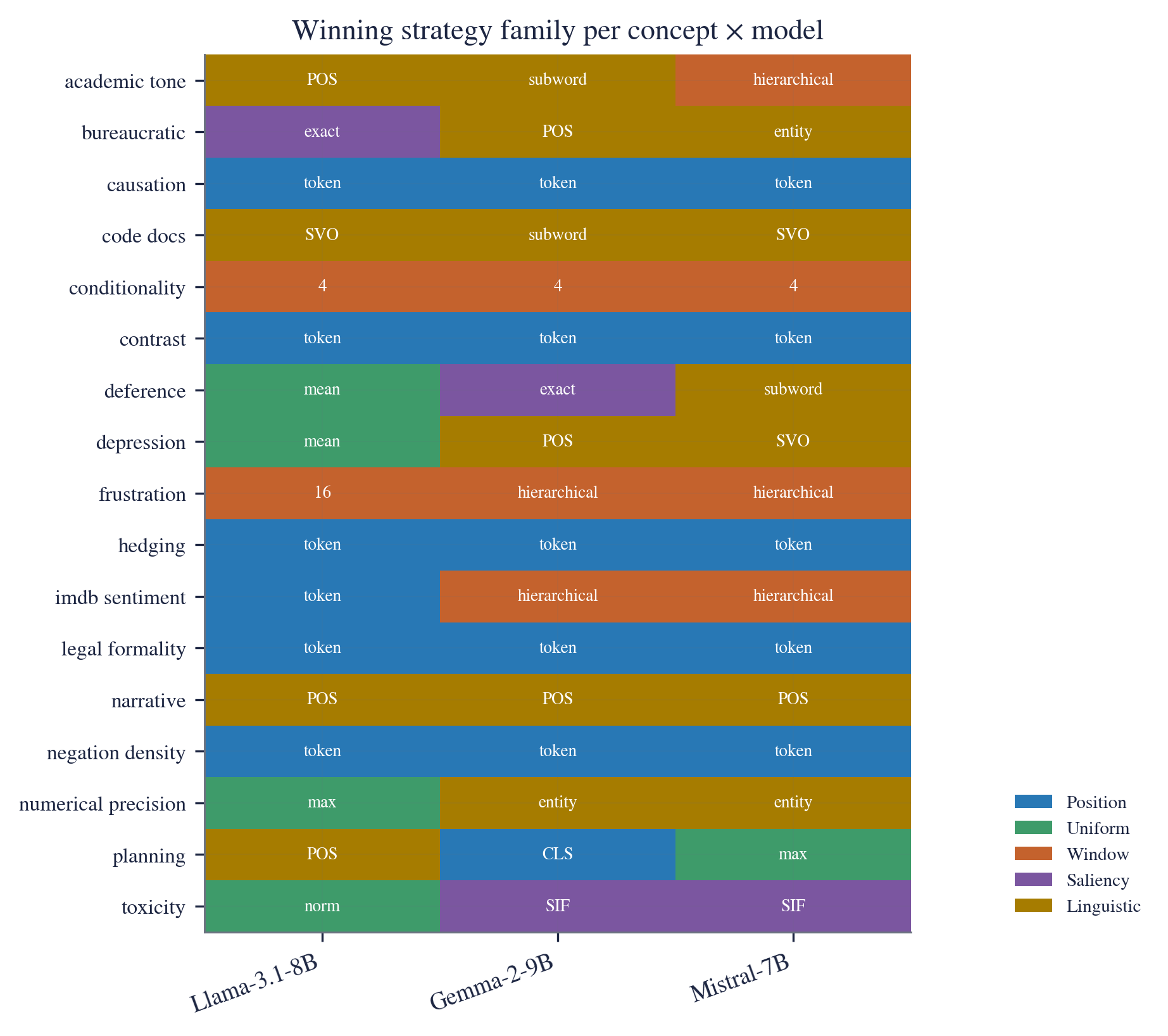}
  \caption{Winner map: which strategy achieves the highest D1 AUROC in each
  concept$\times$model cell?
  \texttt{W4\_hierarchical} wins the majority of cells, particularly on easy
  register-type and dense-lexical concepts.
  The hard syntactic concepts display less consistent winners across models,
  consistent with near-chance AUROC where small sampling fluctuations determine
  the apparent top strategy---confirming that the benchmark has a robust winner in
  the tractable regime and a genuinely unresolved regime at the hard extreme.}
  \label{fig:d1winner}
\end{figure}

\begin{figure}[H]
  \centering
  \includegraphics[width=\linewidth]{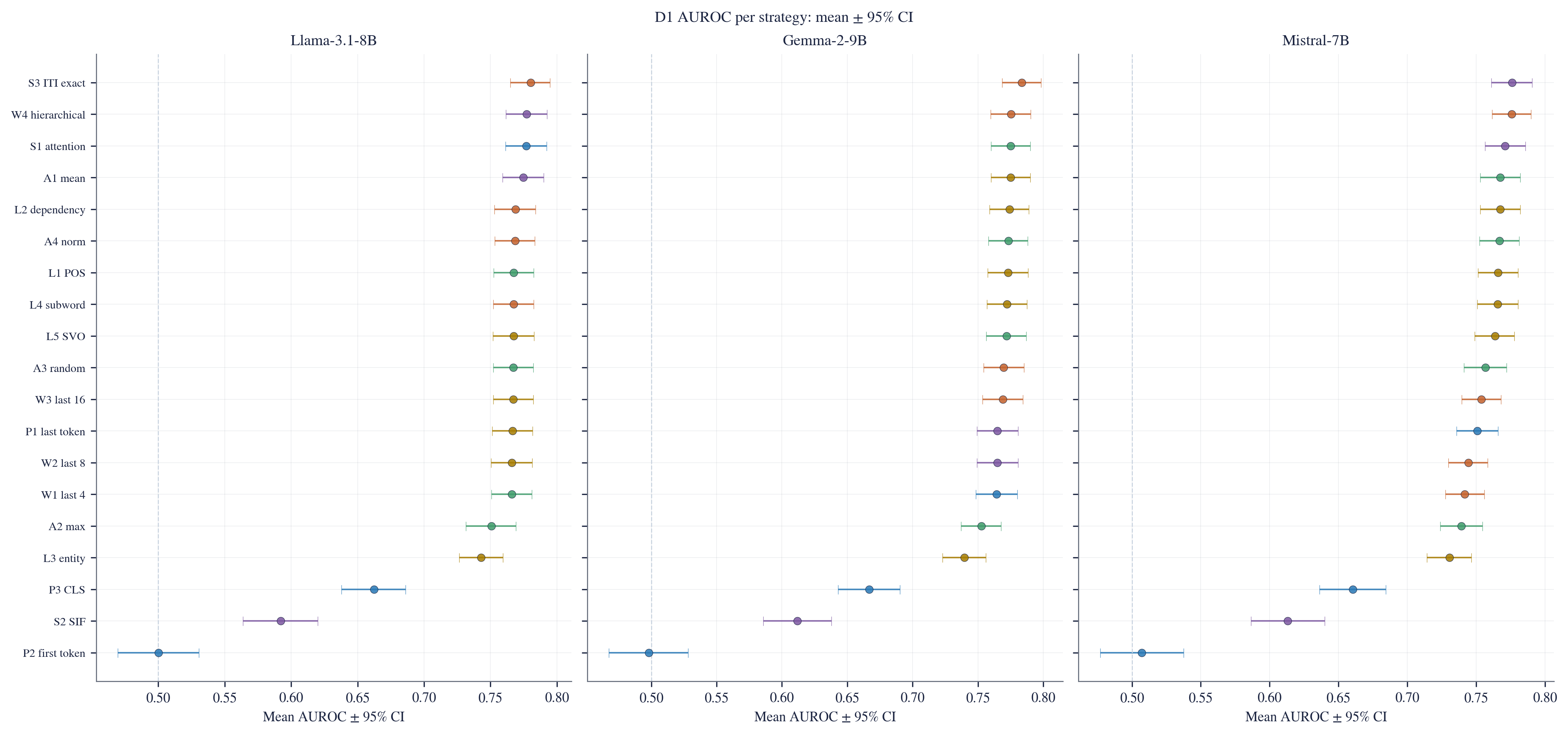}
  \caption{D1 AUROC with 95\% bootstrap confidence intervals per strategy, shown
  separately per model (left: Llama-3.1-8B; centre: Gemma-2-9B; right: Mistral-7B).
  Strategies are sorted by cross-model mean AUROC.
  Within the tractable regime (AUROC $> 0.70$), the top-tier strategies remain
  distinguishable from the position-based defaults even after accounting for
  confidence intervals.
  The Mistral panel shows the widest CIs, consistent with its more uneven
  per-concept difficulty profile.}
  \label{fig:d1ci3panel}
\end{figure}

\begin{figure}[H]
  \centering
  \includegraphics[width=0.72\linewidth]{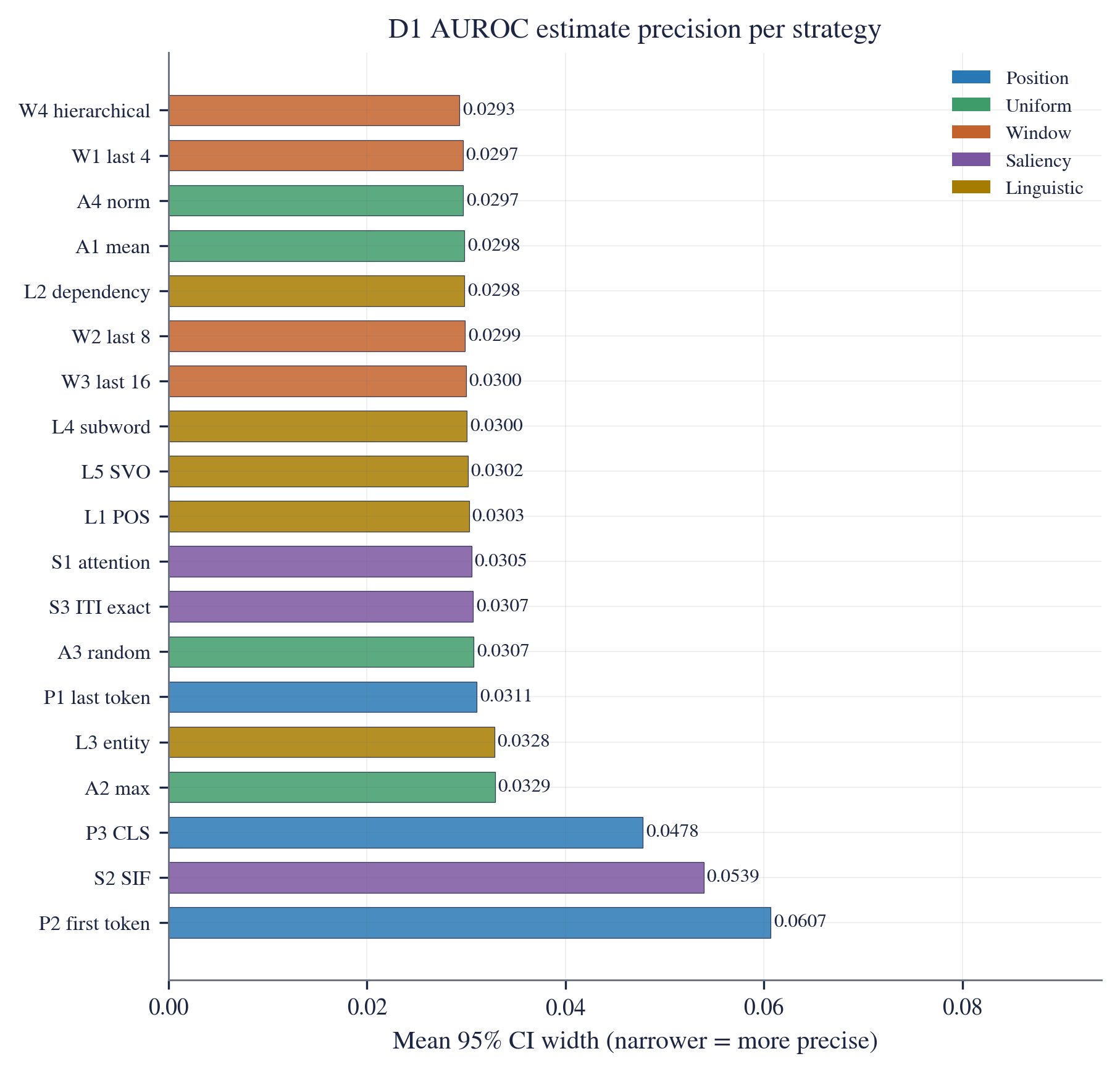}
  \caption{Mean CI half-width per strategy, averaged across all 17 concepts and
  3 models.
  Uniform strategies (\texttt{A1}, \texttt{A3}, \texttt{A4}) exhibit the narrowest
  CIs, consistent with stable mean pooling even in difficult regimes.
  \texttt{P2\_first\_token} has anomalously wide CIs relative to its rank, reflecting
  high bootstrap variance at near-chance AUROC regardless of concept.}
  \label{fig:d1ciwidth}
\end{figure}

\begin{figure}[H]
  \centering
  \includegraphics[width=0.72\linewidth]{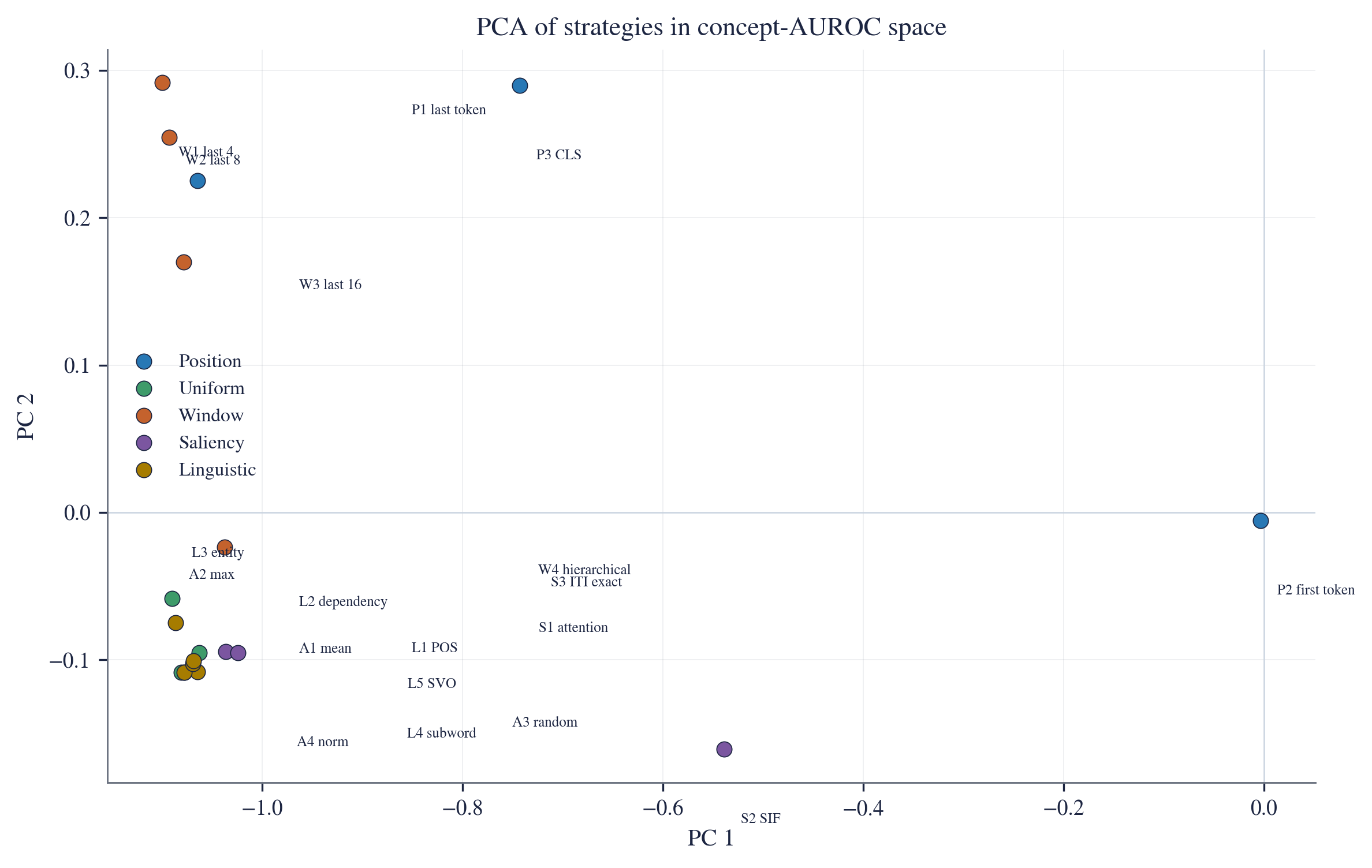}
  \caption{PCA of the $19 \times (17 \times 3)$ D1 AUROC matrix.
  Strategy families separate into recognisable clusters: position strategies occupy
  an isolated corner; uniform and most linguistic strategies overlap in the central
  cluster; \texttt{W4\_hierarchical} lies at the extremity of the window cluster.
  \texttt{S2\_SIF} and \texttt{P2\_first\_token} are clear outliers in the direction
  of lowest overall AUROC.}
  \label{fig:d1pca}
\end{figure}

\begin{figure}[H]
  \centering
  \includegraphics[width=0.78\linewidth]{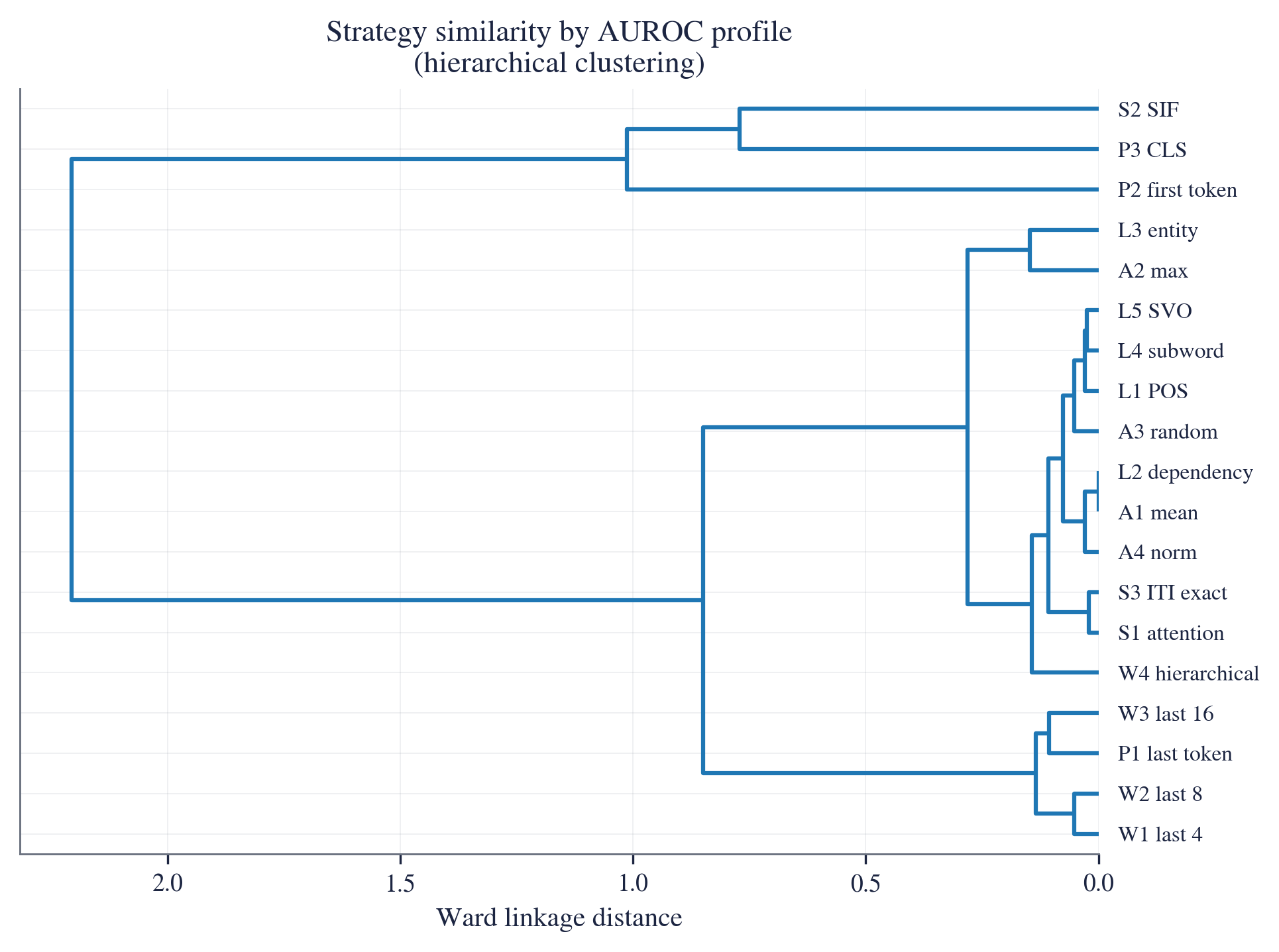}
  \caption{Hierarchical clustering dendrogram of strategies based on their
  cross-concept-model D1 AUROC profiles (Ward linkage, Euclidean distance).
  \texttt{W4\_hierarchical} clusters most closely with the uniform averaging
  strategies (\texttt{A1\_mean}, \texttt{A4\_norm}) rather than with the shorter
  window strategies (\texttt{W1}, \texttt{W2}, \texttt{W3}), suggesting that
  hierarchical pooling produces representations more similar to broad-coverage
  averaging than to narrow trailing-window approaches.
  \texttt{P2\_first\_token} and \texttt{S2\_SIF} form their own cluster, consistent
  with being representationally distinct failures.}
  \label{fig:d1dendro}
\end{figure}

\section{Linguistic Strategy Fallback Rates}
\label{app:fallback}

\begin{figure}[H]
  \centering
  \includegraphics[width=0.72\linewidth]{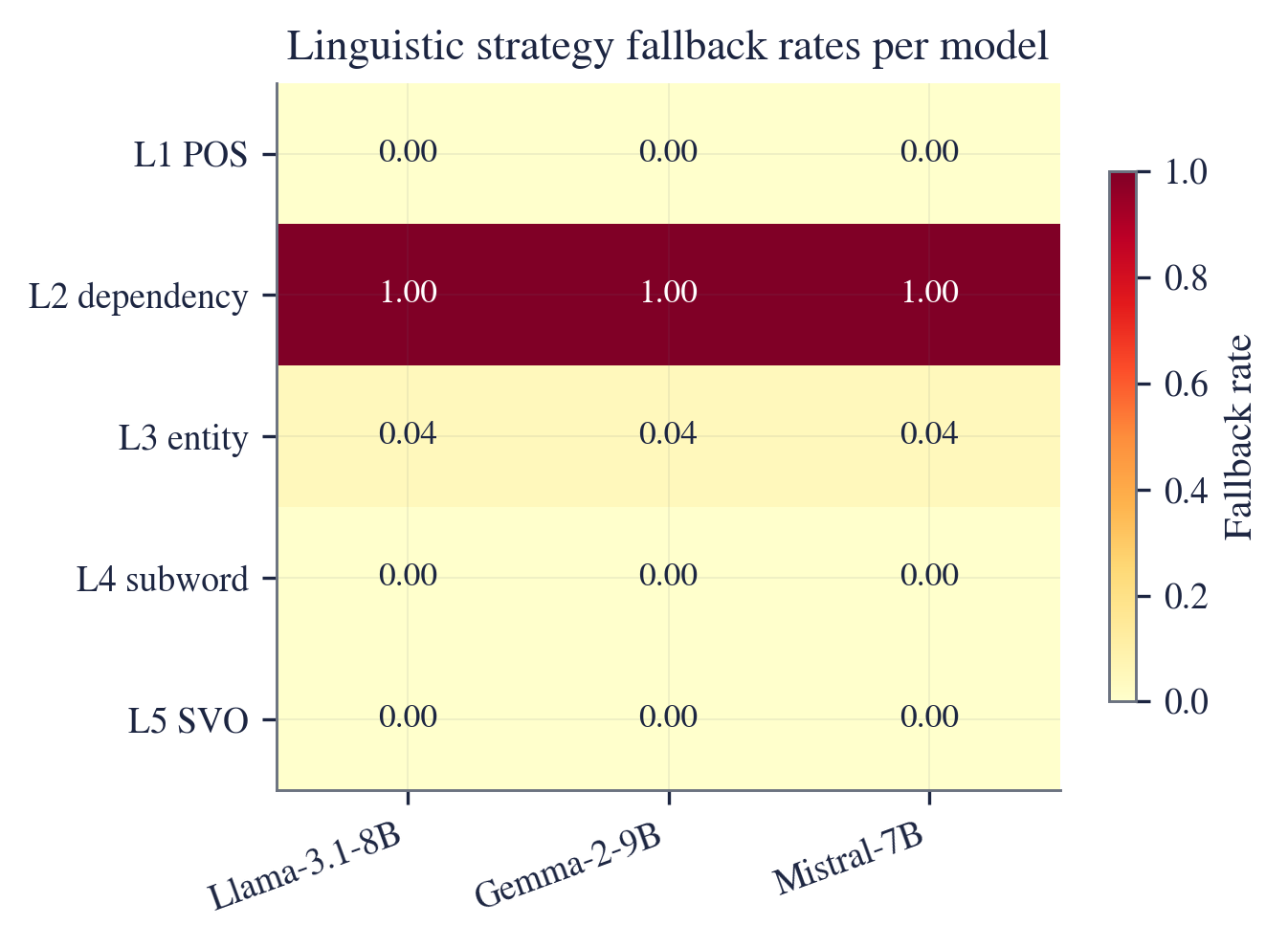}
  \caption{Fallback rates for linguistic strategies L1--L5, per model (darker =
  higher fallback rate).
  \texttt{L2\_dependency\_rel} falls back to mean pooling on 100\% of passages across
  all three models due to a parser implementation issue (zero dependency arcs matched),
  making its behaviour identical to \texttt{A1\_mean}.
  \texttt{L3\_named\_entity} has a 4.36\% fallback rate for passages with no named
  entities.
  L1, L4, and L5 have negligible fallback ($\leq 0.02\%$) and behave as genuinely
  distinct strategies.}
  \label{fig:fallback}
\end{figure}

Linguistic strategies (L1--L5) apply a part-of-speech or dependency parser filter
to select specific token positions.
When the filter matches no tokens in a passage, the strategy falls back to mean
pooling over all tokens---making it equivalent to \texttt{A1\_mean} for that passage.
A strategy with a high fallback rate is not a genuinely distinct pooling choice and
its D1 score cannot be attributed to the linguistic selection criterion.
Figure~\ref{fig:fallback} shows the fallback rates; the key finding is the 100\%
fallback for \texttt{L2\_dependency\_rel} explained in the main paper.

\section{Concept--Strategy Detection Heatmaps}
\label{app:heatmaps}

\begin{figure}[H]
  \centering
  \includegraphics[width=\linewidth]{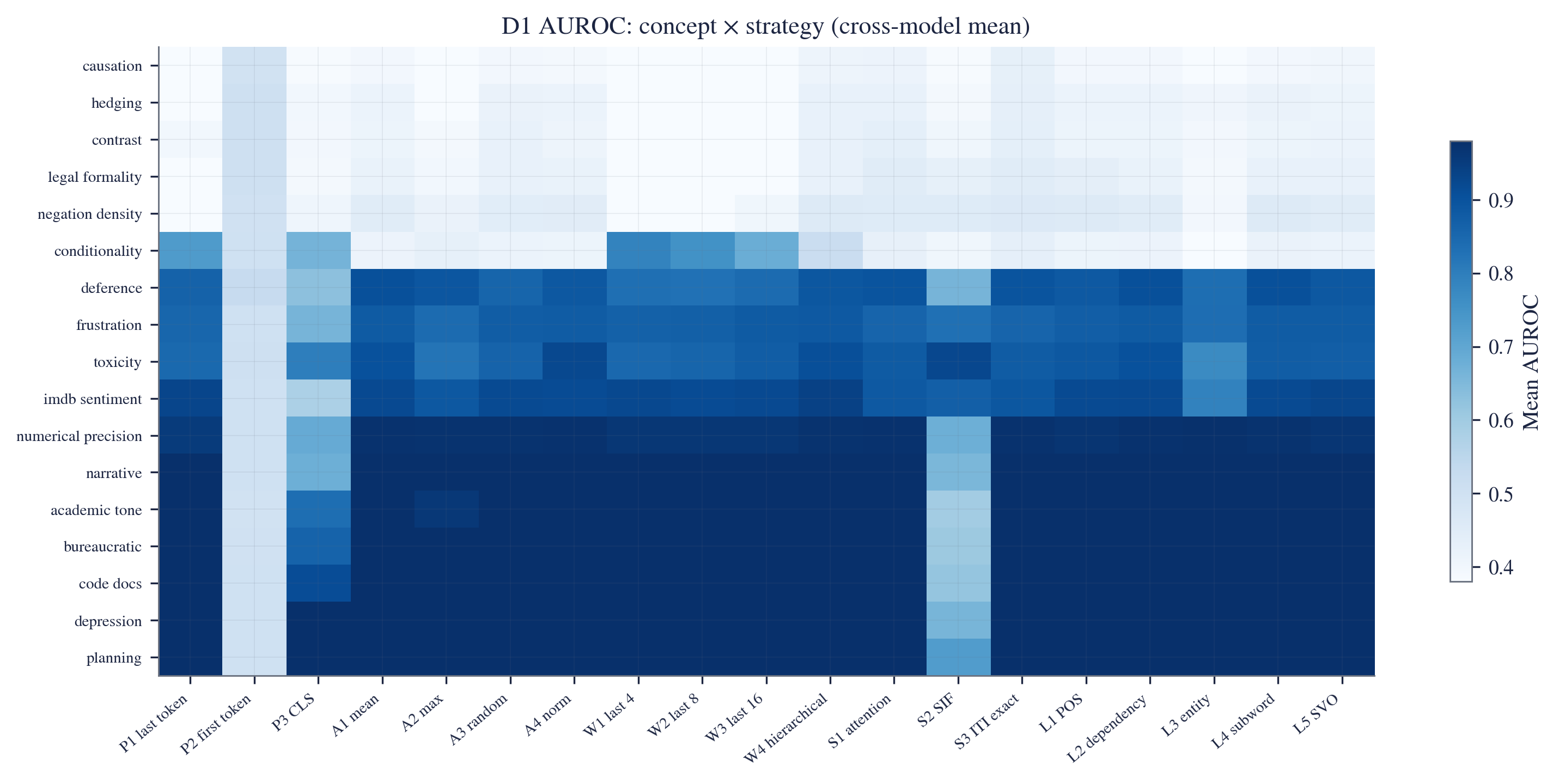}
  \caption{Full concept $\times$ strategy D1 AUROC heatmap (cross-model mean).
  Concepts are sorted by mean AUROC from bottom (easiest) to top (hardest).}
  \label{fig:heatmap_full}
\end{figure}

\begin{figure}[H]
  \centering
  \includegraphics[width=\linewidth]{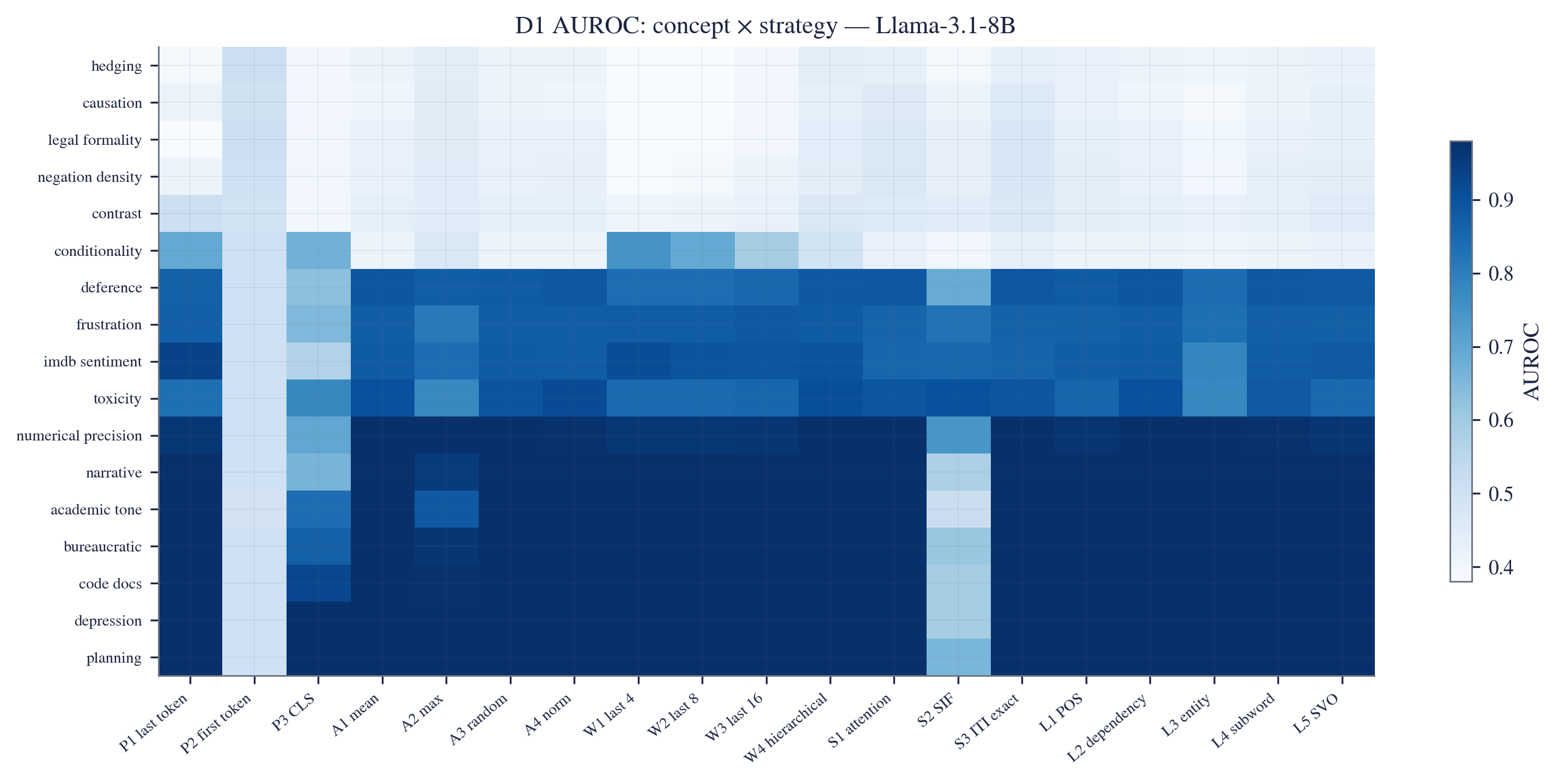}
  \caption{Concept $\times$ strategy D1 AUROC heatmap --- Llama-3.1-8B.}
  \label{fig:heatmap_llama}
\end{figure}

\begin{figure}[H]
  \centering
  \includegraphics[width=\linewidth]{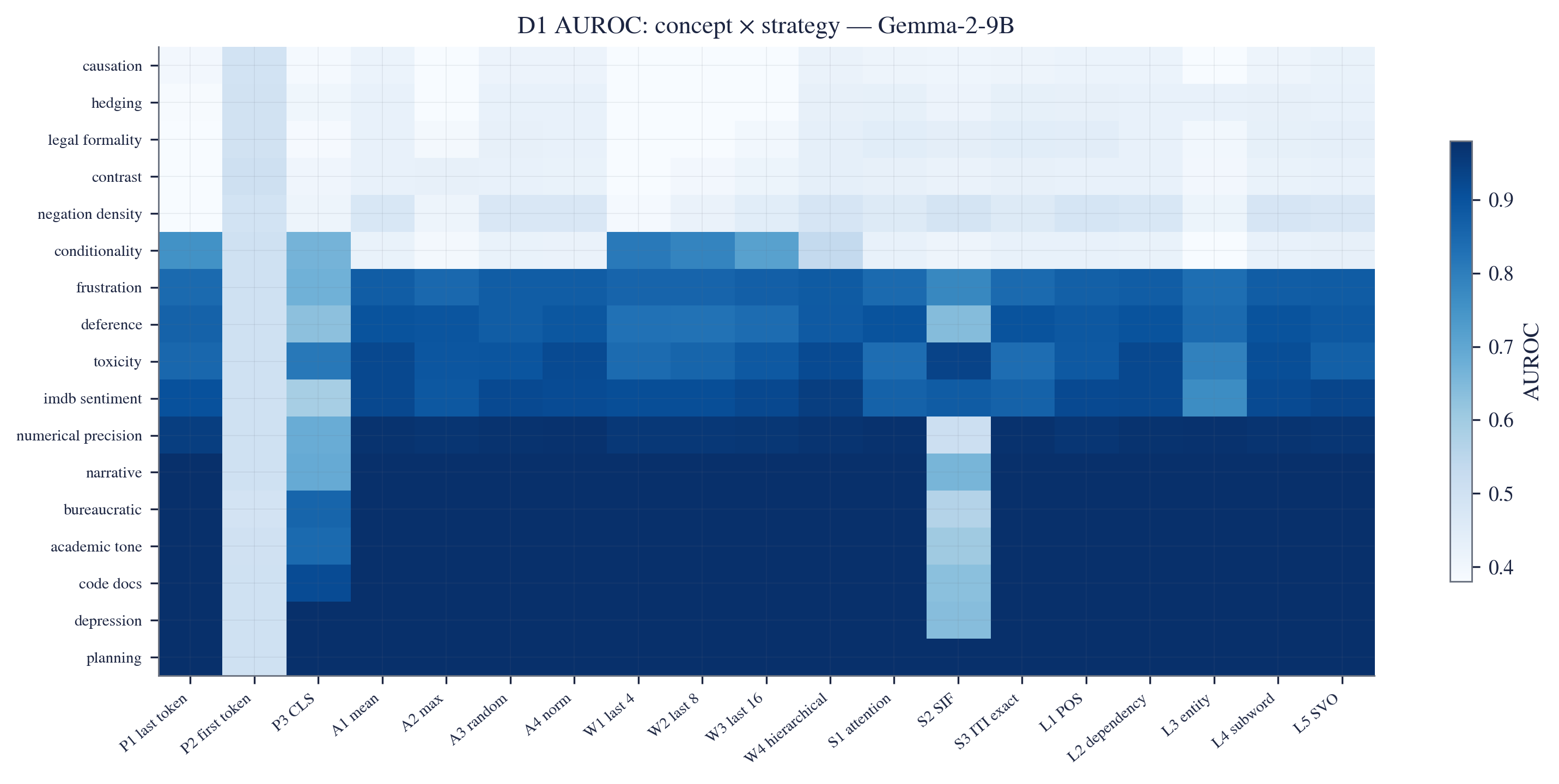}
  \caption{Concept $\times$ strategy D1 AUROC heatmap --- Gemma-2-9B.}
  \label{fig:heatmap_gemma}
\end{figure}

\begin{figure}[H]
  \centering
  \includegraphics[width=\linewidth]{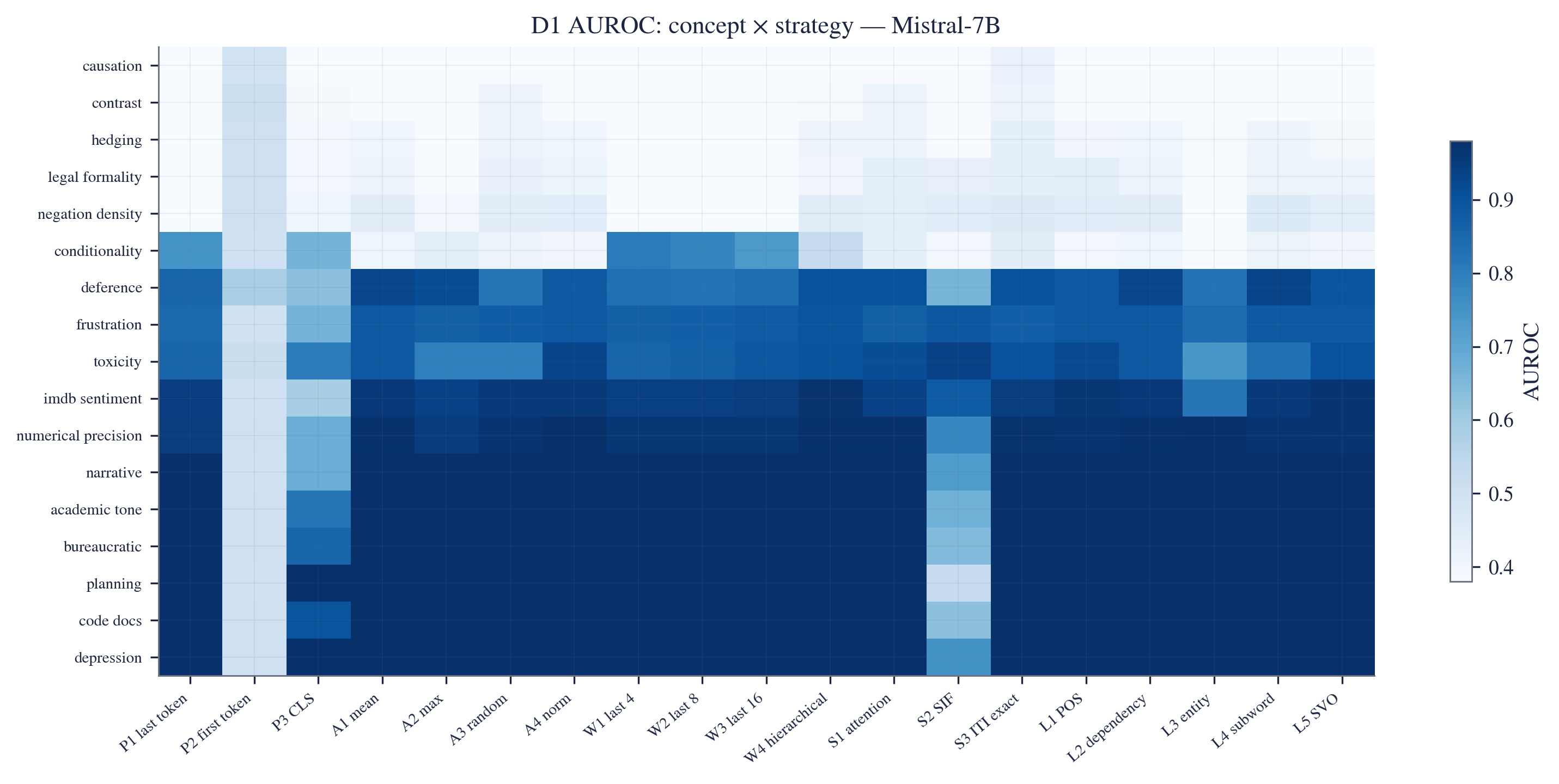}
  \caption{Concept $\times$ strategy D1 AUROC heatmap --- Mistral-7B.}
  \label{fig:heatmap_mistral}
\end{figure}

\begin{figure}[H]
  \centering
  \includegraphics[width=0.60\linewidth]{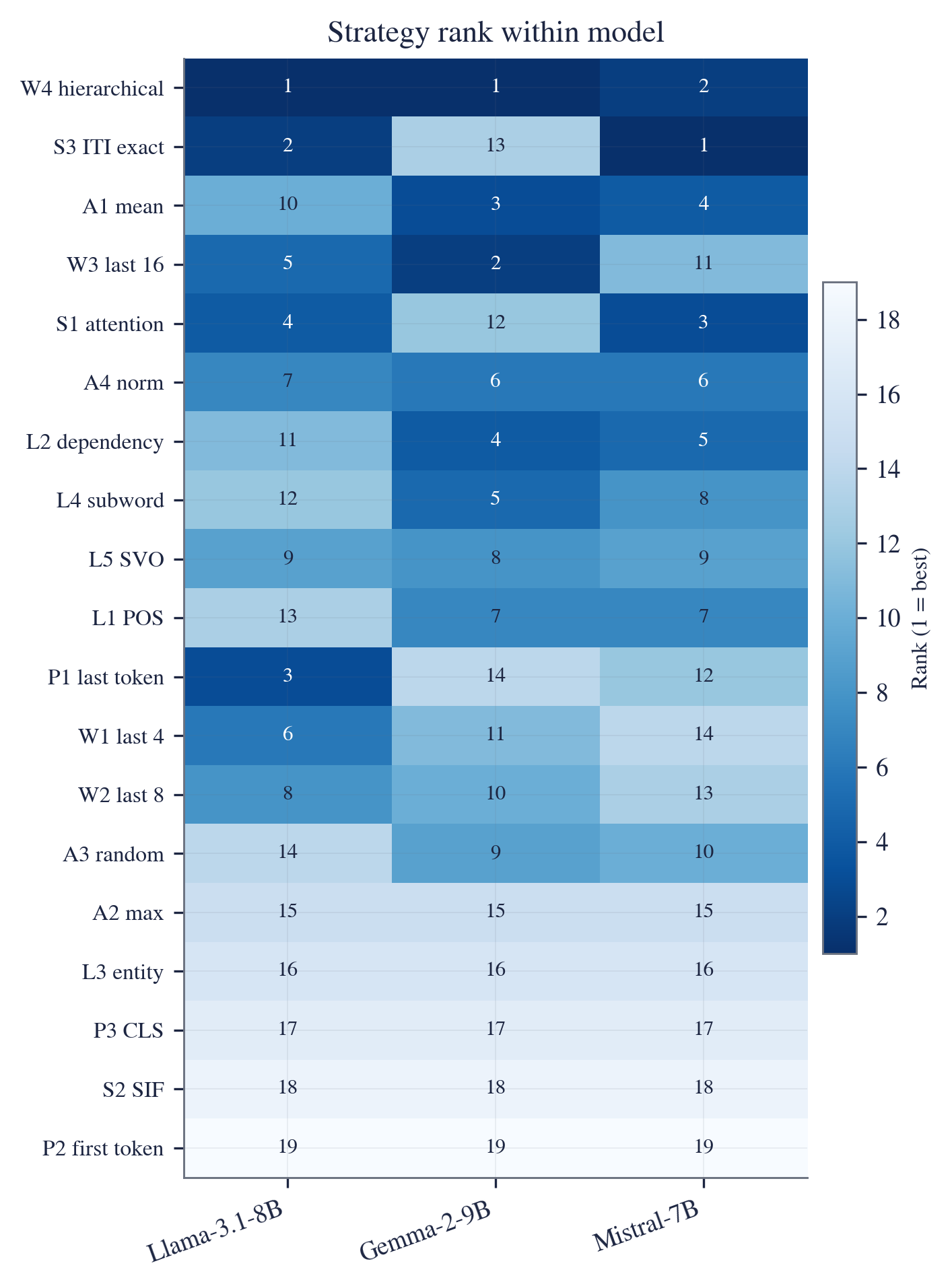}
  \caption{Within-model strategy rank matrix (1 = best per model), sorted by cross-model
  mean rank.
  The rank view surfaces the \texttt{P1\_last\_token} drop on Mistral more clearly
  than the AUROC matrix.}
  \label{fig:rankmatrix}
\end{figure}

Figure~\ref{fig:heatmap_full} shows the full cross-model mean D1 AUROC for every
concept--strategy pair as a colour-coded matrix.
The top-performing strategies are distinguishable by their uniformly darker columns
in the low-AUROC (hard-concept) rows.
Figures~\ref{fig:heatmap_llama}--\ref{fig:heatmap_mistral} provide the model-specific
versions; Figure~\ref{fig:rankmatrix} shows the rank-based variant.

\section{Inter-Rater Consistency: ICC Analysis}
\label{app:icc}

The benchmark's ICC(2,1) reliability metric (Section~3 of the main paper) measures
how consistently the 5-fold probing protocol would rank two independently drawn sets
of passages for the same concept.
The per-model mean ICC values (Llama-3.1-8B: 0.8773; Gemma-2-9B: 0.8586; Mistral-7B: 0.9217) aggregate over all 17~concepts;
here we disaggregate by concept to identify which concepts and models drive the
reliability ceiling.
A concept with consistently low ICC across all models reflects genuine representational
instability rather than an idiosyncratic probe artefact.

\begin{figure}[H]
  \centering
  \includegraphics[width=0.72\linewidth]{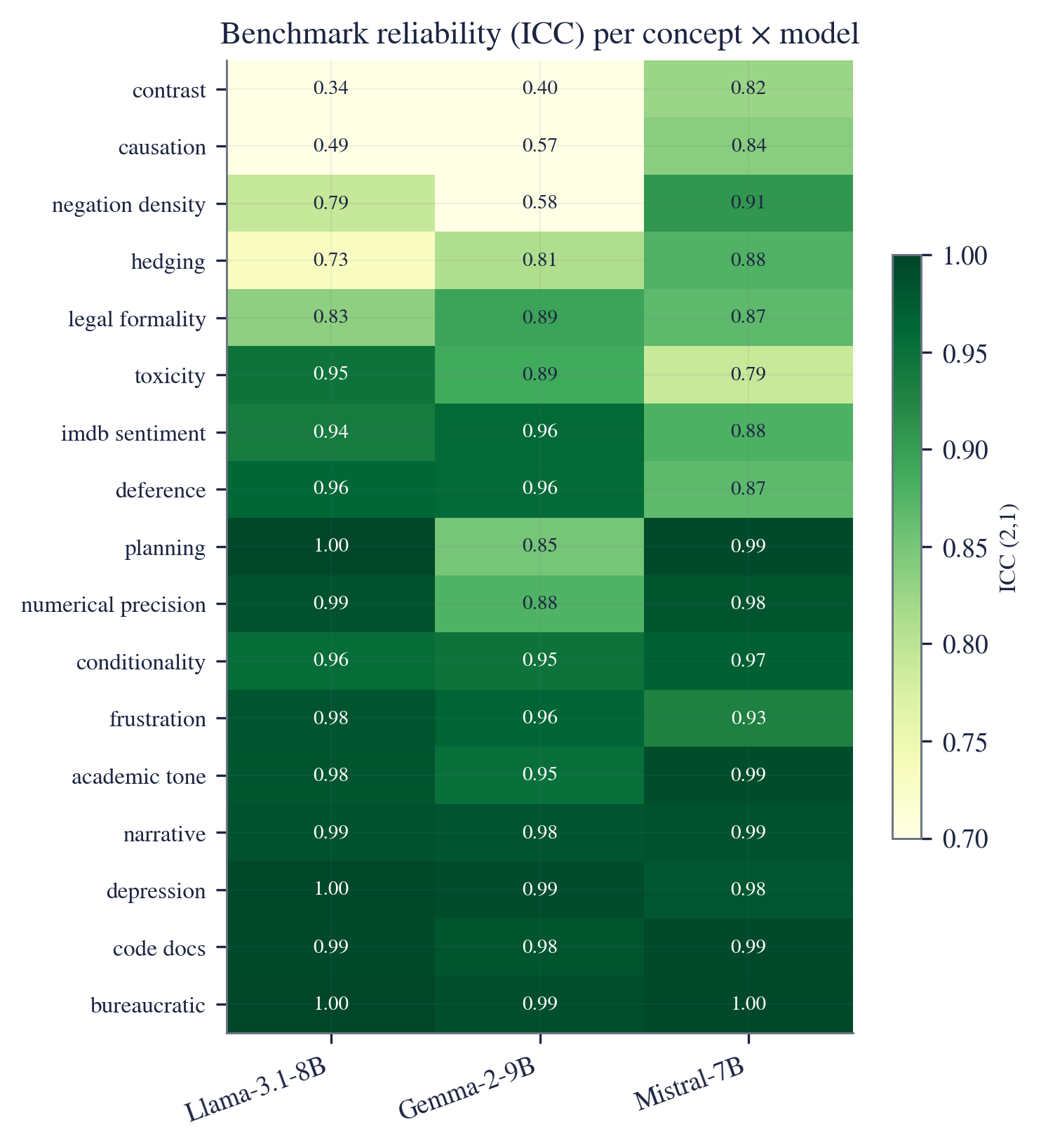}
  \caption{Per-concept ICC heatmap.
  Rows are concepts sorted by cross-model mean ICC (lowest at top); columns are the
  three benchmark models.
  \texttt{contrast} and \texttt{causation} show the weakest reliability on Llama and
  Gemma, consistent with their near-chance D1 AUROC.
  \texttt{toxicity} is the weakest concept on Mistral but not on the other two models,
  pointing to a Mistral-specific representation difference.
  The overall pattern confirms that the benchmark's reliability bottleneck is the set of
  hard syntactic concepts, not the evaluation protocol itself.}
  \label{fig:iccheatmap}
\end{figure}

\begin{figure}[H]
  \centering
  \includegraphics[width=0.60\linewidth]{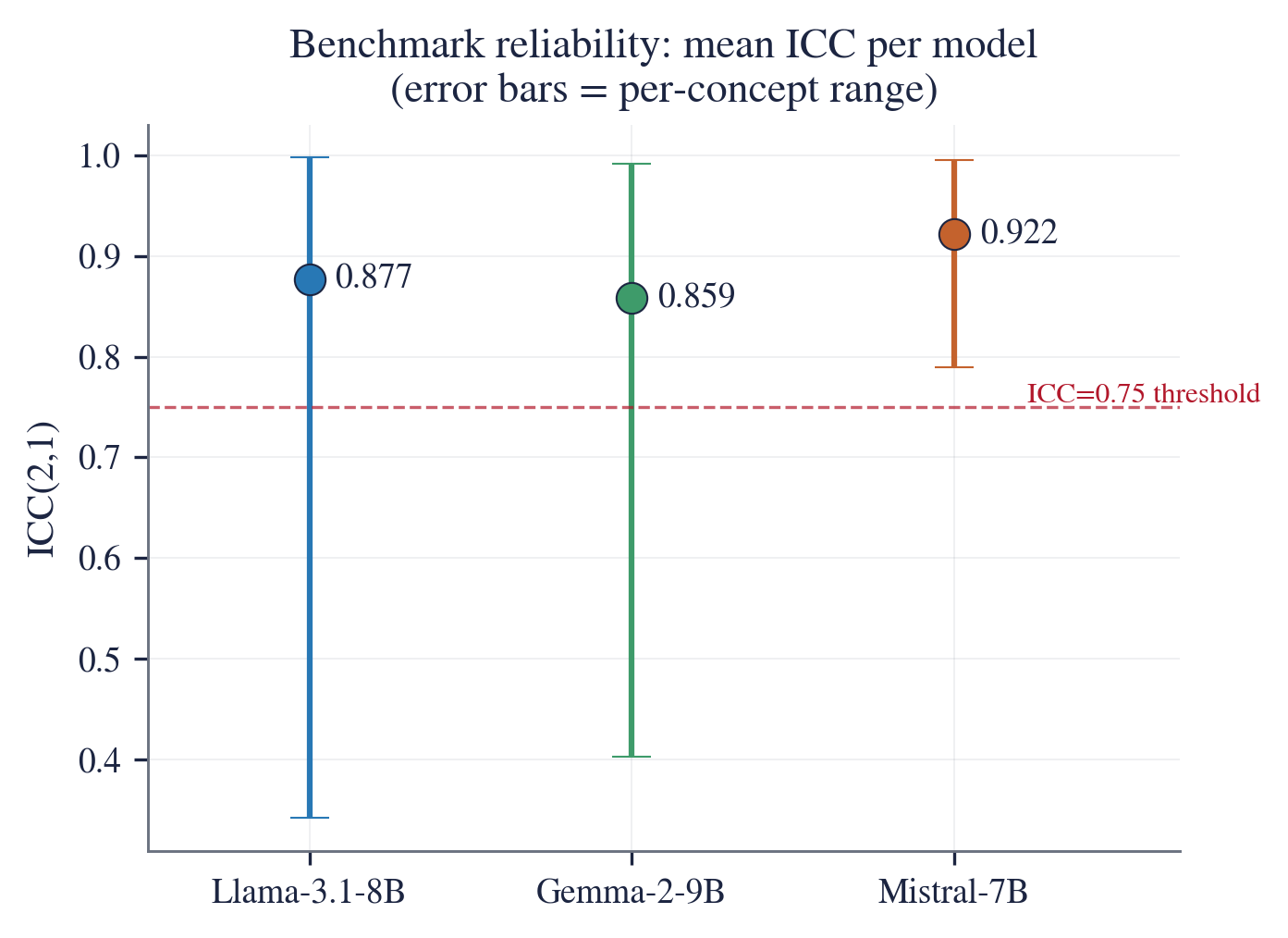}
  \caption{ICC summary dot-plot: mean ICC per model (filled dot) with the
  min--max range across the 17 released concepts (error bar).
  Mistral-7B achieves the highest mean ICC (0.9217) and the tightest spread.
  Gemma-2-9B has the lowest mean (0.8586) and widest spread.
  Even the lowest displayed value remains well above the acceptable threshold
  ($\mathrm{ICC} \geq 0.75$; \citealt{koo2016icc}).}
  \label{fig:iccsummary}
\end{figure}


\section{Construction Method Comparison}
\label{app:construction}

\begin{figure}[H]
  \centering
  \includegraphics[width=0.68\linewidth]{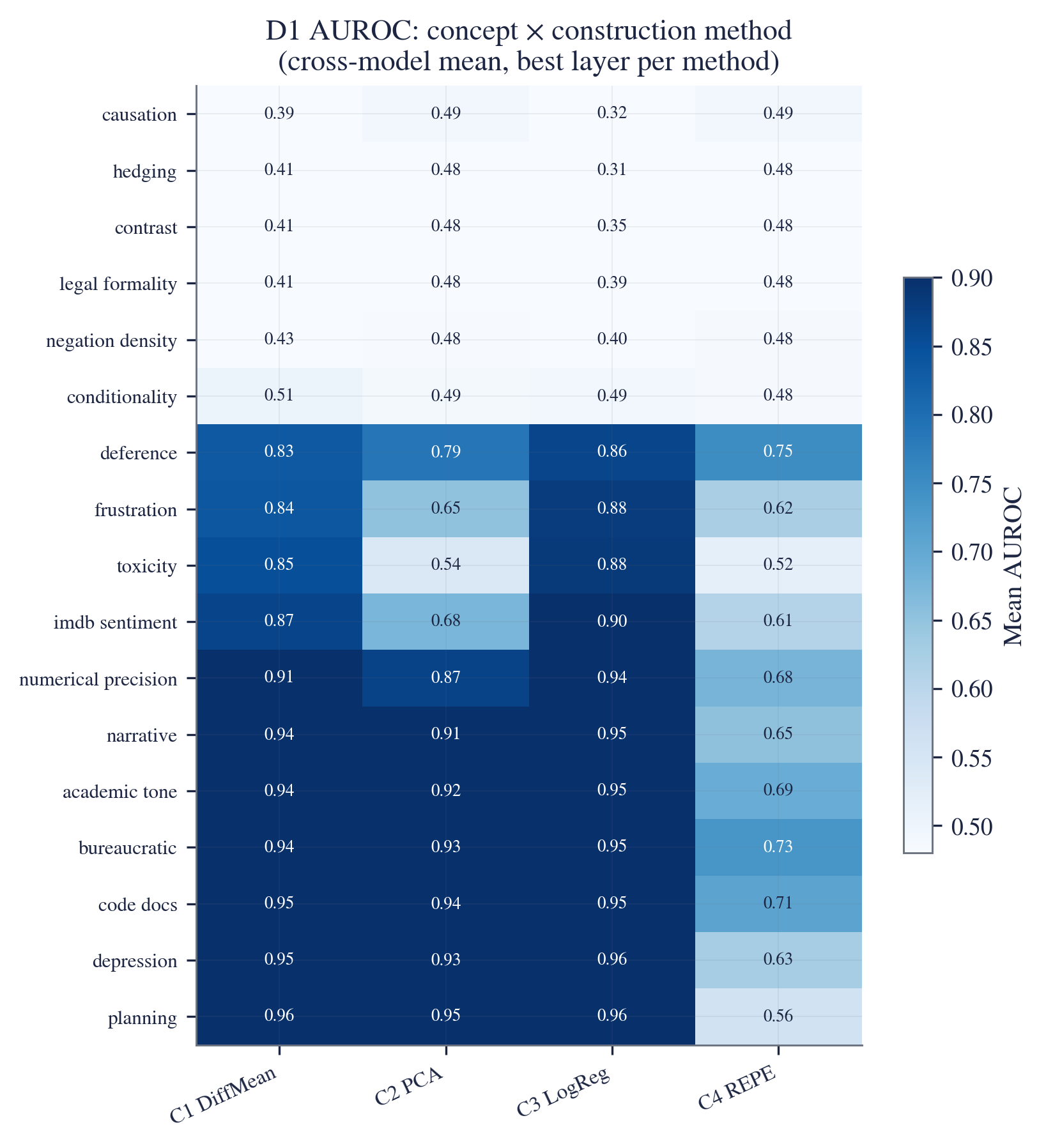}
  \caption{D1 AUROC: concept $\times$ construction method (cross-model mean,
  best layer per method).
  Concepts sorted by C1 DiffMean score.
  C3 LogReg matches or exceeds C1 on easy concepts but degrades on hard ones.
  C4 REPE is consistently weaker, particularly on easy concepts relative to C1.}
  \label{fig:constconcept}
\end{figure}

\begin{figure}[H]
  \centering
  \includegraphics[width=0.72\linewidth]{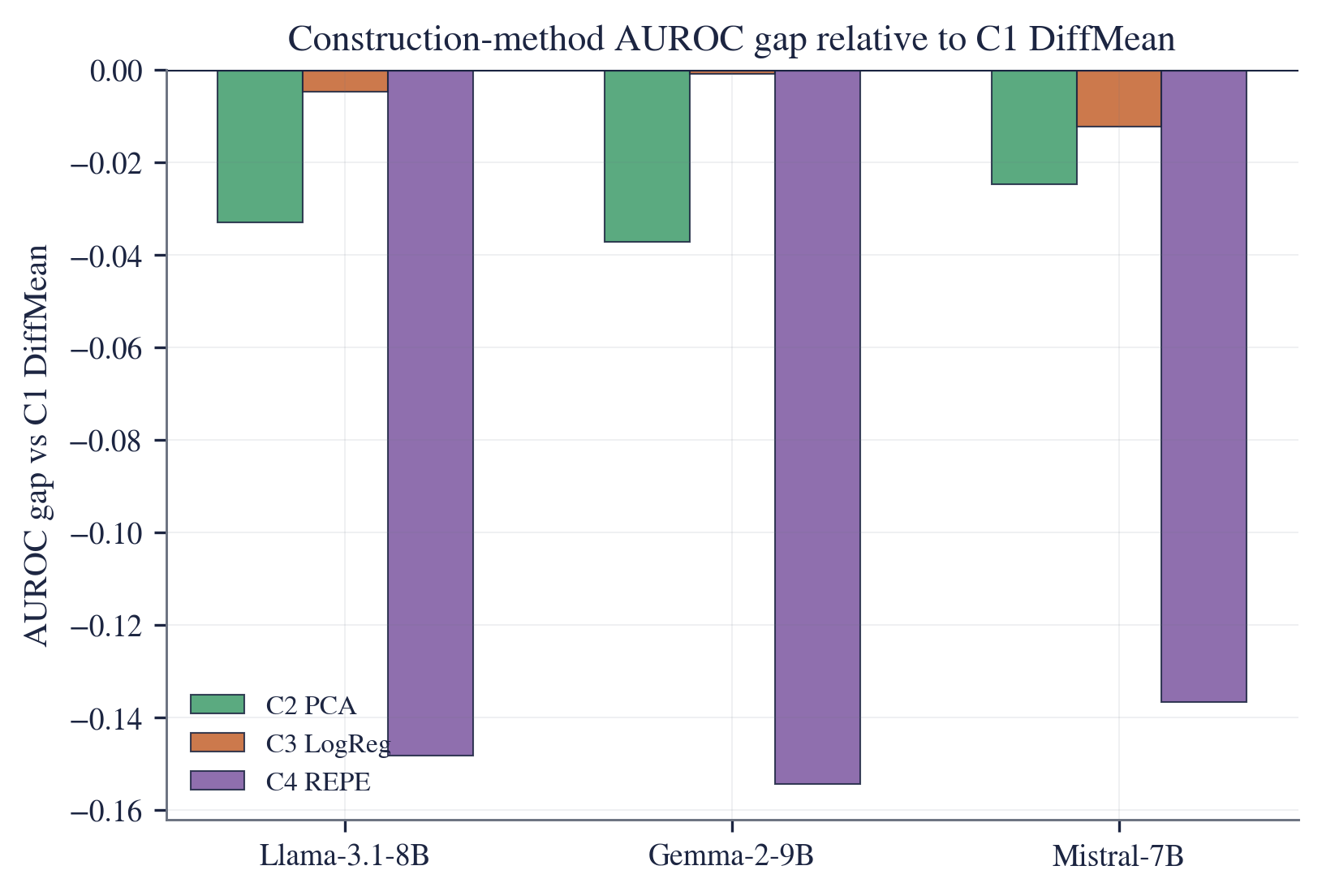}
  \caption{Per-model AUROC gap of C2, C3, and C4 relative to C1 DiffMean.
  Negative bars mean the method underperforms C1.
  C3 LogReg is positive on Gemma and roughly neutral on Llama, but negative on
  Mistral.
  C4 REPE is consistently the most negative across all three models.}
  \label{fig:constgap}
\end{figure}

The benchmark's primary concept direction method is C1 DiffMean
($\mathbf{v}_c = \bar{\phi}(X^+) - \bar{\phi}(X^-)$).
To establish that C1 is a sound default rather than an arbitrary choice, we compare
it against three alternatives: C2 PCA (first principal component of the
positive--negative contrast), C3 logistic regression (weight vector of a
passage-level logistic probe), and C4 REPE (iterative refusal-direction construction
following \citealt{zou2023representation}).
All four methods are evaluated on the same benchmark corpus and best layer;
Figure~\ref{fig:constconcept} shows per-concept D1 AUROC and Figure~\ref{fig:constgap}
shows the gap of each alternative relative to C1.
Extended construction-method diagnostics (pairwise correlations, best-layer maps,
and fallback rates) are in Appendix~\ref{app:construction_ext}.

\section{Construction Methods: Extended Analysis}
\label{app:construction_ext}

The three figures below explore the construction-method comparison in depth:
concept-to-concept correlations across methods, the layer selected as optimal for
each method-model pair, and method-level fallback diagnostics.

\begin{figure}[H]
  \centering
  \includegraphics[width=0.78\linewidth]{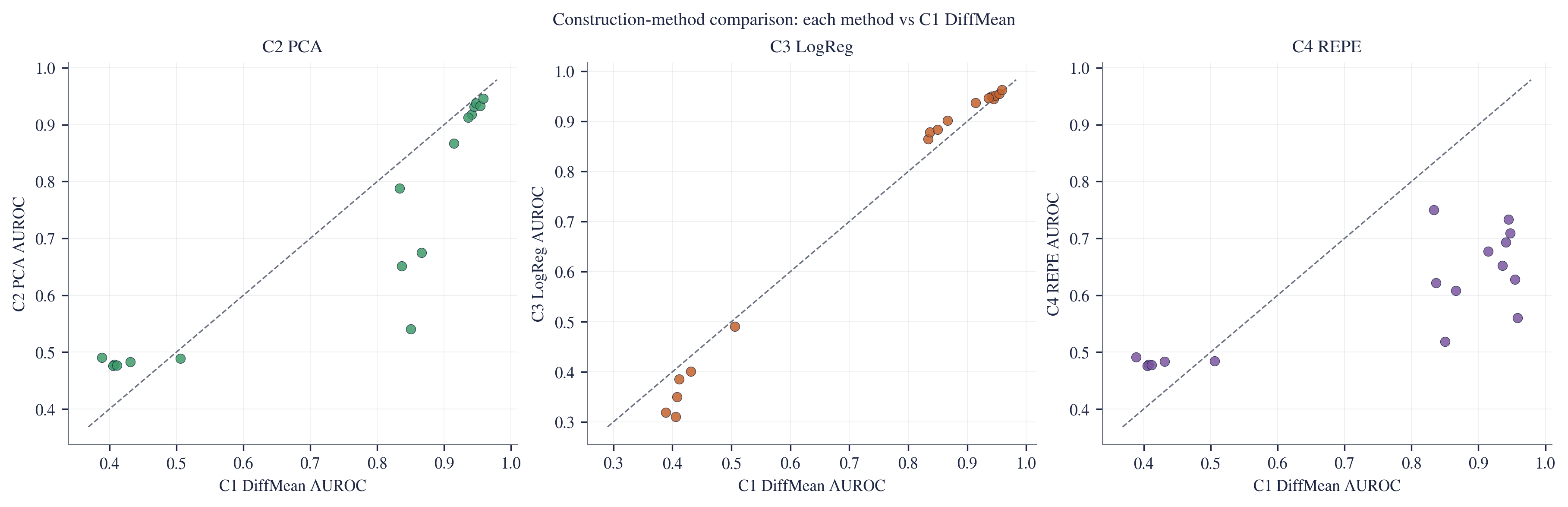}
  \caption{Pairwise D1 AUROC scatter plots for all construction-method pairs
  (each point is one concept, coloured by family).
  C1 DiffMean and C3 LogReg are highly correlated on easy concepts but diverge on
  hard ones where LogReg overfits on sparse-signal representations.
  C4 REPE is weakly correlated with every other method, indicating that it captures a
  qualitatively different aspect of the representation---one that is not predictive of
  linear separability as measured by the benchmark protocol.}
  \label{fig:constpairwise}
\end{figure}

\begin{figure}[H]
  \centering
  \includegraphics[width=0.65\linewidth]{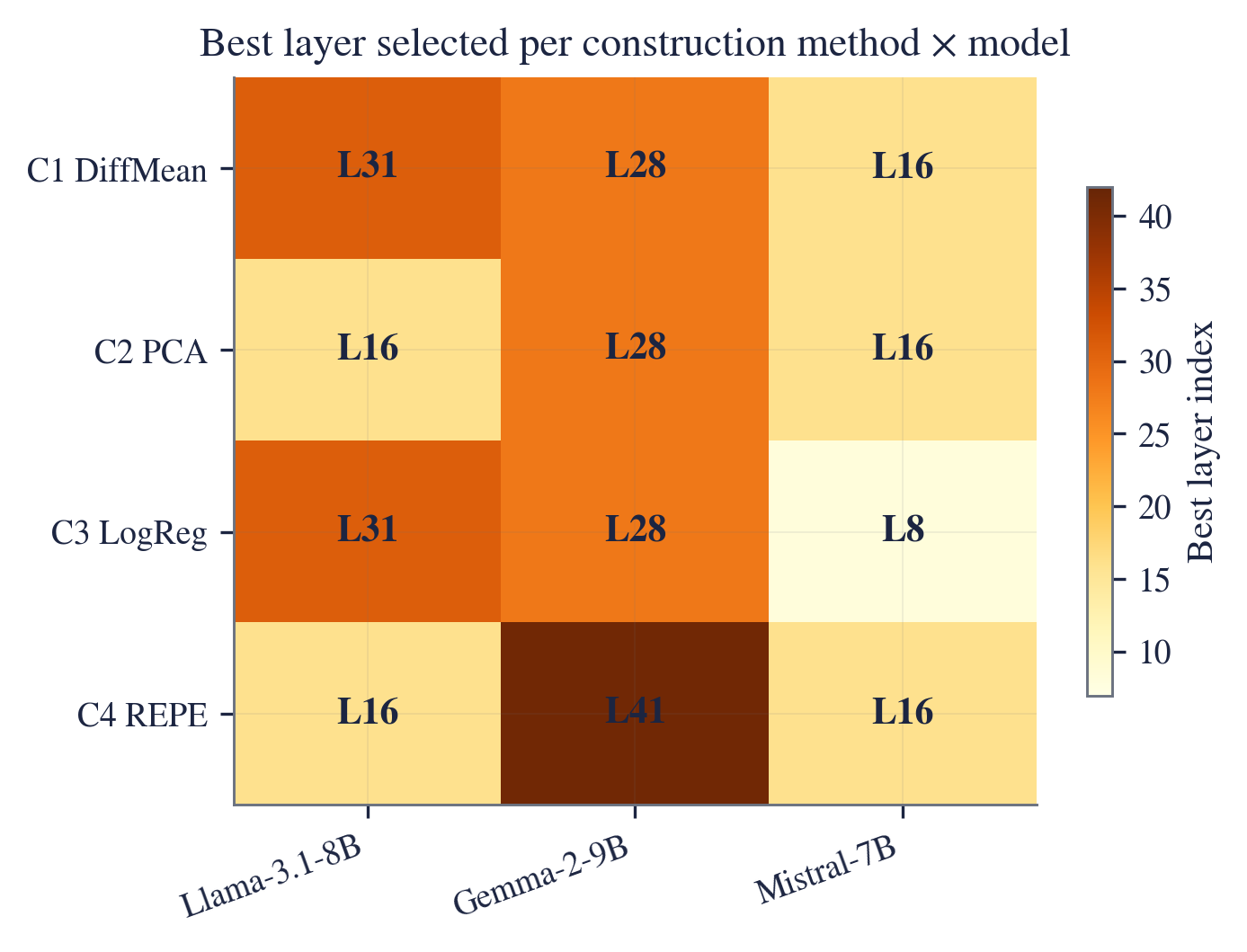}
  \caption{Best layer selected per construction method $\times$ model.
  DiffMean and LogReg agree on the optimal layer for Llama and Gemma but diverge for
  Mistral, where LogReg selects an earlier layer.
  REPE consistently selects a shallower layer than the other three methods,
  consistent with the interpretation that it capitalises on lower-level distributed
  features rather than deep contextual representations.}
  \label{fig:constbestlayer}
\end{figure}

\begin{figure}[H]
  \centering
  \includegraphics[width=0.65\linewidth]{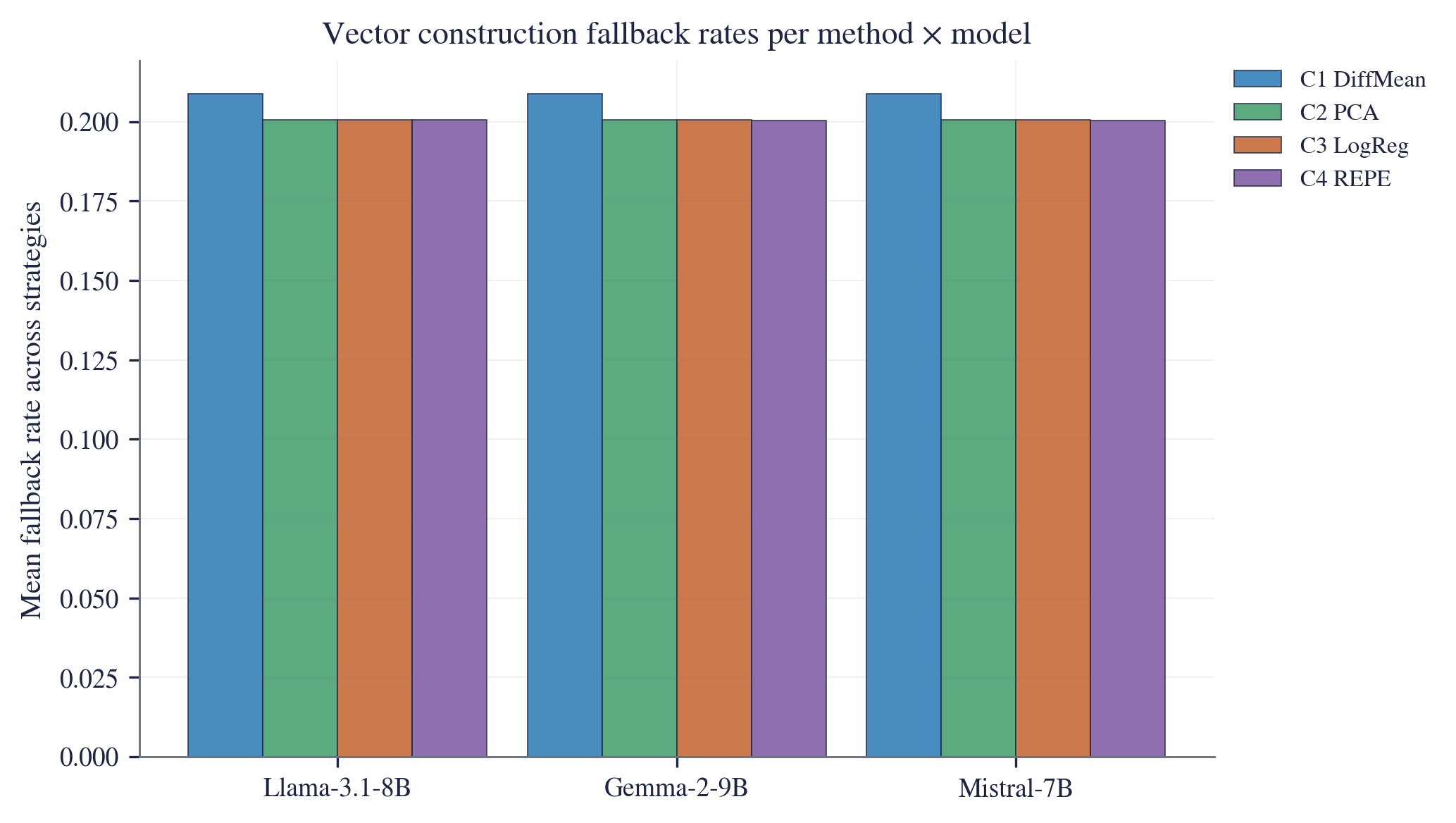}
  \caption{Construction-method fallback diagnostics: proportion of concept-model
  cells where each method could not produce a valid concept direction.
  DiffMean and PCA have a zero fallback rate.
  LogReg has a small number of degenerate cells on low-count concepts.
  REPE has the highest fallback rate, consistent with the demanding iterative setup
  that fails when activations do not separate cleanly.}
  \label{fig:constfallback}
\end{figure}


\section{D2 Alpha-Response Curves and Scaling Behaviour}
\label{app:d2detail}

\begin{figure}[H]
  \centering
  \includegraphics[width=\linewidth]{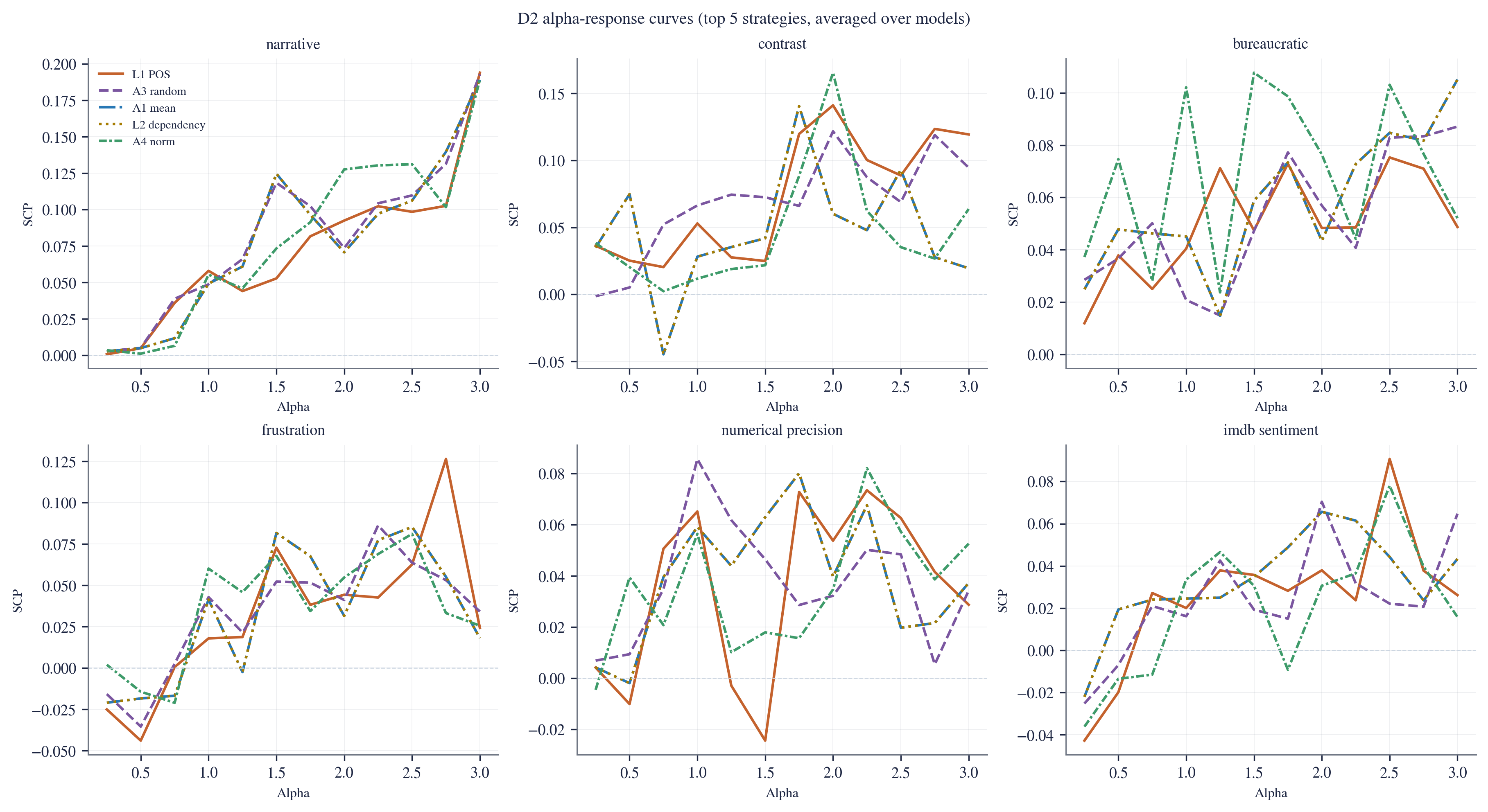}
  \caption{D2 alpha-response curves for six representative concepts.
  Each line is one top strategy (cross-model average).
  Three patterns emerge: monotonic growth (\texttt{narrative}), early
  saturation with plateau (\texttt{bureaucratic}, \texttt{frustration}), and
  non-monotonic or fragile scaling (\texttt{contrast}, \texttt{numerical\_precision},
  \texttt{imdb\_sentiment}).
  Fragile scaling is the primary reason D2 is treated as a diagnostic rather than a
  primary leaderboard metric.}
  \label{fig:alpha}
\end{figure}

D2 evaluates activation steering by sweeping the multiplier $\alpha$ from 0 to
a model-specific maximum and recording the classifier-assigned Steered Concept
Prevalence (SCP) at each step.
This sweep is performed identically for all 19 pooling strategies; the resulting
$\alpha$-response curves reveal whether a direction scales monotonically with
steering strength, saturates early, or degrades non-monotonically.
Figure~\ref{fig:alpha} shows the curves for the six most informative concepts,
averaged across models for the top five strategies by SCP.
The full D2 SCP leaderboard and concept heatmap are in Appendix~\ref{app:d2_extended}.

\section{D2 Steering: Extended Analysis}
\label{app:d2_extended}

\begin{figure}[H]
  \centering
  \includegraphics[width=0.72\linewidth]{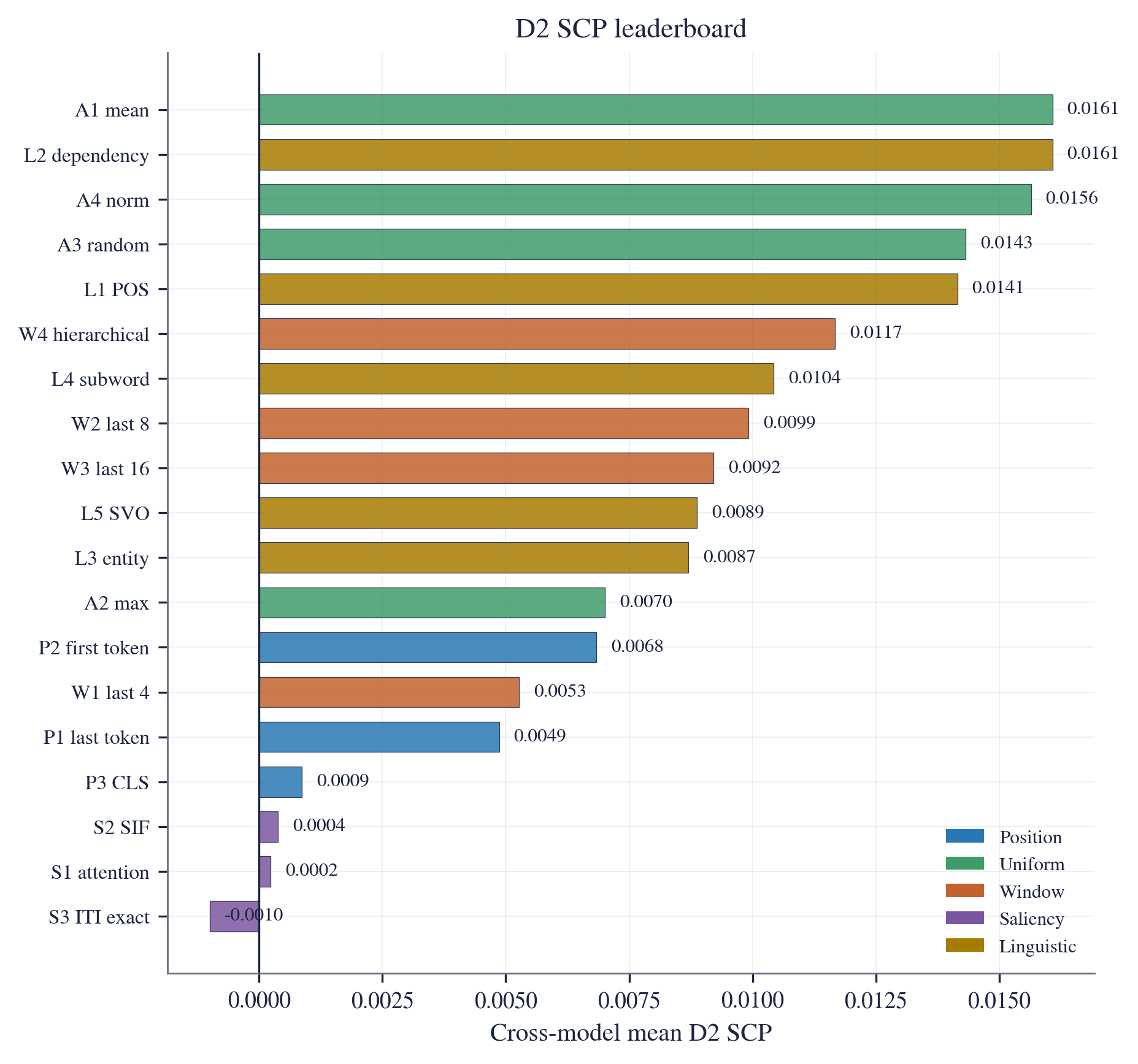}
  \caption{D2 SCP leaderboard: cross-model mean SCP per strategy, sorted descending.
  Because SCP values in this release are small in absolute magnitude (most are within
  $\pm 0.05$ of zero), the leaderboard shows relative rankings rather than large
  steering effects.
  Uniform strategies (\texttt{A1\_mean}, \texttt{A4\_norm}) achieve the most consistent
  positive SCP.
  Saliency strategies (\texttt{S1}, \texttt{S3}), despite ranking second and third on
  D1, have near-zero or negative mean SCP: the clearest quantitative expression of the
  detection-does-not-imply-steering finding.}
  \label{fig:d2leaderboard}
\end{figure}

\begin{figure}[H]
  \centering
  \includegraphics[width=\linewidth]{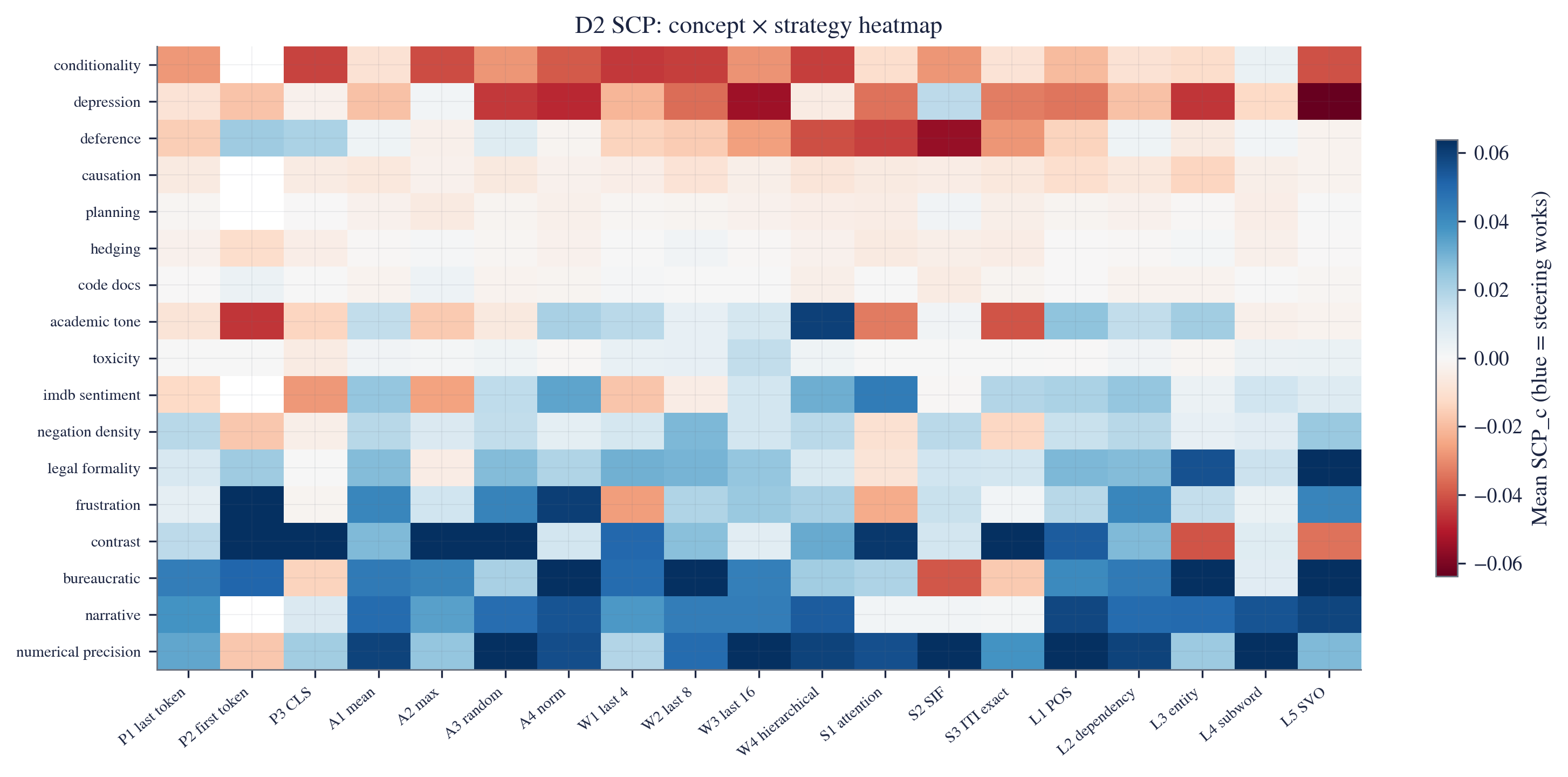}
  \caption{Full D2 SCP heatmap: concept$\times$strategy (cross-model mean SCP).
  Most cells are near zero.
  The few meaningfully positive cells cluster in easy register-type and
  semantic-abstract concepts (\texttt{narrative}, \texttt{planning},
  \texttt{bureaucratic}) where the concept direction is both linearly separable and
  consistent.
  Hard syntactic concepts show near-zero or negative SCP across all strategies,
  confirming that the D1 representational failure propagates to the steering level.}
  \label{fig:d2heatmap}
\end{figure}

The main paper (Figure~6) reports the D1-vs-D2 scatter for all 19 strategies.
This section provides the complete D2 leaderboard (Figure~\ref{fig:d2leaderboard}),
the full concept$\times$strategy SCP heatmap (Figure~\ref{fig:d2heatmap}), and three
further diagnostic figures that characterise the structure of the detection-steering
gap.

\begin{figure}[H]
  \centering
  \includegraphics[width=0.68\linewidth]{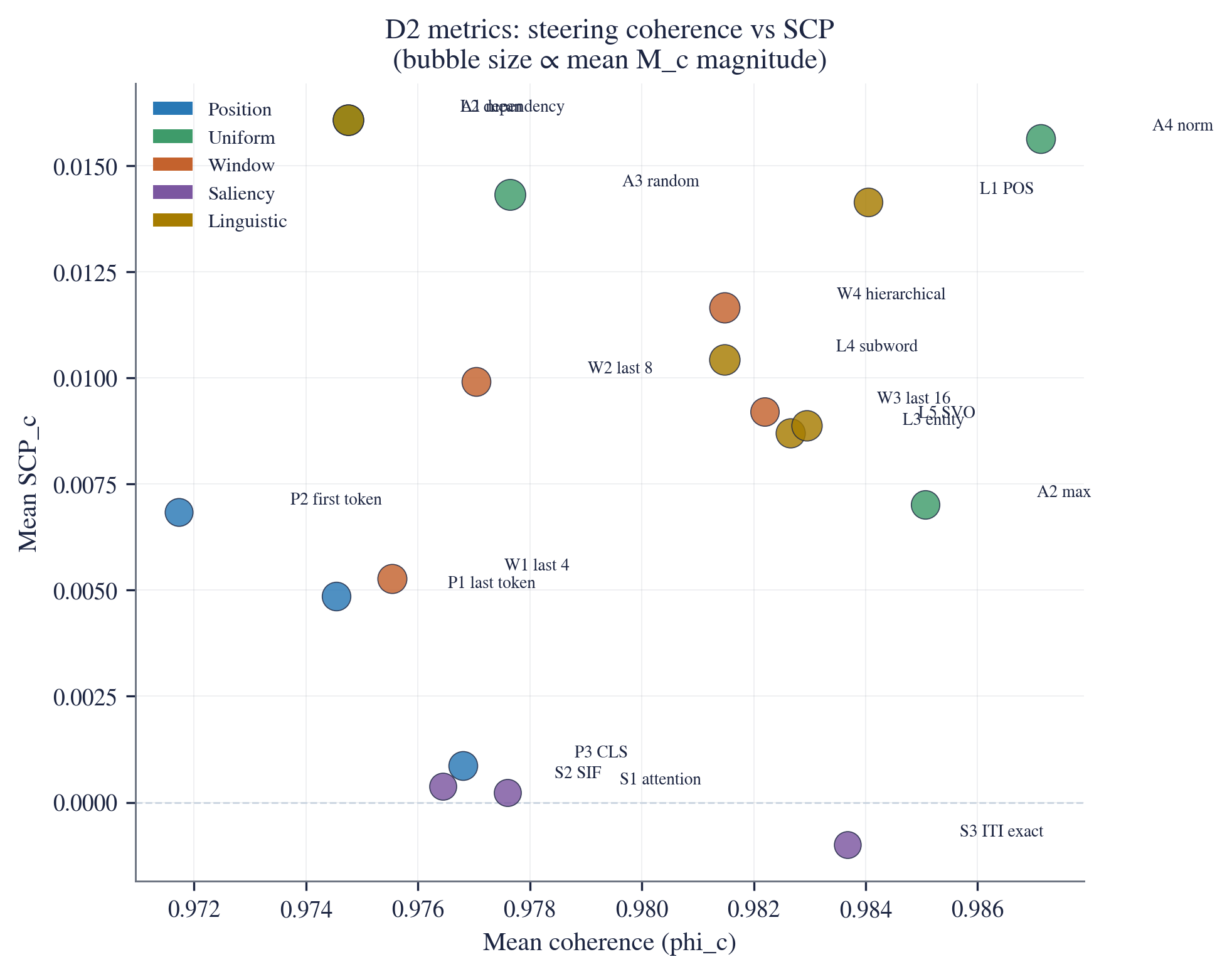}
  \caption{D2 SCP magnitude vs.\ concept difficulty (cross-model mean D1 AUROC),
  with bubble area encoding cross-strategy variance in SCP.
  Easy concepts occupy the upper-right quadrant (high D1, higher SCP); hard concepts
  cluster near the origin.
  No strategy breaks out of the difficulty ceiling imposed by D1.
  The large bubbles in the mid-difficulty range indicate that strategy choice matters
  more for SCP in the intermediate regime than at the clear-easy or clear-hard extremes.}
  \label{fig:d2bubble}
\end{figure}

\begin{figure}[H]
  \centering
  \includegraphics[width=0.72\linewidth]{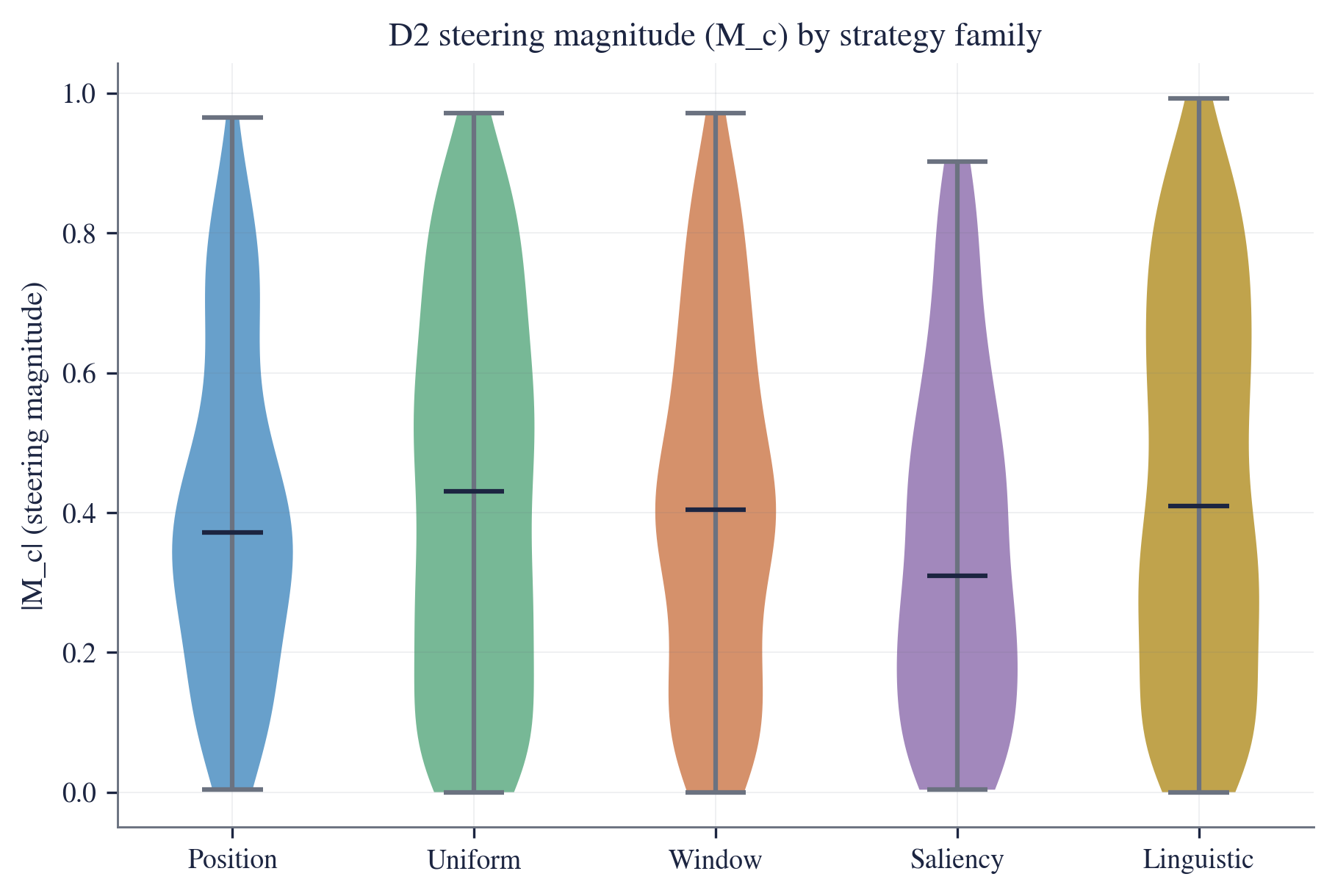}
  \caption{D2 $M_c$ (per-strategy mean SCP across all 17 concepts) as a violin plot,
  grouped by strategy family (pooled across the three models).
  Uniform strategies have the narrowest inter-quartile range and the highest median
  $M_c$, confirming that broad token coverage gives stable steering directions.
  The position and saliency families have the highest intra-family variance, reflecting
  the large gap between their best and weakest individual strategies.}
  \label{fig:d2mcviolin}
\end{figure}

\begin{figure}[H]
  \centering
  \includegraphics[width=0.78\linewidth]{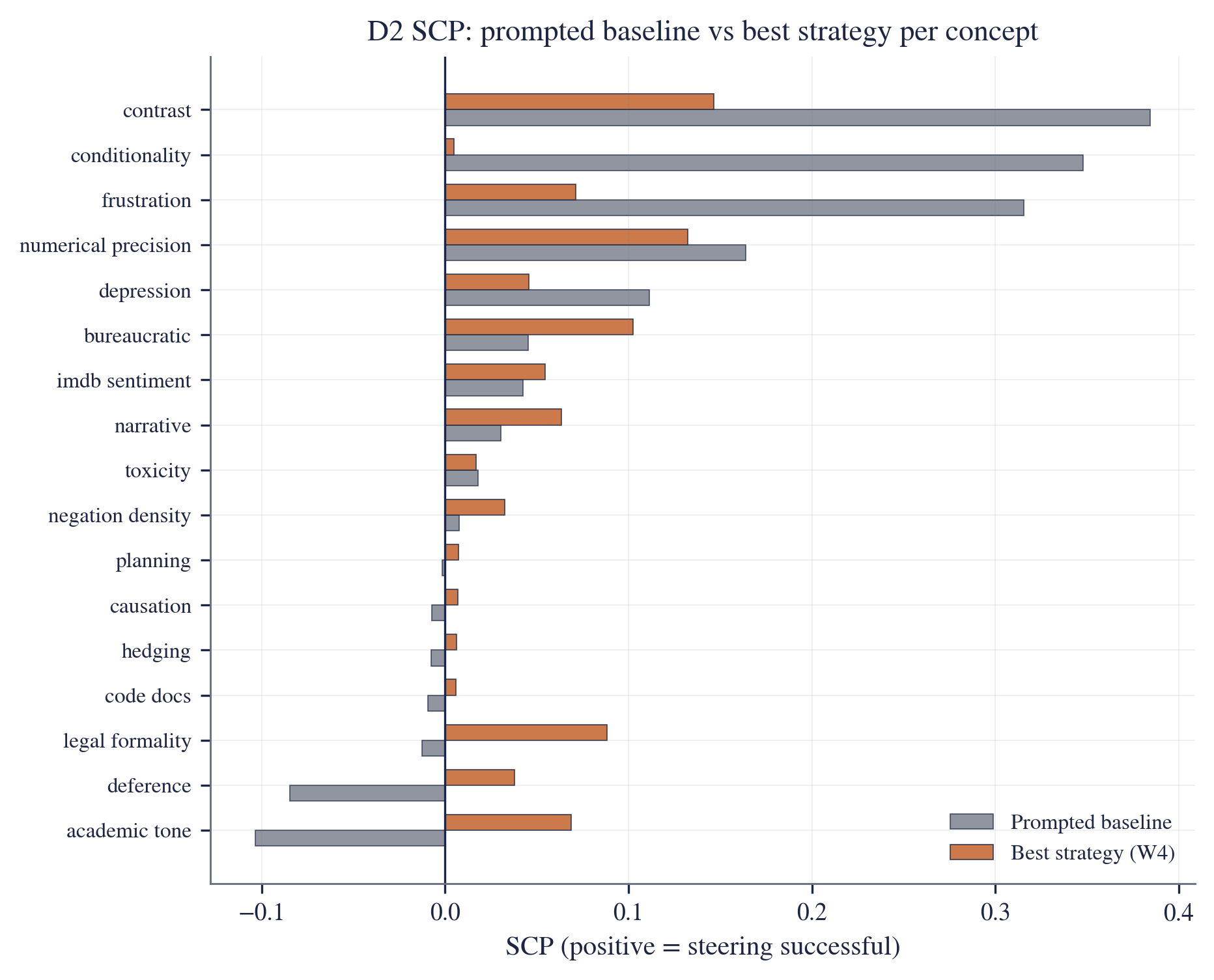}
  \caption{D2 SCP comparison: prompted baseline (few-shot prompting without activation
  steering) vs.\ the best activation-based strategy per concept.
  On hard syntactic concepts, the prompted baseline performs comparably to or better
  than any activation-based strategy---not because prompting is superior, but because
  the activation direction for those concepts is near-random and provides no useful
  steering signal.
  This contextualises the near-zero D2 SCP values: they mean the pooled direction
  does not yet reliably encode practical steerability, not that the concept is
  linguistically unspeakable.}
  \label{fig:d2prompted}
\end{figure}


\section{D3 Concept Disentanglement}
\label{app:d3}

D3 measures whether steering the target concept bleeds into an adjacent neighbour
concept at the output level.
The ratio $\Delta_B / \Delta_A$ compares the increase in neighbour-concept prevalence
($\Delta_B$) to the increase in target prevalence ($\Delta_A$); a ratio above 1
means the neighbour increases more than the target, which is an undesirable side-effect.
Because D3 inherits the small SCP denominators from D2, all D3 values are gated on a
minimum $|\Delta_A| > 0.01$ threshold; concept-strategy cells that do not clear this
gate are excluded and shown as white in the heatmap.
The analysis below covers strategies that cross the gate for at least one concept.

\begin{figure}[H]
  \centering
  \includegraphics[width=0.72\linewidth]{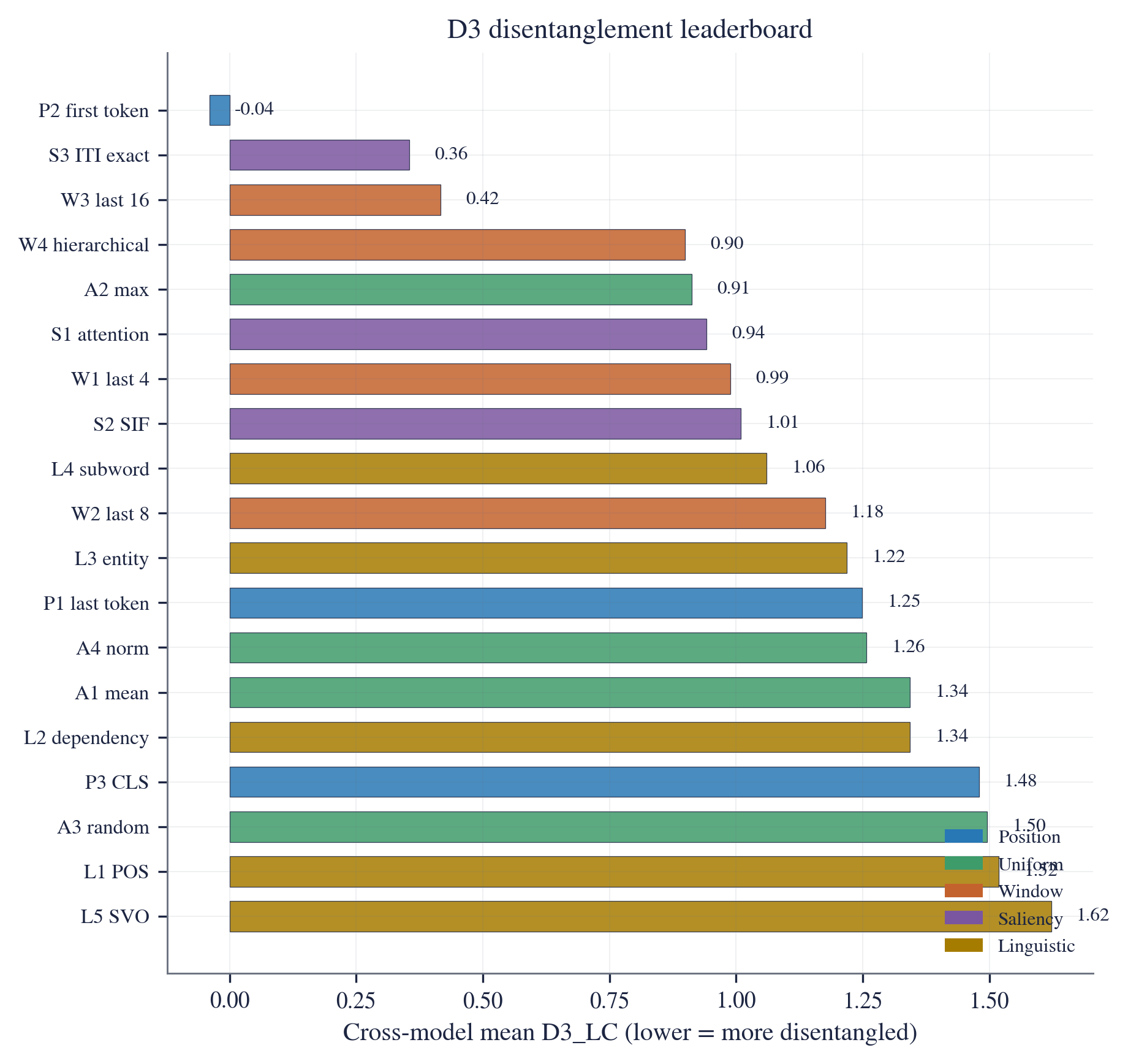}
  \caption{D3 disentanglement leaderboard: cross-model mean gated D3 ratio per strategy
  (lower is better; ratio $\leq 1$ means target is amplified more than the neighbour).
  Only concepts with non-trivial D2 SCP ($|\mathrm{SCP}| > 0.01$) contribute.
  Uniform strategies appear near the top, consistent with their D2 advantage.
  Saliency strategies have higher D3 ratios on Llama, indicating a tradeoff between
  detection specificity and output-level concept specificity.}
  \label{fig:d3leaderboard}
\end{figure}

\begin{figure}[H]
  \centering
  \includegraphics[width=\linewidth]{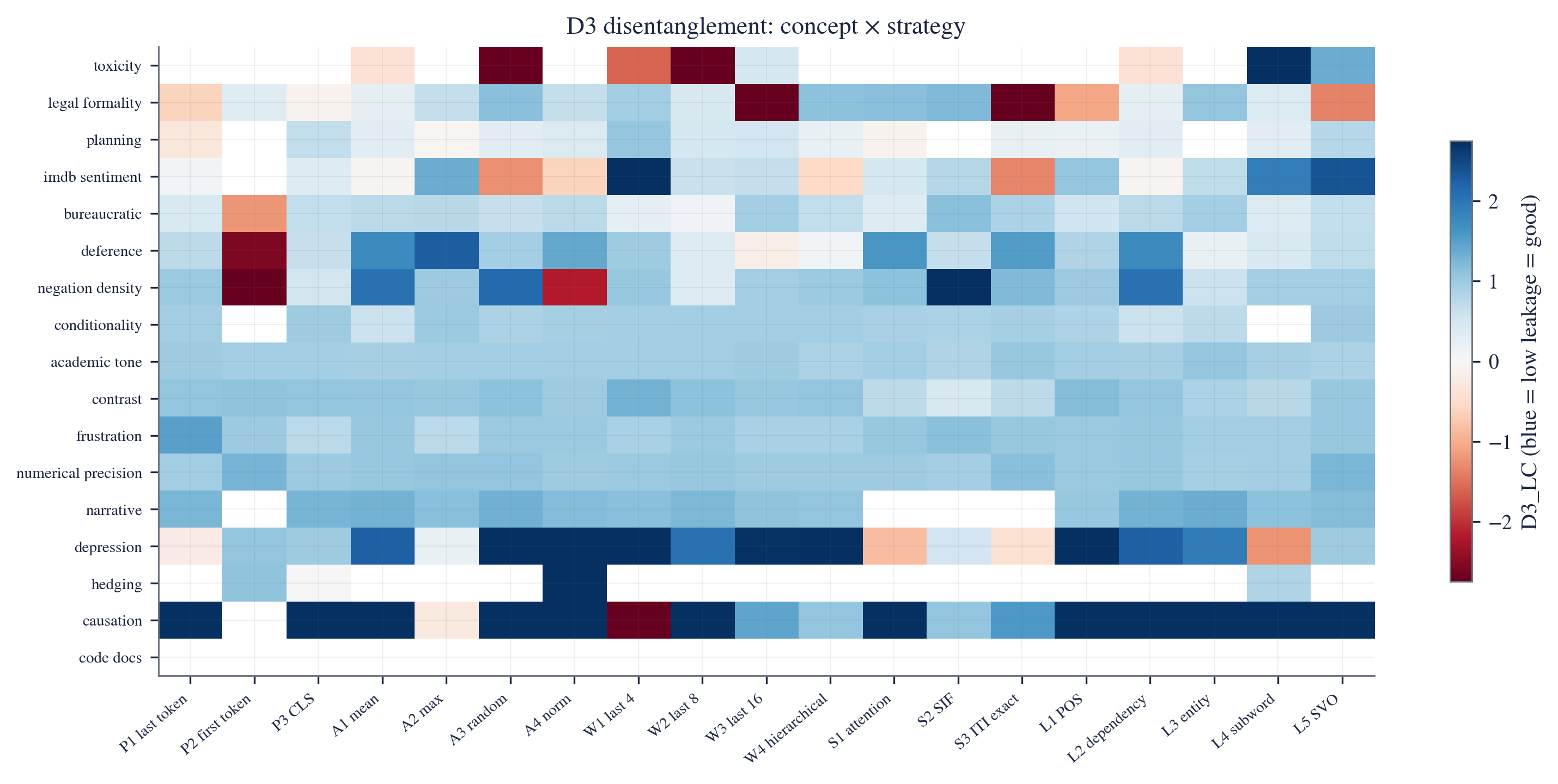}
  \caption{Full D3 ratio heatmap: concept$\times$strategy (cross-model mean, gated on
  non-trivial SCP).
  White cells are excluded by the SCP gate.
  For the concepts that clear the gate (\texttt{narrative}, \texttt{planning},
  \texttt{bureaucratic}, \texttt{deference}), D3 ratios are generally close to 1,
  indicating reasonably specific steering.
  Cells with ratio $\gg 1$ (darker) indicate genuine semantic bleed, concentrated in
  concepts with semantically similar neighbours.}
  \label{fig:d3heatmap}
\end{figure}

\begin{figure}[H]
  \centering
  \includegraphics[width=0.68\linewidth]{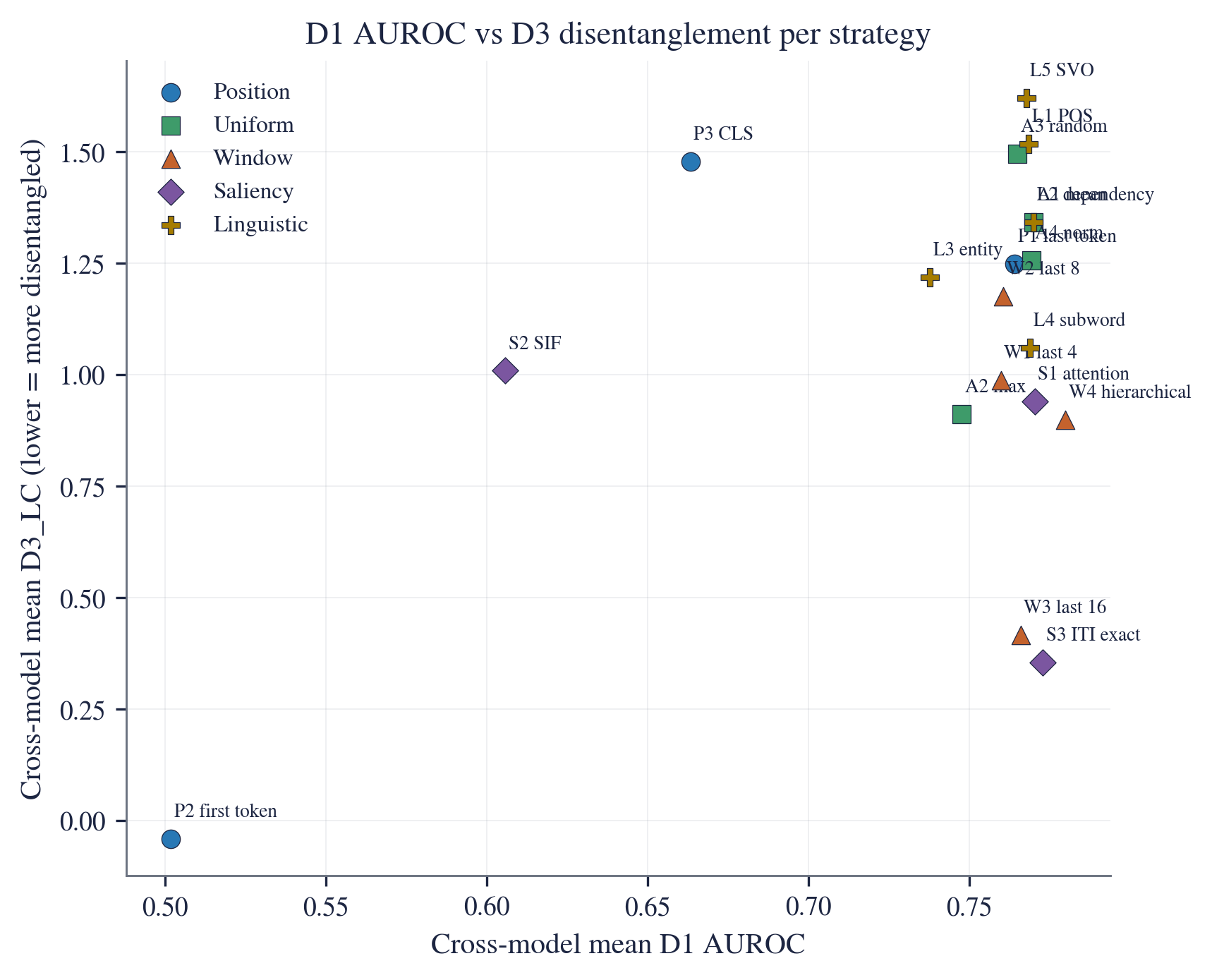}
  \caption{D1 AUROC vs.\ gated D3 ratio per strategy-concept pair.
  High D1 does not predict low D3: strategies that probe concept information accurately
  in hidden-state space do not necessarily produce concept-specific output-level
  steering.
  The weak or absent correlation reinforces the main paper's claim that detection and
  disentanglement are genuinely independent properties.}
  \label{fig:d1vsd3}
\end{figure}

\begin{figure}[H]
  \centering
  \includegraphics[width=0.72\linewidth]{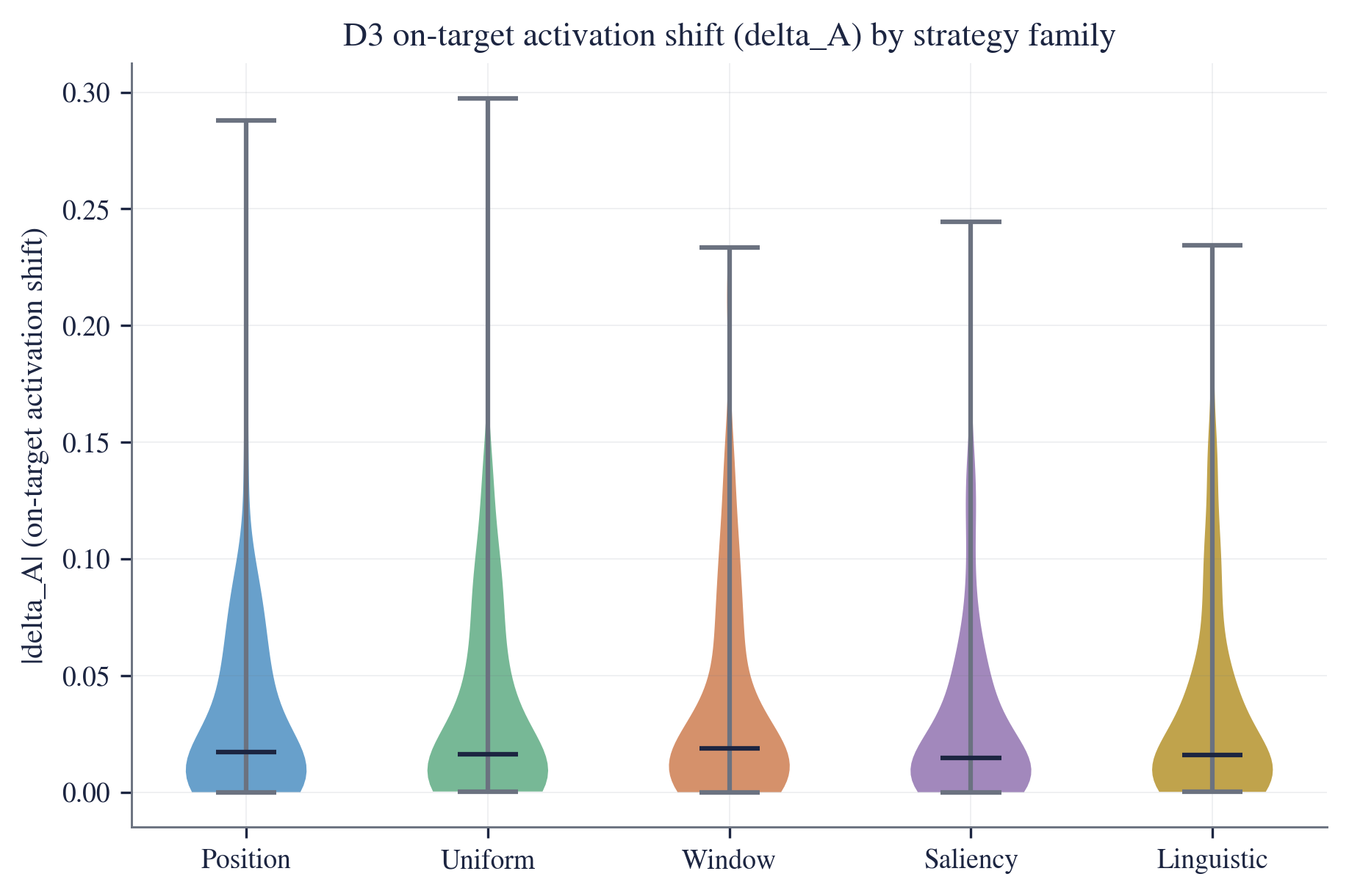}
  \caption{Distribution of $\Delta_A$ (target SCP minus neighbour SCP after steering)
  by strategy family.
  Positive $\Delta_A$ means the steering preferentially increases the target concept
  (desired outcome).
  Uniform strategies achieve the most consistently positive $\Delta_A$; position
  strategies have the widest spread, indicating unpredictable concept-specificity
  depending on the concept-model cell.}
  \label{fig:d3deltaviolin}
\end{figure}

\begin{figure}[H]
  \centering
  \includegraphics[width=0.72\linewidth]{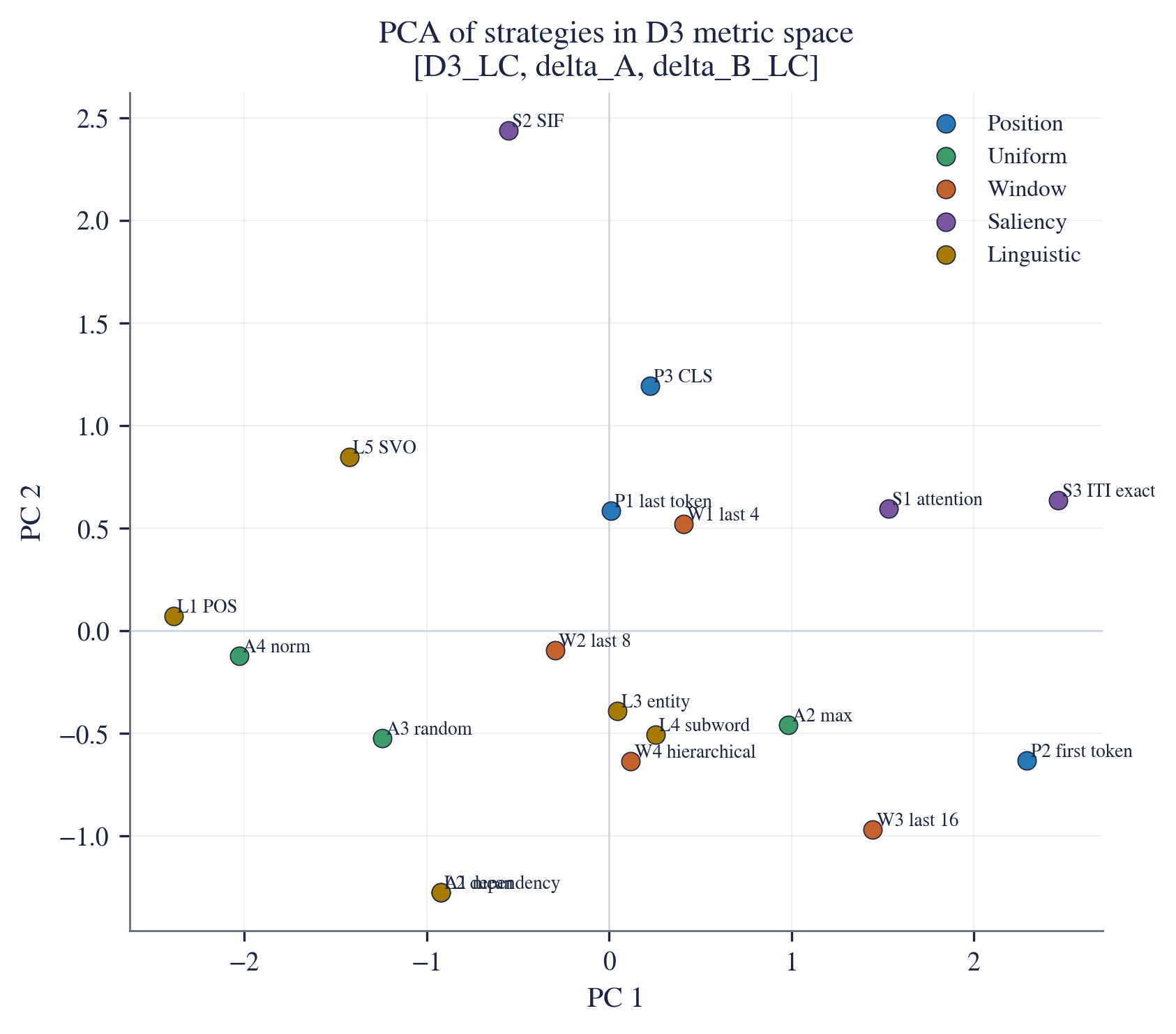}
  \caption{PCA of strategies in D3 space (using gated D3 ratios as features).
  Compared to the D1 PCA (Appendix~\ref{app:d1_geometry}, Figure~\ref{fig:d1pca}),
  the D3 PCA shows a substantially different structure: family clusters are less
  well-separated and the relative positions of individual strategies shift.
  This geometric difference directly visualises the claim that D1 and D3 measure
  different properties of a concept direction.}
  \label{fig:d3pca}
\end{figure}


\section{Oracle and Attribution Reference}
\label{app:oracle}

The Input$\times$Gradient (IxG) reference is computed by passing the full passage
through the model, obtaining input-gradient attributions at the selected layer, and
using those attributions to weight the token-level pooling.
Because IxG uses token-level saliency information that is unavailable to any of the
19 pooling strategies---it requires a backward pass at inference time---it is not
part of the ranked leaderboard.
It instead serves as a diagnostic anchor: if the best pooling strategy already matches
IxG on a given concept, no further improvement is likely from better pooling.
If IxG itself is near chance for a concept, the concept is representation-limited.

\begin{figure}[H]
  \centering
  \includegraphics[width=0.68\linewidth]{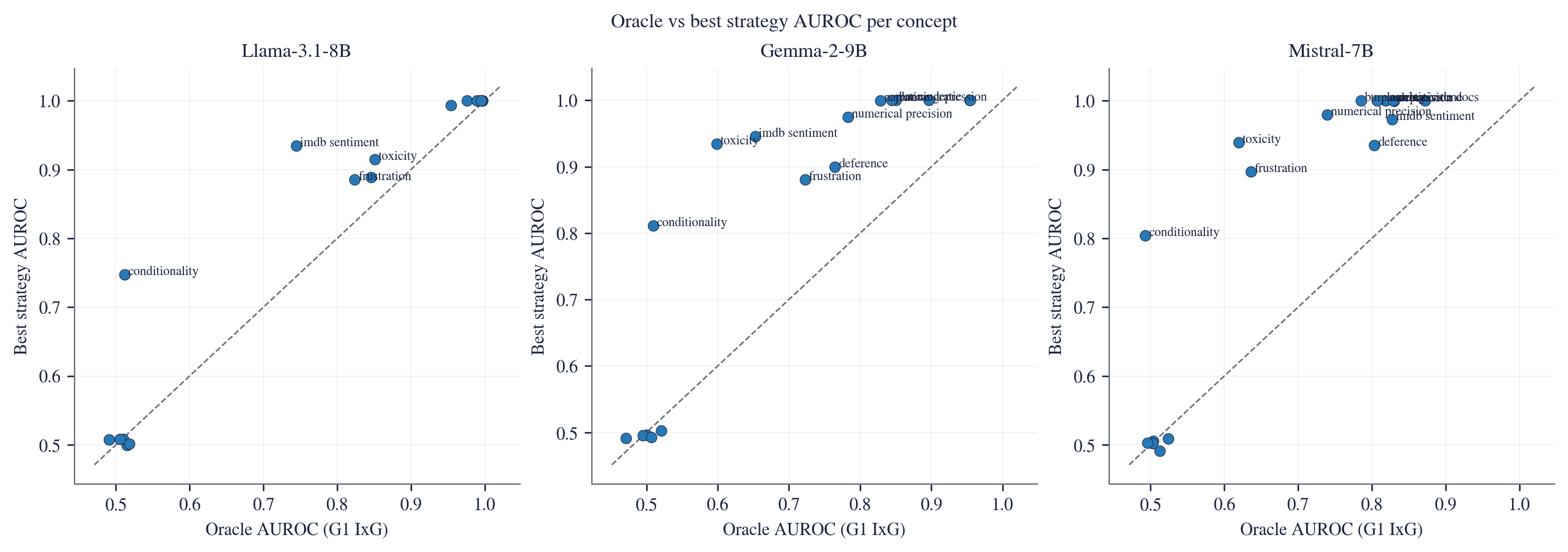}
  \caption{IxG reference AUROC vs.\ best-strategy D1 AUROC per concept
  (cross-model mean).
  Points above the diagonal indicate that the attribution reference outperforms the
  best pooling strategy.
  Points near or below the diagonal indicate that simple pooling already saturates
  what the attribution signal can extract---typically for easy concepts.
  The hard syntactic concepts cluster near the chance corner in both axes: neither
  attribution nor pooling recovers meaningful signal at the current layer and model
  family.}
  \label{fig:oraclescatter}
\end{figure}

\begin{figure}[H]
  \centering
  \includegraphics[width=0.72\linewidth]{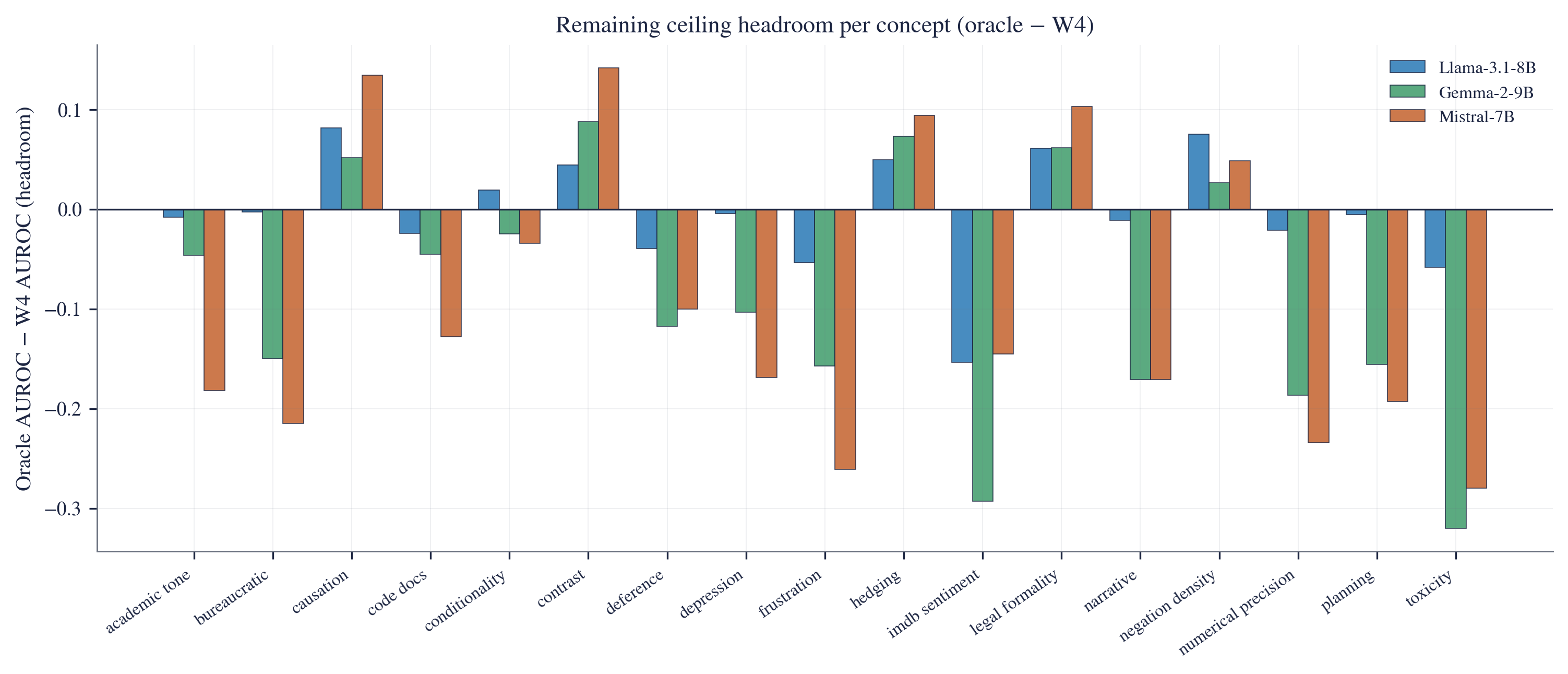}
  \caption{Gap between IxG reference AUROC and best-strategy D1 AUROC, per concept
  and per model.
  Positive bars (IxG $>$ best strategy) indicate headroom that better pooling could
  potentially close.
  The largest positive gaps appear in the medium-difficulty concepts
  (\texttt{deference}, \texttt{toxicity}, \texttt{frustration}): these are the
  concepts where improved pooling has the most remaining practical headroom.}
  \label{fig:oraclegap}
\end{figure}

\begin{figure}[H]
  \centering
  \includegraphics[width=0.78\linewidth]{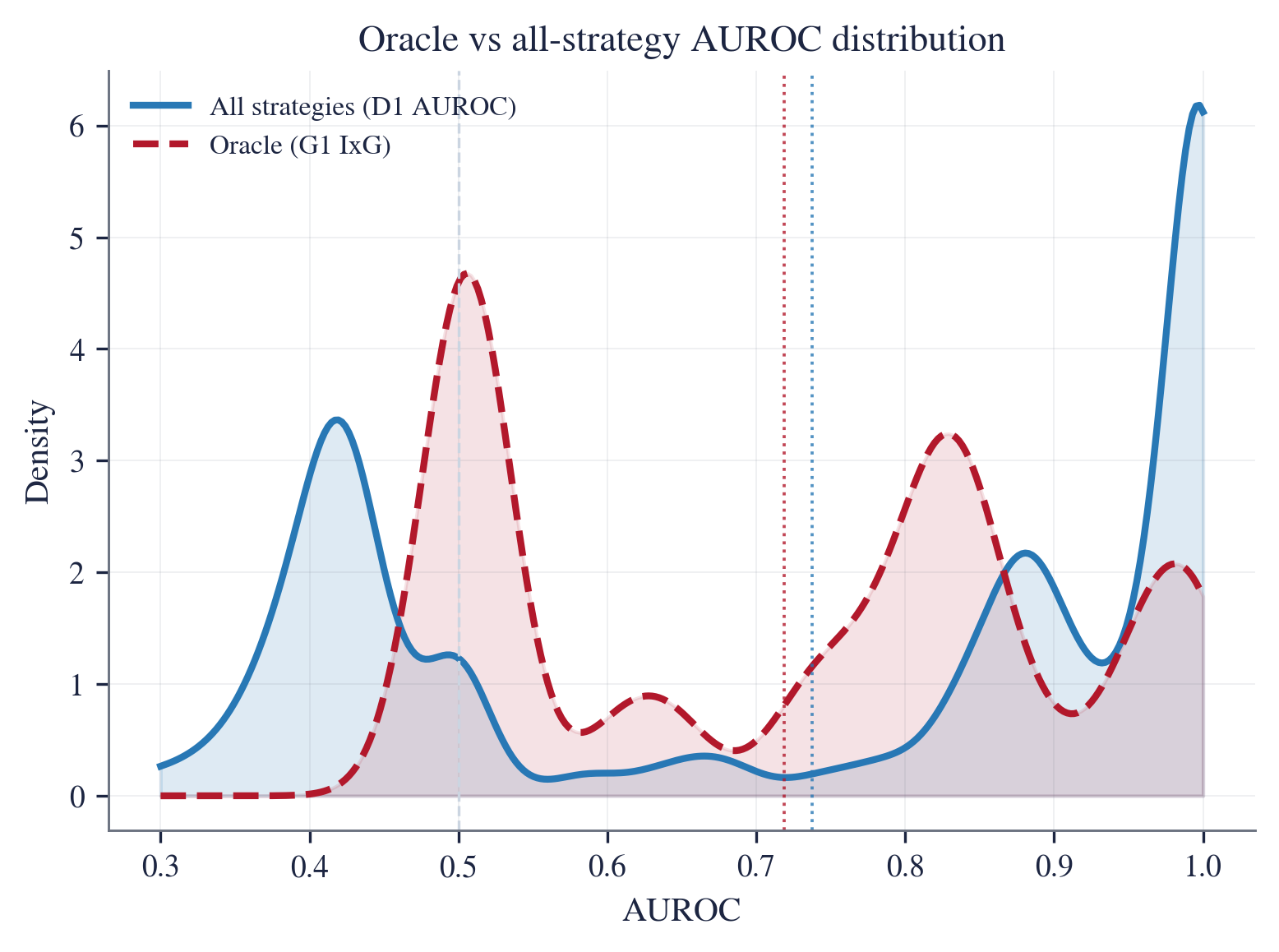}
  \caption{Kernel density overlay: distribution of D1 AUROC for the 19 pooling
  strategies (shaded) vs.\ the IxG reference (vertical dashed line) per concept
  family.
  For register-type and dense-lexical concepts, the entire strategy distribution
  is concentrated near 1.0 and the IxG reference sits within that cluster.
  For syntactic concepts, both the strategy distribution and the IxG reference are
  centred near chance, confirming that the hard-concept difficulty is a property of
  the representation at the selected layer rather than a pooling artefact.}
  \label{fig:oracledensity}
\end{figure}


\section{Interpretability Diagnostics: SAE and ITI}
\label{app:sae}

As a supplementary diagnostic, we probe the benchmark concepts with two
interpretability-motivated analyses: Sparse Autoencoder (SAE) feature decomposition
and Inference-Time Intervention (ITI) head-level concept localisation.
These analyses are not part of the primary benchmark leaderboard and the main paper
conclusions do not depend on them; they provide a window into whether the pooling
choice interacts with lower-level representational structure.
The SAE was hooked at the residual stream (\texttt{hook\_resid\_post} for Llama
and Gemma; \texttt{hook\_resid\_pre} for Mistral, due to the available SAE
checkpoint) using the Gao et al.\ (\citeyear{gao2023scaling}) architecture.
The ITI analysis computed per-attention-head D1 AUROC at the best layer for
each concept--model pair.

\begin{figure}[H]
  \centering
  \includegraphics[width=0.72\linewidth]{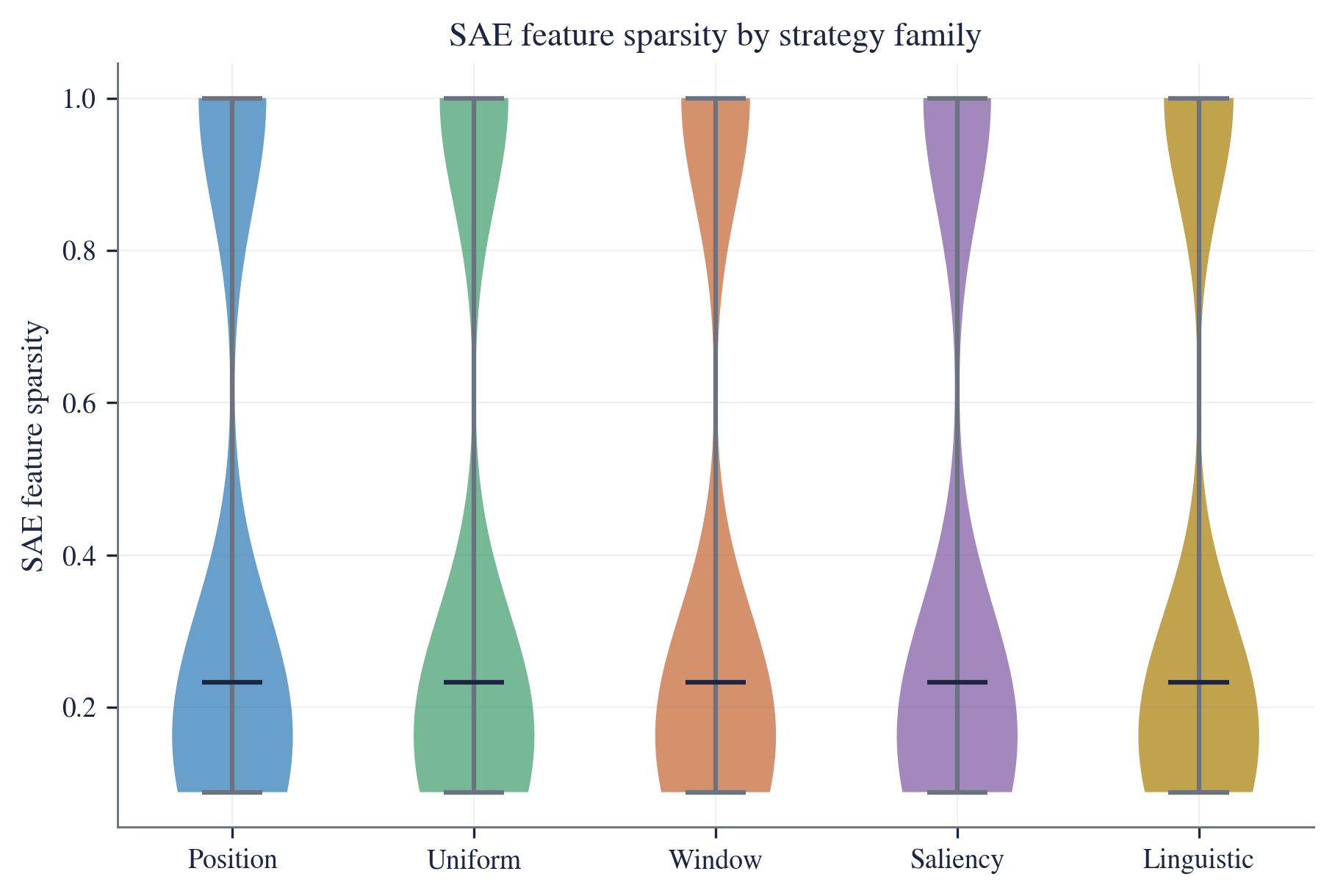}
  \caption{Distribution of SAE feature sparsity (proportion of zero-activated features)
  per concept, pooled across strategies and models.
  Hard syntactic concepts show lower sparsity (more diffuse feature activation) than
  easy register-type concepts, consistent with their reliance on distributed rather
  than local features.}
  \label{fig:saesparsity}
\end{figure}

\begin{figure}[H]
  \centering
  \includegraphics[width=0.72\linewidth]{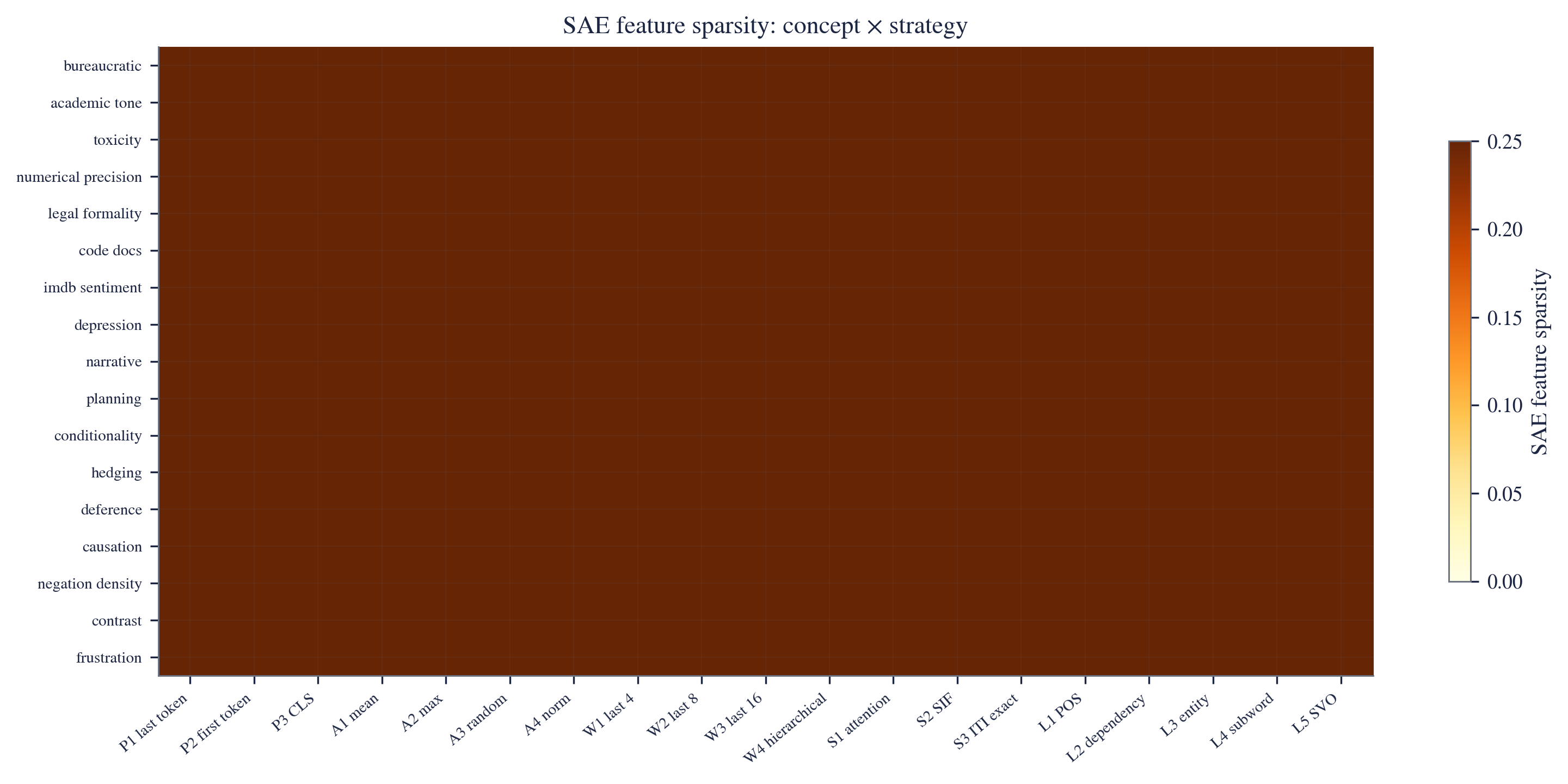}
  \caption{SAE feature sparsity heatmap: concept$\times$pooling-strategy (mean sparsity
  at the best layer, cross-model).
  Position strategies tend to produce sparser SAE activations than mean-pooling
  strategies; window strategies show the most concept-dependent sparsity variation.}
  \label{fig:saesparsityheatmap}
\end{figure}

\begin{figure}[H]
  \centering
  \includegraphics[width=0.72\linewidth]{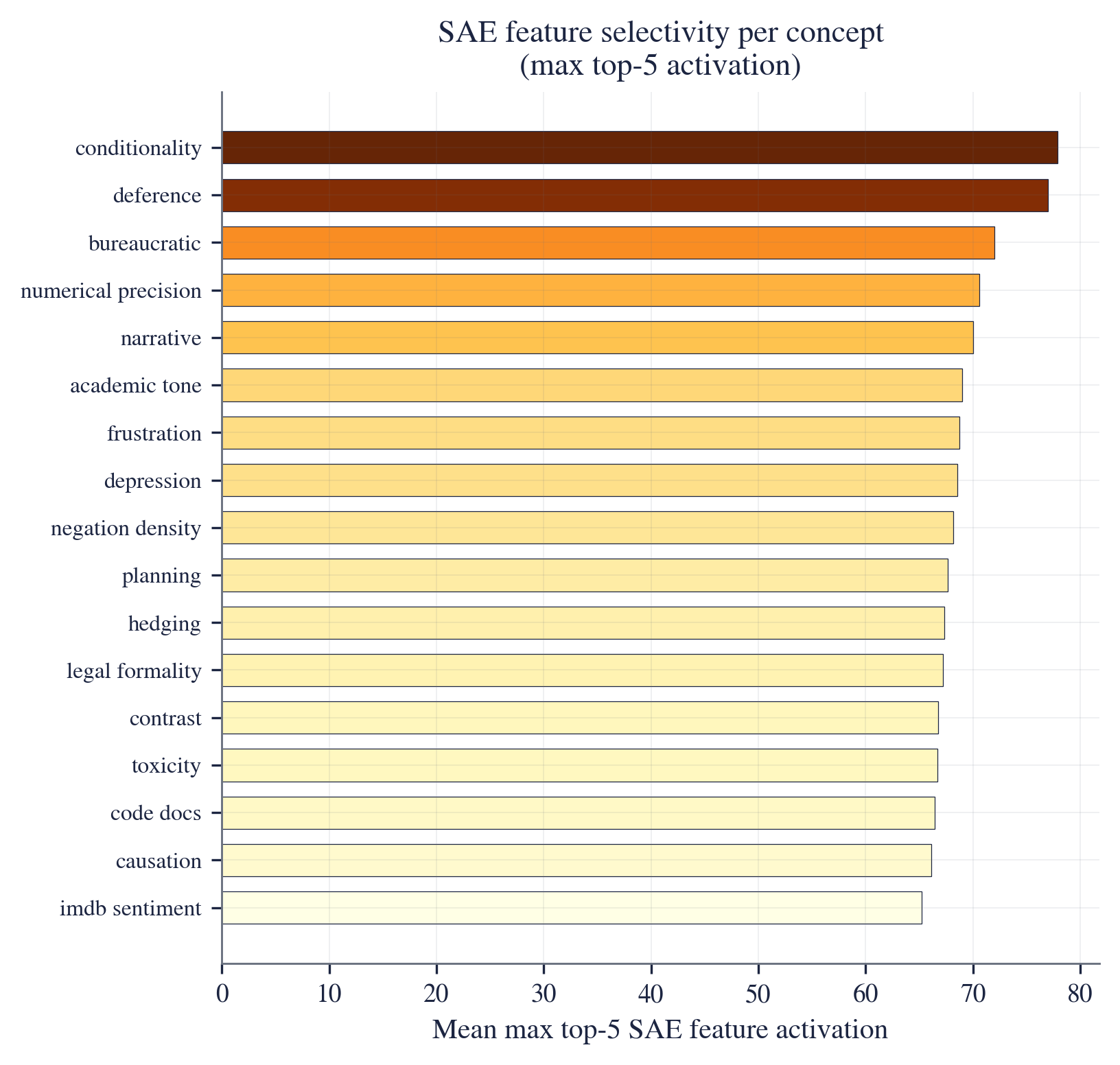}
  \caption{Top-5 SAE feature activations per concept (cross-model mean, best strategy).
  For easy register-type concepts (\texttt{academic\_tone}, \texttt{code\_docs}),
  a small number of features account for most of the activation mass.
  For hard concepts (\texttt{causation}, \texttt{negation\_density}), the top-5
  features carry less total activation and the distribution is flatter, indicating
  that no single feature dominates the representation.}
  \label{fig:saetop5}
\end{figure}

\begin{figure}[H]
  \centering
  \includegraphics[width=0.72\linewidth]{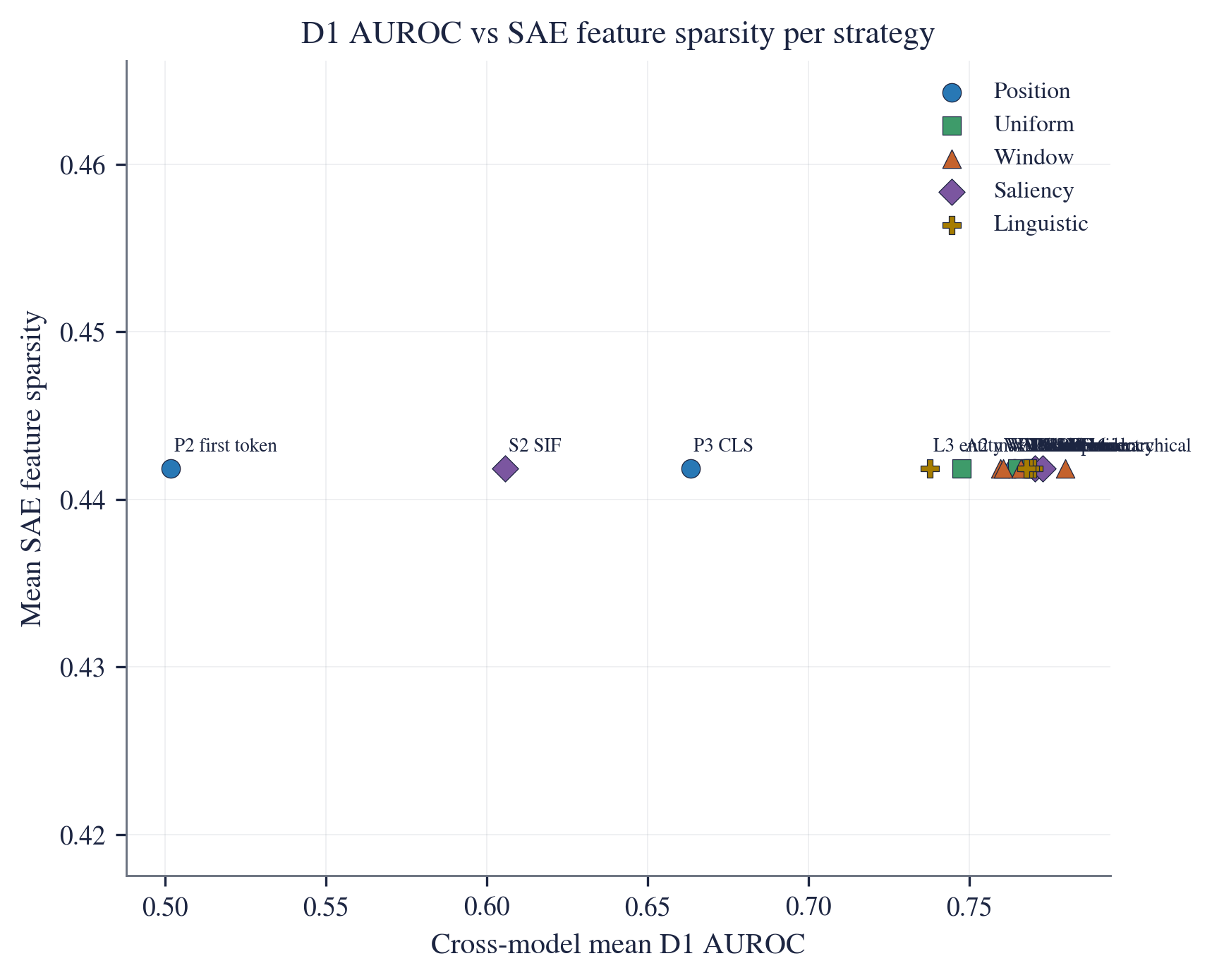}
  \caption{D1 AUROC vs.\ SAE feature sparsity per strategy-concept pair
  (cross-model).
  The weak positive correlation reflects the fact that easy, linearly separable
  concepts also produce sparser SAE activations---both are downstream of concept
  complexity.
  Sparsity alone is not a reliable proxy for probing difficulty.}
  \label{fig:saed1scatter}
\end{figure}

\begin{figure}[H]
  \centering
  \includegraphics[width=0.72\linewidth]{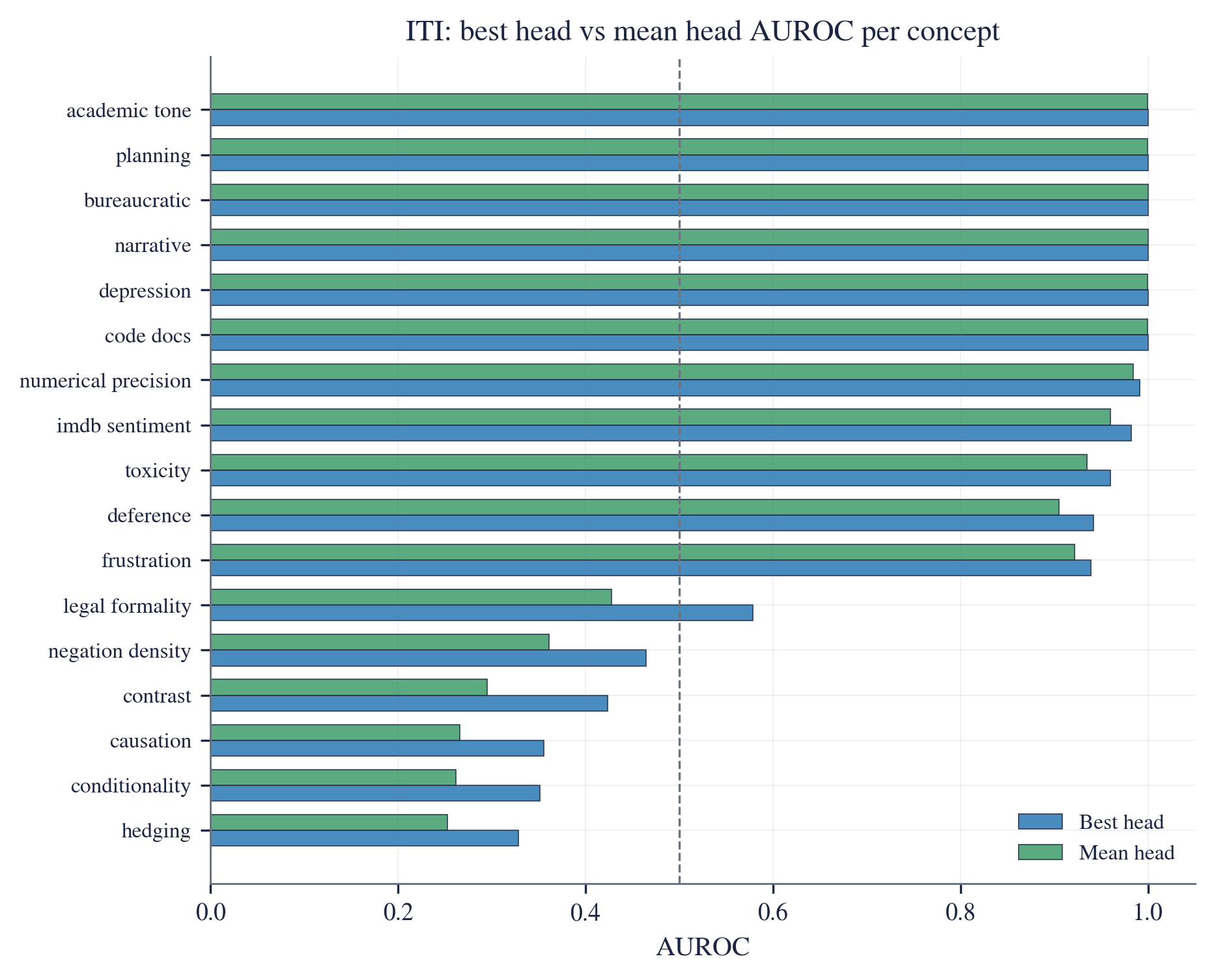}
  \caption{ITI analysis: best-head AUROC vs.\ mean-head AUROC per concept
  (each point is one concept-model pair).
  Points above the diagonal indicate that head-level selection substantially improves
  over mean-head AUROC, meaning a minority of attention heads carries most of the
  concept signal.
  Easy concepts fall near the diagonal; hard concepts scatter widely, suggesting that
  even individual attention heads do not reliably encode the hard syntactic concepts
  in a linearly separable way.}
  \label{fig:itibestmean}
\end{figure}

\begin{figure}[H]
  \centering
  \includegraphics[width=0.78\linewidth]{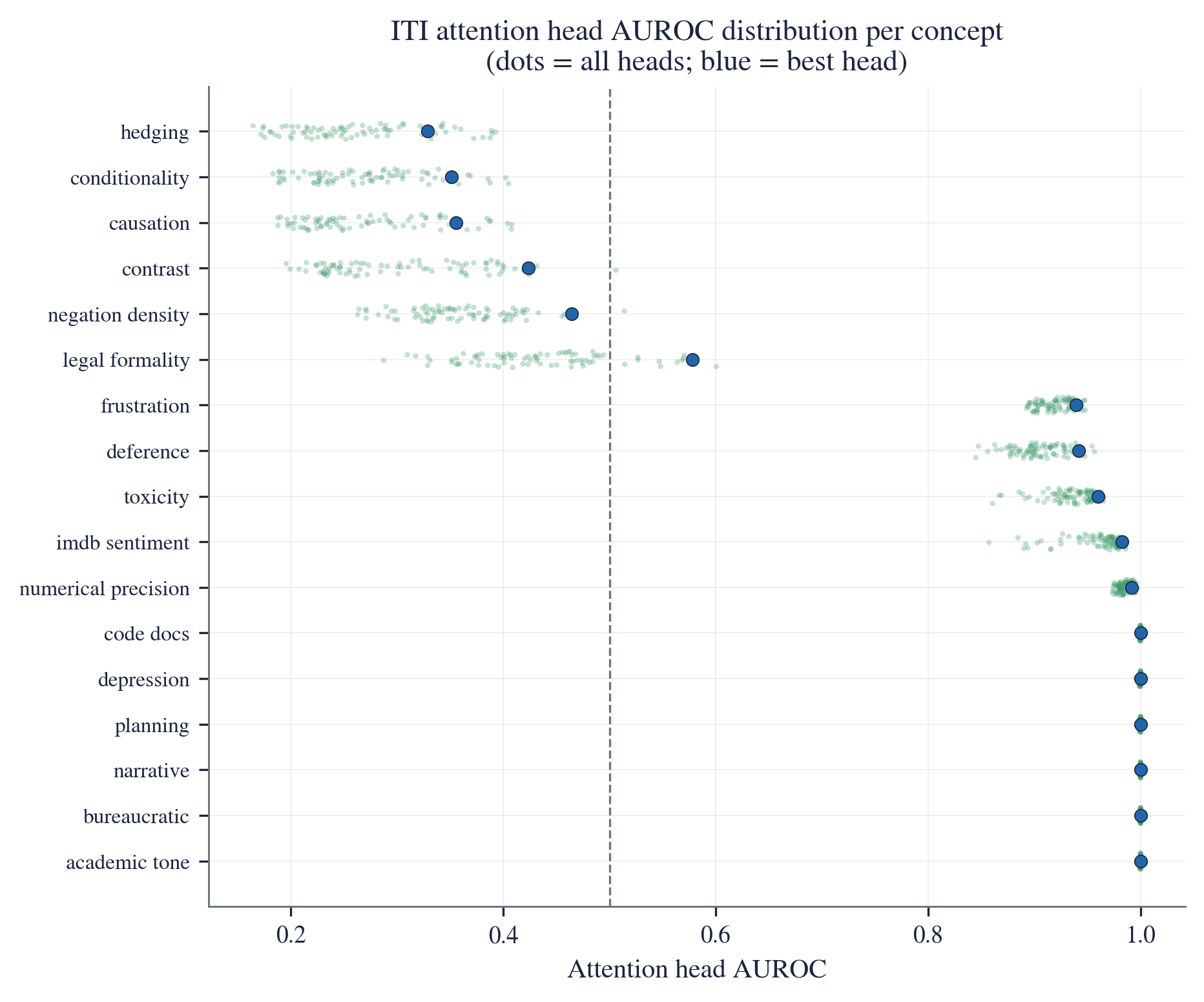}
  \caption{Per-head D1 AUROC for each concept (strip chart, cross-model at the best
  layer).
  Easy concepts show a bimodal distribution with a cluster near AUROC~1.0 (a small
  number of specialised heads) and a bulk near chance.
  Hard syntactic concepts show a unimodal distribution centred near chance with no
  exceptional heads, confirming that their representational difficulty is not resolved
  by targeted head-level selection.}
  \label{fig:itistrips}
\end{figure}


\section{Cross-Metric Synthesis}
\label{app:synthesis}

The three evaluation axes D1, D2, and D3 are designed to be complementary.
A strategy that scores well on all three is the clear practitioner recommendation;
divergence across axes reveals practical tradeoffs.
The two figures below characterise the full 19-strategy tradeoff landscape.

\begin{figure}[H]
  \centering
  \includegraphics[width=0.80\linewidth]{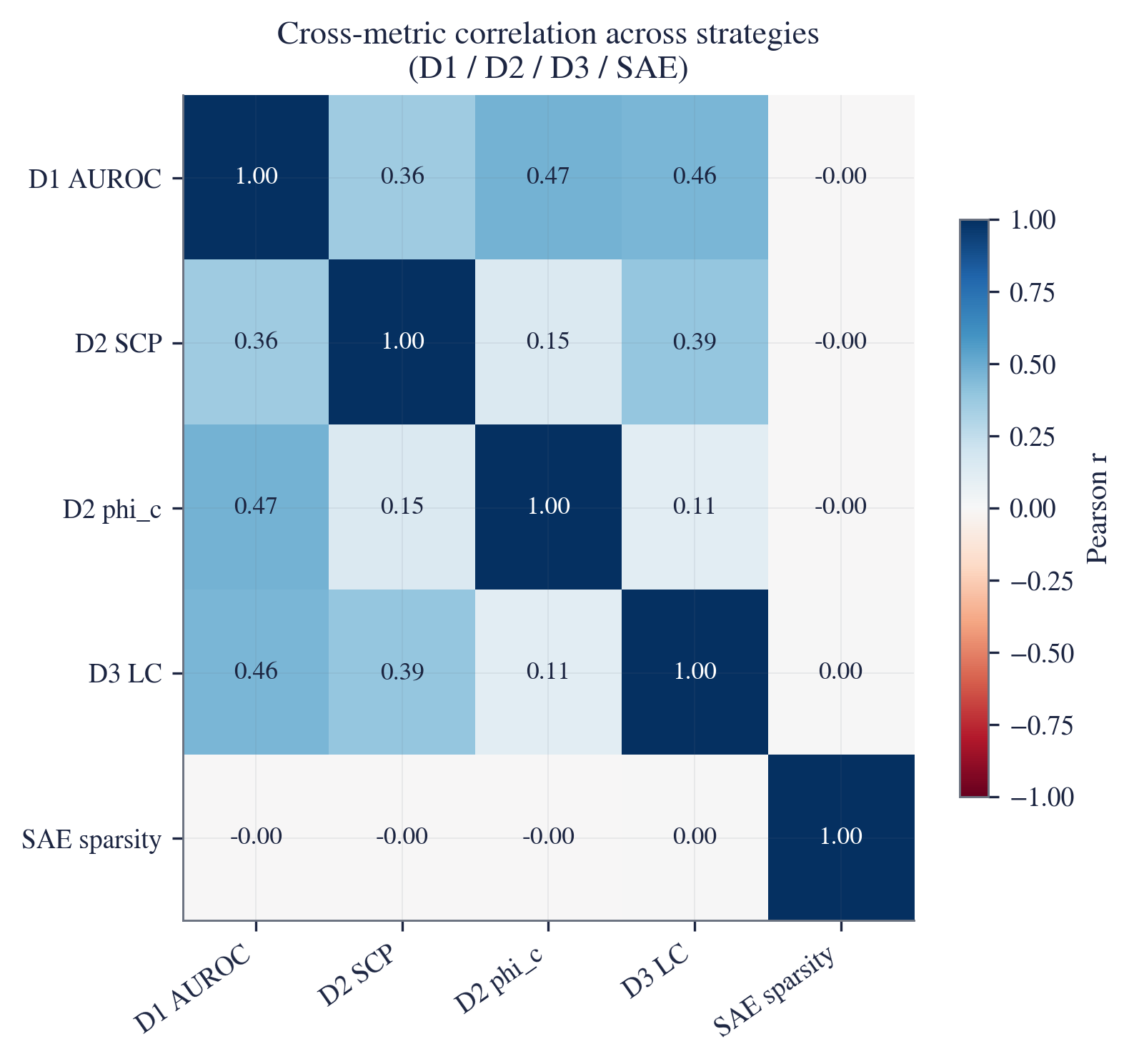}
  \caption{Cross-metric correlation matrix and scatter plots for D1 (mean AUROC),
  D2 (mean SCP), and D3 (mean gated ratio).
  D1 and D2 are weakly positively correlated, but the correlation is driven by the
  easy-concept mass rather than meaningful covariation in the harder regime.
  D1 and D3, and D2 and D3, show even weaker correlation, confirming that
  disentanglement cannot be predicted from detection performance alone.
  No single summary axis captures all three dimensions of strategy quality.}
  \label{fig:crossmetric}
\end{figure}

\begin{figure}[H]
  \centering
  \includegraphics[width=0.78\linewidth]{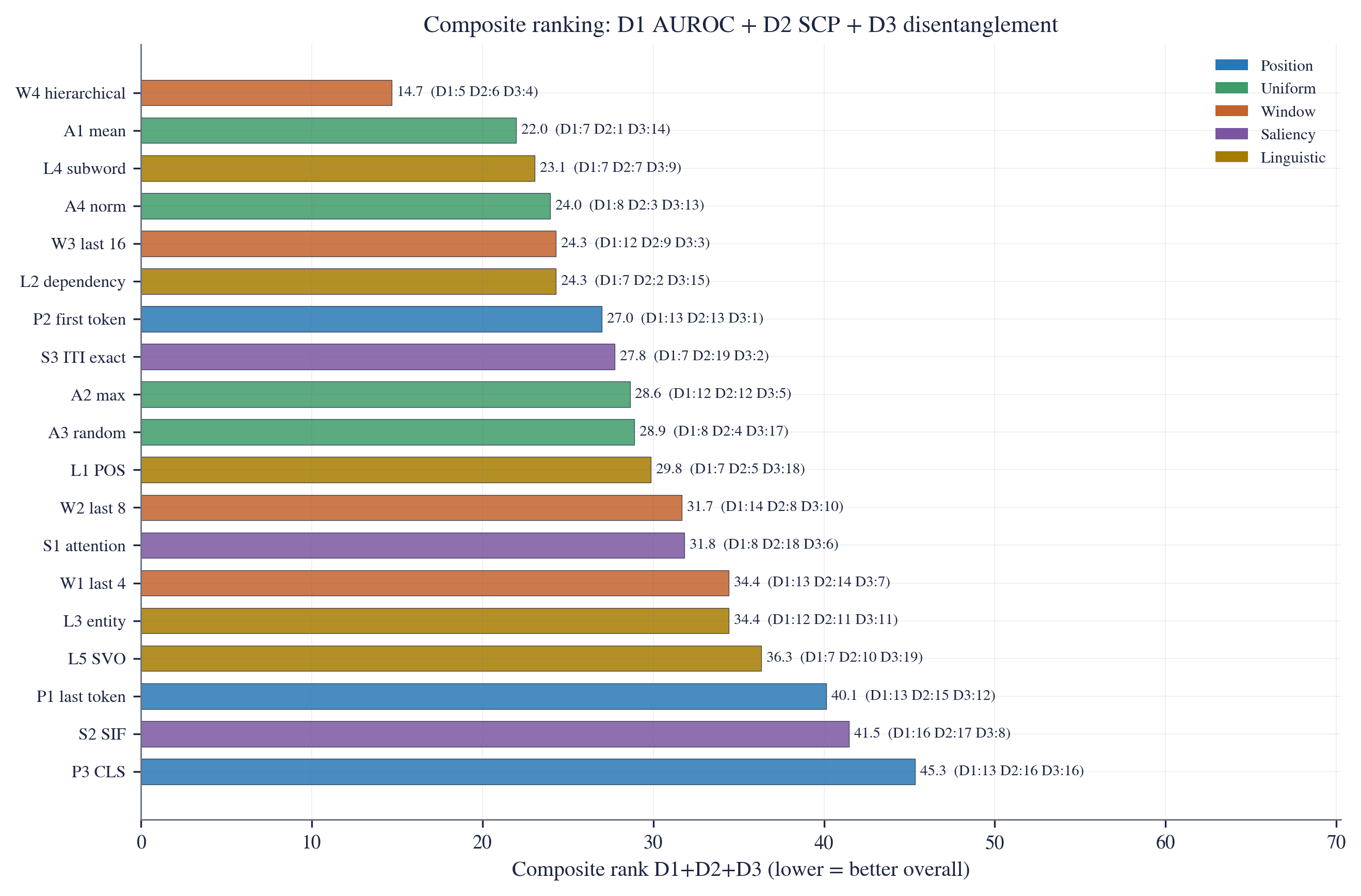}
  \caption{Composite rank bar: sum of D1, D2, and D3 ranks per strategy
  (lower total = better across all three axes).
  This is an informal summary; D1 remains the primary benchmark ranking.
  \texttt{A1\_mean} and \texttt{A4\_norm} achieve the best composite ranks, driven
  by consistency across D2 and D3 even when they are not always first on D1.
  \texttt{W4\_hierarchical} ranks first on D1 but falls in the composite due to
  weaker D2/D3 performance---a concrete illustration of the practitioner tradeoff
  between detection power and downstream steerability.}
  \label{fig:compositerank}
\end{figure}


\section{Descriptive Concept Families}
\label{app:families}

The family labels are descriptive metadata only; they carry no routing
implications for the benchmark leaderboard and are included as navigation aids.

\begin{table}[t]
\centering
\caption{Descriptive concept-family summary used for appendix navigation only.}
\label{tab:families}
\small
\begin{tabular}{p{2.2cm}p{5.6cm}p{4.5cm}}
\toprule
Family & Concepts & Intended signal profile \\
\midrule
Sparse-lexical & \texttt{hedging}, \texttt{legal\_formality}, \texttt{frustration}, \texttt{numerical\_precision} & A small number of local markers or numeric tokens dominate the signal. \\
Dense-lexical & \texttt{imdb\_sentiment}, \texttt{toxicity}, \texttt{depression} & Signal is spread across many tokens and document-level tone cues. \\
Syntactic & \texttt{causation}, \texttt{contrast}, \texttt{conditionality}, \texttt{negation\_density} & The concept depends on connective structure, clause relations, or syntactic operators. \\
Register & \texttt{academic\_tone}, \texttt{code\_docs}, \texttt{bureaucratic} & Distributed stylistic cues define a discourse register. \\
Semantic-abstract & \texttt{narrative}, \texttt{deference}, \texttt{planning} & High-level discourse intent or stance distributed over the passage. \\
\bottomrule
\end{tabular}
\end{table}


\section{Compute}
\label{app:compute}

The full \poolbench{} sweep ran on H100 80\,GB GPUs.
Activation extraction for all three models at three candidate layers over
37{,}693 passages required approximately 18 GPU-hours.
D1 AUROC evaluation (logistic probe training, 5-fold cross-validation, 19
strategies $\times$ 17 concepts $\times$ 3 models) required approximately 4
GPU-hours.
D2 SCP generation (12 steering coefficients $\times$ 10 prompts per
strategy--concept pair in a single batched generation call) required approximately
22 GPU-hours.
The total sweep cost was approximately 44 GPU-hours.

The released pre-extracted activations allow D1--D3 evaluation without rerunning
extraction.
Extending the leaderboard to a new pooling strategy on a fixed model requires only
re-running D1 probe training ($\approx 1$--2 GPU-hours) and optionally D2 SCP
generation ($\approx 6$--7 GPU-hours).
Extending to a new model requires a fresh activation extraction pass
($\approx 6$--8 GPU-hours) before D1--D3 evaluation.


\section{Artifact Release: Permanent URLs}
\label{app:artifacts}

Table~\ref{tab:artifacts} lists the permanent release URLs for all benchmark
artifacts.
These are the primary entry points for reproduction and extension.
The benchmark corpus, pre-extracted activations for all three models at all
candidate layers, fine-tuned BERT scorer models for 15 concepts, steering vectors,
and the evaluation code are all available at the URLs below.
\texttt{toxicity} and \texttt{depression} scorer outputs are excluded from the
public release per NeurIPS data-ethics guidance.

\begin{table}[t]
\centering
\caption{Released \poolbench{} artifacts.}
\label{tab:artifacts}
\small
\begin{tabular}{p{3.2cm}p{9.2cm}}
\toprule
Artifact & URL \\
\midrule
Corpus dataset & \url{https://huggingface.co/datasets/nips234678/poolbench} \\
Activation release & \url{https://huggingface.co/datasets/nips234678/poolbench-activations} \\
Scorer models & \url{https://huggingface.co/nips234678/poolbench-bert-scorers} \\
Steering vectors & \url{https://huggingface.co/datasets/nips234678/poolbench-steering-vectors} \\
Evaluation code & \url{https://github.com/ayushi-agarwal/poolbench} \\
Statistical results & \url{https://github.com/ayushi-agarwal/poolbench/results} \\
\bottomrule
\end{tabular}
\end{table}

\section{Full Per-Concept Per-Strategy Results}
\label{app:fullresults}

Table~\ref{tab:fullresults} reports D1 AUROC, D2 SCP$_c$, D3~LD, and D3~LC
for every model $\times$ concept $\times$ strategy combination at the released best layer.
Strategies follow family order: Position (P1--P3), Uniform (A1--A4), Window (W1--W4),
Saliency (S1--S3), Linguistic (L1--L5); the cross-strategy mean is shown in italics.
\textdagger\ \texttt{L2\_dependency\_rel} is degenerate (see main text); values retained for completeness.

\begingroup
\scriptsize
\setlength{\tabcolsep}{4pt}
\setlength{\LTleft}{0pt}\setlength{\LTright}{0pt}
\begin{longtable}[l]{@{}p{1.9cm}p{2.9cm}p{3.2cm}rrrrrr@{}}
\caption{Full per-concept per-strategy leaderboard for all three models.
D1\,=\,AUROC; D2\,SCP\,=\,signed confidence percentile; D2\,M\,=\,mean SCP;
D2\,$\phi$\,=\,phi coefficient; D3\,LD\,=\,linear disentanglement; D3\,LC\,=\,linear consistency.
Cells marked \texttt{---} indicate the strategy produced a zero-norm steering vector.}\\
\label{tab:fullresults} \\
\toprule
\textbf{Model} & \textbf{Concept} & \textbf{Strategy} & \textbf{D1} & \textbf{D2 SCP} & \textbf{D2 M} & \textbf{D2 $\phi$} & \textbf{D3 LD} & \textbf{D3 LC} \\
\midrule
\endfirsthead
\multicolumn{9}{c}{{\tablename\ \thetable{} -- continued from previous page}} \\
\toprule
\textbf{Model} & \textbf{Concept} & \textbf{Strategy} & \textbf{D1} & \textbf{D2 SCP} & \textbf{D2 M} & \textbf{D2 $\phi$} & \textbf{D3 LD} & \textbf{D3 LC} \\
\midrule
\endhead
\midrule
\multicolumn{9}{r}{\small\textit{continued on next page}} \\
\endfoot
\bottomrule
\endlastfoot
\multicolumn{9}{l}{\textbf{Llama-3.1-8B}} \\
\midrule
Llama-3.1-8B & Academic Tone & P1 last token & 0.9970 & 0.0514 & -0.7203 & 1.0286 & 0.9997 & 0.9684 \\
Llama-3.1-8B & Academic Tone & P2 first token & 0.4893 & -0.0543 & 0.1399 & 0.9770 & 1.1300 & 0.8629 \\
Llama-3.1-8B & Academic Tone & P3 CLS & 0.8387 & 0.0674 & -0.6993 & 1.0491 & 1.0220 & 0.9619 \\
Llama-3.1-8B & Academic Tone & A1 mean & 0.9998 & 0.0438 & -0.6224 & 1.0235 & 0.9853 & 1.0343 \\
Llama-3.1-8B & Academic Tone & A2 max & 0.8840 & -0.0218 & -0.5524 & 0.9600 & 0.9745 & 0.8487 \\
Llama-3.1-8B & Academic Tone & A3 random & 0.9998 & 0.0414 & -0.7063 & 1.0386 & 0.9845 & 1.0363 \\
Llama-3.1-8B & Academic Tone & A4 norm & 0.9999 & 0.0031 & -0.4825 & 1.1455 & --- & --- \\
\end{longtable}
\endgroup


\FloatBarrier
\clearpage
\clearpage
\section{Additional Released Leaderboards and Diagnostics}
\label{app:released_matrices}
The benchmark release includes full model$\times$concept$\times$strategy AUROC matrices at each best layer. The tables below reproduce those released best-layer matrices in abbreviated strategy notation to keep the appendix navigable.
\begin{table}[!htbp]
\centering
\caption{Cross-model concept-by-concept gains of the best released strategy over the last-token and mean-pooling baselines. Gains are averaged over the three benchmark models.}
\label{tab:concept_gains}
\small
\begin{tabular}{lcccc}
\toprule
Concept & Best strategy & Best AUROC & Gain vs. \texttt{P1\_last\_token} & Gain vs. \texttt{A1\_mean} \\
\midrule
\texttt{academic\_tone} & \texttt{W4\_hierarchical} & 1.000 & +0.003 & +0.000 \\
\texttt{bureaucratic} & \texttt{W4\_hierarchical} & 1.000 & +0.003 & +0.000 \\
\texttt{causation} & \texttt{P2\_first\_token} & 0.494 & +0.115 & +0.099 \\
\texttt{code\_docs} & \texttt{L5\_SVO} & 1.000 & +0.001 & +0.000 \\
\texttt{conditionality} & \texttt{W1\_mean\_last\_4} & 0.788 & +0.056 & +0.373 \\
\texttt{contrast} & \texttt{P2\_first\_token} & 0.503 & +0.105 & +0.090 \\
\texttt{deference} & \texttt{L4\_subword\_root} & 0.906 & +0.042 & +0.000 \\
\texttt{depression} & \texttt{L5\_SVO} & 1.000 & +0.005 & +0.000 \\
\texttt{frustration} & \texttt{W4\_hierarchical} & 0.885 & +0.030 & +0.006 \\
\texttt{hedging} & \texttt{P2\_first\_token} & 0.503 & +0.128 & +0.086 \\
\texttt{imdb\_sentiment} & \texttt{W4\_hierarchical} & 0.939 & +0.010 & +0.019 \\
\texttt{legal\_formality} & \texttt{P2\_first\_token} & 0.502 & +0.182 & +0.081 \\
\texttt{narrative} & \texttt{L1\_POS\_filtered} & 1.000 & +0.014 & +0.000 \\
\texttt{negation\_density} & \texttt{P2\_first\_token} & 0.499 & +0.130 & +0.050 \\
\texttt{numerical\_precision} & \texttt{L3\_named\_entity} & 0.977 & +0.025 & +0.003 \\
\texttt{planning} & \texttt{P3\_CLS} & 1.000 & +0.005 & +0.000 \\
\texttt{toxicity} & \texttt{S2\_SIF} & 0.926 & +0.078 & +0.021 \\
\bottomrule
\end{tabular}
\end{table}
\begin{table}[!htbp]
\centering
\caption{Supervision gap requested by the methodology. The gap is the D1 AUROC of \texttt{S3\_ITI\_exact} minus the strongest unsupervised strategy on the same model.}
\label{tab:supervision_gap}
\small
\begin{tabular}{lcccc}
\toprule
Model & Best unsupervised strategy & Unsupervised AUROC & \texttt{S3\_ITI\_exact} AUROC & Gap \\
\midrule
Llama-3.1-8B & \texttt{W4\_hierarchical} & 0.780 & 0.777 & -0.003 \\
Gemma-2-9B & \texttt{W4\_hierarchical} & 0.784 & 0.765 & -0.019 \\
Mistral-7B & \texttt{W4\_hierarchical} & 0.776 & 0.776 & +0.000 \\
\bottomrule
\end{tabular}
\end{table}
\begin{table}[!htbp]
\centering
\caption{Fallback rates for linguistic strategies from the released parser logs. \texttt{L2\_dependency\_rel} is fully degenerate in the current release: it falls back to mean pooling on every passage for every model.}
\label{tab:fallback_rates}
\small
\begin{tabular}{lccccc}
\toprule
Model & \texttt{L1} & \texttt{L2} & \texttt{L3} & \texttt{L4} & \texttt{L5} \\
\midrule
Llama-3.1-8B & 0.02\% & 100.00\% & 4.36\% & 0.00\% & 0.00\% \\
Gemma-2-9B & 0.02\% & 100.00\% & 4.36\% & 0.00\% & 0.00\% \\
Mistral-7B & 0.02\% & 100.00\% & 4.36\% & 0.00\% & 0.00\% \\
\bottomrule
\end{tabular}
\end{table}
\clearpage
\begin{table}[p]
\centering
\caption{Full released D1 matrix for Llama-3.1-8B at best layer L31. Columns use the abbreviated strategy IDs defined in the main text.}
\label{tab:master_1}
\scriptsize
\resizebox{\linewidth}{!}{%
\begin{tabular}{lccccccccccccccccccc}
\toprule
Concept & P1 & P2 & P3 & A1 & A2 & A3 & A4 & W1 & W2 & W3 & W4 & S1 & S2 & S3 & L1 & L2 & L3 & L4 & L5 \\
\midrule
\texttt{academic\_tone} & 0.997 & 0.489 & 0.839 & 1.000 & 0.884 & 1.000 & 1.000 & 1.000 & 1.000 & 1.000 & 1.000 & 1.000 & 0.515 & 1.000 & 1.000 & 1.000 & 0.999 & 1.000 & 1.000 \\
\texttt{bureaucratic} & 0.997 & 0.501 & 0.867 & 1.000 & 0.966 & 1.000 & 1.000 & 1.000 & 1.000 & 1.000 & 1.000 & 1.000 & 0.613 & 1.000 & 1.000 & 1.000 & 1.000 & 1.000 & 1.000 \\
\texttt{causation} & 0.418 & 0.499 & 0.396 & 0.408 & 0.447 & 0.412 & 0.406 & 0.378 & 0.378 & 0.401 & 0.433 & 0.453 & 0.411 & 0.460 & 0.425 & 0.408 & 0.392 & 0.416 & 0.431 \\
\texttt{code\_docs} & 1.000 & 0.500 & 0.925 & 0.999 & 0.973 & 0.999 & 0.999 & 0.999 & 0.999 & 0.999 & 1.000 & 0.999 & 0.593 & 0.999 & 0.999 & 0.999 & 0.995 & 1.000 & 1.000 \\
\texttt{conditionality} & 0.693 & 0.500 & 0.668 & 0.416 & 0.465 & 0.417 & 0.415 & 0.747 & 0.693 & 0.592 & 0.492 & 0.425 & 0.401 & 0.432 & 0.417 & 0.416 & 0.407 & 0.415 & 0.425 \\
\texttt{contrast} & 0.508 & 0.496 & 0.398 & 0.433 & 0.448 & 0.433 & 0.430 & 0.406 & 0.411 & 0.426 & 0.465 & 0.461 & 0.448 & 0.467 & 0.443 & 0.433 & 0.422 & 0.435 & 0.451 \\
\texttt{deference} & 0.867 & 0.500 & 0.630 & 0.889 & 0.872 & 0.878 & 0.888 & 0.839 & 0.839 & 0.853 & 0.885 & 0.888 & 0.686 & 0.888 & 0.879 & 0.889 & 0.840 & 0.884 & 0.882 \\
\texttt{depression} & 0.998 & 0.502 & 0.987 & 1.000 & 0.999 & 1.000 & 1.000 & 1.000 & 1.000 & 1.000 & 1.000 & 1.000 & 0.591 & 1.000 & 1.000 & 1.000 & 0.999 & 1.000 & 1.000 \\
\texttt{frustration} & 0.870 & 0.500 & 0.652 & 0.874 & 0.809 & 0.872 & 0.874 & 0.875 & 0.877 & 0.885 & 0.877 & 0.860 & 0.828 & 0.863 & 0.866 & 0.874 & 0.834 & 0.872 & 0.870 \\
\texttt{hedging} & 0.391 & 0.507 & 0.395 & 0.416 & 0.442 & 0.415 & 0.417 & 0.359 & 0.375 & 0.394 & 0.441 & 0.430 & 0.390 & 0.434 & 0.424 & 0.416 & 0.405 & 0.420 & 0.421 \\
\texttt{imdb\_sentiment} & 0.934 & 0.500 & 0.566 & 0.878 & 0.839 & 0.877 & 0.877 & 0.911 & 0.894 & 0.893 & 0.898 & 0.856 & 0.851 & 0.858 & 0.874 & 0.878 & 0.783 & 0.876 & 0.883 \\
\texttt{legal\_formality} & 0.338 & 0.508 & 0.401 & 0.424 & 0.447 & 0.424 & 0.425 & 0.373 & 0.384 & 0.399 & 0.445 & 0.466 & 0.429 & 0.472 & 0.435 & 0.424 & 0.403 & 0.424 & 0.431 \\
\texttt{narrative} & 0.987 & 0.500 & 0.662 & 1.000 & 0.954 & 1.000 & 1.000 & 0.997 & 0.998 & 0.999 & 1.000 & 1.000 & 0.579 & 1.000 & 1.000 & 1.000 & 0.998 & 1.000 & 1.000 \\
\texttt{negation\_density} & 0.411 & 0.501 & 0.400 & 0.426 & 0.450 & 0.422 & 0.429 & 0.372 & 0.391 & 0.412 & 0.443 & 0.465 & 0.428 & 0.472 & 0.438 & 0.426 & 0.399 & 0.432 & 0.439 \\
\texttt{numerical\_precision} & 0.963 & 0.500 & 0.697 & 0.977 & 0.993 & 0.977 & 0.974 & 0.962 & 0.963 & 0.964 & 0.975 & 0.977 & 0.742 & 0.976 & 0.967 & 0.977 & 0.977 & 0.974 & 0.965 \\
\texttt{planning} & 1.000 & 0.500 & 1.000 & 1.000 & 0.995 & 1.000 & 1.000 & 1.000 & 1.000 & 1.000 & 1.000 & 1.000 & 0.657 & 1.000 & 1.000 & 1.000 & 1.000 & 1.000 & 1.000 \\
\texttt{toxicity} & 0.834 & 0.501 & 0.776 & 0.905 & 0.775 & 0.895 & 0.915 & 0.846 & 0.846 & 0.854 & 0.909 & 0.891 & 0.904 & 0.891 & 0.858 & 0.905 & 0.780 & 0.883 & 0.848 \\
\bottomrule
\end{tabular}%
}
\end{table}
\clearpage
\begin{table}[p]
\centering
\caption{Full released D1 matrix for Gemma-2-9B at best layer L28. Columns use the abbreviated strategy IDs defined in the main text.}
\label{tab:master_2}
\scriptsize
\resizebox{\linewidth}{!}{%
\begin{tabular}{lccccccccccccccccccc}
\toprule
Concept & P1 & P2 & P3 & A1 & A2 & A3 & A4 & W1 & W2 & W3 & W4 & S1 & S2 & S3 & L1 & L2 & L3 & L4 & L5 \\
\midrule
\texttt{academic\_tone} & 0.997 & 0.498 & 0.845 & 1.000 & 0.999 & 1.000 & 1.000 & 0.999 & 0.999 & 0.999 & 1.000 & 1.000 & 0.601 & 1.000 & 1.000 & 1.000 & 0.998 & 1.000 & 1.000 \\
\texttt{bureaucratic} & 0.997 & 0.490 & 0.858 & 1.000 & 1.000 & 1.000 & 1.000 & 0.999 & 0.999 & 0.999 & 1.000 & 1.000 & 0.563 & 1.000 & 1.000 & 1.000 & 1.000 & 1.000 & 1.000 \\
\texttt{causation} & 0.395 & 0.491 & 0.388 & 0.415 & 0.330 & 0.415 & 0.413 & 0.377 & 0.338 & 0.364 & 0.419 & 0.410 & 0.406 & 0.410 & 0.417 & 0.415 & 0.381 & 0.409 & 0.420 \\
\texttt{code\_docs} & 0.998 & 0.500 & 0.916 & 1.000 & 1.000 & 1.000 & 0.999 & 0.994 & 0.993 & 0.995 & 1.000 & 0.999 & 0.631 & 0.999 & 1.000 & 1.000 & 0.995 & 1.000 & 1.000 \\
\texttt{conditionality} & 0.753 & 0.500 & 0.664 & 0.421 & 0.391 & 0.420 & 0.420 & 0.811 & 0.784 & 0.714 & 0.534 & 0.425 & 0.410 & 0.425 & 0.423 & 0.421 & 0.360 & 0.424 & 0.428 \\
\texttt{contrast} & 0.378 & 0.503 & 0.406 & 0.425 & 0.429 & 0.424 & 0.422 & 0.366 & 0.395 & 0.411 & 0.433 & 0.427 & 0.419 & 0.427 & 0.423 & 0.425 & 0.396 & 0.421 & 0.425 \\
\texttt{deference} & 0.865 & 0.500 & 0.631 & 0.899 & 0.894 & 0.876 & 0.890 & 0.830 & 0.826 & 0.840 & 0.882 & 0.900 & 0.641 & 0.900 & 0.888 & 0.899 & 0.846 & 0.899 & 0.888 \\
\texttt{depression} & 0.992 & 0.501 & 0.989 & 1.000 & 0.998 & 1.000 & 1.000 & 0.997 & 0.997 & 0.997 & 1.000 & 0.999 & 0.638 & 0.999 & 1.000 & 1.000 & 0.999 & 1.000 & 1.000 \\
\texttt{frustration} & 0.846 & 0.500 & 0.672 & 0.877 & 0.849 & 0.876 & 0.876 & 0.858 & 0.860 & 0.872 & 0.880 & 0.846 & 0.778 & 0.846 & 0.870 & 0.877 & 0.837 & 0.875 & 0.877 \\
\texttt{hedging} & 0.383 & 0.496 & 0.403 & 0.426 & 0.358 & 0.425 & 0.425 & 0.367 & 0.379 & 0.382 & 0.427 & 0.430 & 0.413 & 0.430 & 0.428 & 0.426 & 0.424 & 0.427 & 0.424 \\
\texttt{imdb\_sentiment} & 0.905 & 0.500 & 0.586 & 0.922 & 0.887 & 0.921 & 0.916 & 0.911 & 0.911 & 0.922 & 0.946 & 0.863 & 0.879 & 0.863 & 0.920 & 0.922 & 0.766 & 0.918 & 0.931 \\
\texttt{legal\_formality} & 0.345 & 0.496 & 0.385 & 0.426 & 0.392 & 0.428 & 0.423 & 0.380 & 0.378 & 0.398 & 0.433 & 0.444 & 0.437 & 0.444 & 0.441 & 0.426 & 0.399 & 0.431 & 0.433 \\
\texttt{narrative} & 0.979 & 0.500 & 0.689 & 1.000 & 0.999 & 1.000 & 0.999 & 0.991 & 0.994 & 0.996 & 0.999 & 0.998 & 0.660 & 0.998 & 1.000 & 1.000 & 0.998 & 0.999 & 0.999 \\
\texttt{negation\_density} & 0.369 & 0.492 & 0.410 & 0.472 & 0.410 & 0.471 & 0.470 & 0.388 & 0.418 & 0.446 & 0.480 & 0.455 & 0.488 & 0.455 & 0.482 & 0.472 & 0.411 & 0.479 & 0.470 \\
\texttt{numerical\_precision} & 0.947 & 0.500 & 0.684 & 0.972 & 0.968 & 0.971 & 0.974 & 0.960 & 0.961 & 0.962 & 0.970 & 0.973 & 0.511 & 0.973 & 0.966 & 0.972 & 0.975 & 0.968 & 0.965 \\
\texttt{planning} & 0.992 & 0.500 & 1.000 & 1.000 & 1.000 & 1.000 & 1.000 & 0.999 & 0.997 & 0.998 & 1.000 & 0.999 & 0.992 & 0.999 & 1.000 & 1.000 & 0.998 & 1.000 & 1.000 \\
\texttt{toxicity} & 0.851 & 0.501 & 0.811 & 0.923 & 0.891 & 0.895 & 0.919 & 0.844 & 0.856 & 0.884 & 0.919 & 0.838 & 0.934 & 0.838 & 0.886 & 0.923 & 0.791 & 0.912 & 0.869 \\
\bottomrule
\end{tabular}%
}
\end{table}
\clearpage
\begin{table}[p]
\centering
\caption{Full released D1 matrix for Mistral-7B at best layer L16. Columns use the abbreviated strategy IDs defined in the main text.}
\label{tab:master_3}
\scriptsize
\resizebox{\linewidth}{!}{%
\begin{tabular}{lccccccccccccccccccc}
\toprule
Concept & P1 & P2 & P3 & A1 & A2 & A3 & A4 & W1 & W2 & W3 & W4 & S1 & S2 & S3 & L1 & L2 & L3 & L4 & L5 \\
\midrule
\texttt{academic\_tone} & 0.995 & 0.503 & 0.822 & 1.000 & 0.997 & 1.000 & 1.000 & 0.998 & 0.998 & 0.999 & 1.000 & 1.000 & 0.671 & 1.000 & 1.000 & 1.000 & 0.999 & 1.000 & 1.000 \\
\texttt{bureaucratic} & 0.997 & 0.502 & 0.858 & 1.000 & 1.000 & 1.000 & 1.000 & 0.999 & 0.999 & 0.999 & 1.000 & 1.000 & 0.646 & 1.000 & 1.000 & 1.000 & 1.000 & 1.000 & 1.000 \\
\texttt{causation} & 0.324 & 0.491 & 0.370 & 0.360 & 0.293 & 0.357 & 0.360 & 0.241 & 0.233 & 0.266 & 0.378 & 0.381 & 0.331 & 0.422 & 0.348 & 0.360 & 0.345 & 0.361 & 0.347 \\
\texttt{code\_docs} & 0.999 & 0.504 & 0.896 & 1.000 & 0.999 & 1.000 & 1.000 & 0.999 & 0.999 & 0.999 & 1.000 & 1.000 & 0.630 & 1.000 & 1.000 & 1.000 & 0.998 & 1.000 & 1.000 \\
\texttt{conditionality} & 0.748 & 0.500 & 0.665 & 0.406 & 0.437 & 0.409 & 0.400 & 0.804 & 0.782 & 0.736 & 0.527 & 0.435 & 0.397 & 0.441 & 0.394 & 0.406 & 0.364 & 0.416 & 0.399 \\
\texttt{contrast} & 0.307 & 0.509 & 0.385 & 0.379 & 0.303 & 0.410 & 0.379 & 0.245 & 0.246 & 0.284 & 0.382 & 0.416 & 0.340 & 0.410 & 0.368 & 0.379 & 0.370 & 0.378 & 0.376 \\
\texttt{deference} & 0.860 & 0.582 & 0.630 & 0.929 & 0.913 & 0.818 & 0.885 & 0.830 & 0.824 & 0.836 & 0.903 & 0.900 & 0.662 & 0.904 & 0.887 & 0.929 & 0.824 & 0.935 & 0.894 \\
\texttt{depression} & 0.996 & 0.503 & 0.988 & 0.999 & 0.997 & 0.998 & 0.999 & 0.997 & 0.996 & 0.997 & 0.999 & 1.000 & 0.751 & 1.000 & 0.999 & 0.999 & 0.997 & 0.999 & 1.000 \\
\texttt{frustration} & 0.847 & 0.499 & 0.666 & 0.887 & 0.868 & 0.875 & 0.888 & 0.868 & 0.872 & 0.883 & 0.897 & 0.870 & 0.890 & 0.872 & 0.888 & 0.887 & 0.843 & 0.885 & 0.888 \\
\texttt{hedging} & 0.351 & 0.505 & 0.397 & 0.408 & 0.329 & 0.414 & 0.405 & 0.282 & 0.307 & 0.327 & 0.410 & 0.408 & 0.370 & 0.433 & 0.400 & 0.408 & 0.385 & 0.409 & 0.389 \\
\texttt{imdb\_sentiment} & 0.947 & 0.500 & 0.584 & 0.959 & 0.938 & 0.955 & 0.956 & 0.942 & 0.941 & 0.949 & 0.973 & 0.940 & 0.881 & 0.947 & 0.962 & 0.959 & 0.819 & 0.959 & 0.968 \\
\texttt{legal\_formality} & 0.277 & 0.502 & 0.396 & 0.413 & 0.353 & 0.423 & 0.412 & 0.300 & 0.309 & 0.337 & 0.400 & 0.435 & 0.428 & 0.432 & 0.437 & 0.413 & 0.380 & 0.413 & 0.408 \\
\texttt{narrative} & 0.990 & 0.500 & 0.681 & 1.000 & 0.999 & 1.000 & 1.000 & 0.994 & 0.996 & 0.998 & 1.000 & 1.000 & 0.730 & 1.000 & 1.000 & 1.000 & 0.999 & 1.000 & 1.000 \\
\texttt{negation\_density} & 0.326 & 0.503 & 0.408 & 0.447 & 0.395 & 0.442 & 0.445 & 0.289 & 0.322 & 0.353 & 0.447 & 0.439 & 0.449 & 0.455 & 0.449 & 0.447 & 0.376 & 0.460 & 0.440 \\
\texttt{numerical\_precision} & 0.947 & 0.503 & 0.680 & 0.974 & 0.952 & 0.970 & 0.976 & 0.962 & 0.961 & 0.961 & 0.973 & 0.974 & 0.779 & 0.972 & 0.970 & 0.974 & 0.980 & 0.970 & 0.968 \\
\texttt{planning} & 0.994 & 0.500 & 1.000 & 1.000 & 1.000 & 1.000 & 1.000 & 1.000 & 1.000 & 1.000 & 1.000 & 1.000 & 0.530 & 1.000 & 1.000 & 1.000 & 0.999 & 1.000 & 1.000 \\
\texttt{toxicity} & 0.859 & 0.515 & 0.805 & 0.887 & 0.796 & 0.795 & 0.933 & 0.859 & 0.869 & 0.889 & 0.899 & 0.910 & 0.939 & 0.905 & 0.920 & 0.887 & 0.742 & 0.831 & 0.904 \\
\bottomrule
\end{tabular}%
}
\end{table}

\FloatBarrier


\begin{thebibliography}{99}

\bibitem[Arora et al.(2017)]{arora2017simple}
Sanjeev Arora, Yingyu Liang, and Tengyu Ma.
\newblock A simple but tough-to-beat baseline for sentence embeddings.
\newblock In \emph{ICLR}, 2017.

\bibitem[Burns et al.(2023)]{burns2023discovering}
Collin Burns, Haotian Ye, Dan Klein, and Jacob Steinhardt.
\newblock Discovering latent knowledge in language models without supervision.
\newblock In \emph{ICLR}, 2023.

\bibitem[Demsar(2006)]{demsar2006statistical}
Janez Demsar.
\newblock Statistical comparisons of classifiers over multiple data sets.
\newblock \emph{Journal of Machine Learning Research}, 7:1--30, 2006.

\bibitem[Devlin et al.(2019)]{devlin2019bert}
Jacob Devlin, Ming-Wei Chang, Kenton Lee, and Kristina Toutanova.
\newblock BERT: Pre-training of deep bidirectional transformers for language
  understanding.
\newblock In \emph{NAACL-HLT}, 2019.

\bibitem[Dubey et al.(2024)]{dubey2024llama3}
Abhimanyu Dubey et al.
\newblock The Llama 3 herd of models.
\newblock \emph{arXiv:2407.21783}, 2024.

\bibitem[Efron and Tibshirani(1993)]{efron1993bootstrap}
Bradley Efron and Robert Tibshirani.
\newblock \emph{An Introduction to the Bootstrap}.
\newblock Chapman and Hall, 1993.

\bibitem[Engels et al.(2024)]{engels2024not}
Joshua Engels, Isaac Liao, Eric J. Michaud, Wes Gurnee, and Max Tegmark.
\newblock Not all language model features are linear.
\newblock \emph{arXiv:2405.14860}, 2024.

\bibitem[Friedman(1940)]{friedman1940rankings}
Milton Friedman.
\newblock A comparison of alternative tests of significance for the problem of
  $m$ rankings.
\newblock \emph{Annals of Mathematical Statistics}, 11(1):86--92, 1940.

\bibitem[Gebru et al.(2018)]{gebru2018datasheets}
Timnit Gebru et al.
\newblock Datasheets for datasets.
\newblock \emph{arXiv:1803.09010}, 2018.

\bibitem[Hanley and McNeil(1982)]{hanley1982roc}
James Hanley and Barbara McNeil.
\newblock The meaning and use of the area under a receiver operating
  characteristic curve.
\newblock \emph{Radiology}, 143(1):29--36, 1982.

\bibitem[Honnibal et al.(2020)]{honnibal2020spacy}
Matthew Honnibal, Ines Montani, Sofie Van Landeghem, and Adriane Boyd.
\newblock spaCy: Industrial-strength natural language processing in Python.
\newblock 2020.

\bibitem[Izacard et al.(2022)]{izacard2022contriever}
Gautier Izacard, Mathilde Caron, Lucas Hosseini, Sebastian Riedel,
  Piotr Bojanowski, Armand Joulin, and Edouard Grave.
\newblock Unsupervised dense information retrieval with contrastive learning.
\newblock \emph{Transactions on Machine Learning Research}, 2022.

\bibitem[Jiang et al.(2023)]{jiang2023mistral}
Albert Jiang et al.
\newblock Mistral 7B.
\newblock \emph{arXiv:2310.06825}, 2023.

\bibitem[Kohavi(1995)]{kohavi1995cv}
Ron Kohavi.
\newblock A study of cross-validation and bootstrap for accuracy estimation and
  model selection.
\newblock In \emph{IJCAI}, 1995.

\bibitem[Li et al.(2023)]{li2023iti}
Kenneth Li, Oam Patel, Fernanda Viegas, Hanspeter Pfister, and Martin Wattenberg.
\newblock Inference-time intervention: Eliciting truthful answers from a language
  model.
\newblock In \emph{NeurIPS}, 2023.

\bibitem[Liang et al.(2022)]{liang2022holistic}
Percy Liang et al.
\newblock HELM: Holistic evaluation of language models.
\newblock \emph{arXiv:2211.09110}, 2022.

\bibitem[Marks and Tegmark(2023)]{marks2023geometry}
Samuel Marks and Max Tegmark.
\newblock The geometry of truth: Emergent linear structure in large language
  model representations of true/false datasets.
\newblock \emph{arXiv:2310.06824}, 2023.

\bibitem[Mesnard et al.(2024)]{mesnard2024gemma}
Thomas Mesnard et al.
\newblock Gemma: Open models based on Gemini research and technology.
\newblock \emph{arXiv:2403.08295}, 2024.

\bibitem[Rimsky et al.(2023)]{rimsky2023steering}
Nina Rimsky, Nick Gabrieli, Julian Schulz, Meg Tong, Evan Hubinger, and
  Alexander Matt Turner.
\newblock Steering Llama 2 via contrastive activation addition.
\newblock \emph{arXiv:2312.06681}, 2023.

\bibitem[Simonyan et al.(2014)]{simonyan2014saliency}
Karen Simonyan, Andrea Vedaldi, and Andrew Zisserman.
\newblock Deep inside convolutional networks: Visualising image classification
  models and saliency maps.
\newblock In \emph{ICLR Workshop}, 2014.

\bibitem[Srivastava et al.(2022)]{srivastava2022bigbench}
Aarohi Srivastava et al.
\newblock Beyond the imitation game: Quantifying and extrapolating the
  capabilities of language models.
\newblock \emph{arXiv:2206.04615}, 2022.

\bibitem[Turner et al.(2023)]{turner2023activation}
Alexander Matt Turner, Lisa Thiergart, David Udell, Gavin Leech, Ulisse Mini,
  and Monte MacDiarmid.
\newblock Activation addition: Steering language models without optimization.
\newblock \emph{arXiv:2308.10248}, 2023.

\bibitem[Zou et al.(2023)]{zou2023representation}
Andy Zou et al.
\newblock Representation engineering: A top-down approach to AI transparency.
\newblock \emph{arXiv:2310.01405}, 2023.

\bibitem[Zou et al.(2025)]{zou2025axbench}
Andy Zou, Long Phan, Justin Wang, Derek Duenas, Maxwell Lin, Maksym Andriushchenko,
  Rowan Wang, Zico Kolter, Matt Fredrikson, and Dan Hendrycks.
\newblock AxBench: Steering LLMs? Even simple baselines outperform activation addition.
\newblock \emph{arXiv:2501.17148}, 2025.

\bibitem[Davidson et al.(2017)]{davidson2017hate}
Thomas Davidson, Dana Warmsley, Michael Macy, and Ingmar Weber.
\newblock Automated hate speech detection and the problem of offensive language.
\newblock In \emph{ICWSM}, 2017.

\bibitem[Gao et al.(2024)]{gao2023scaling}
Leo Gao, Tom Dup\'{e} la Tour, Henk Tillman, Gabriel Goh, Rajan Tesi,
  Shan Carter, Chris Olah, John Schulman, Ilya Sutskever, and Jan Leike.
\newblock Scaling and evaluating sparse autoencoders.
\newblock \emph{arXiv:2406.04093}, 2024.

\bibitem[Intel(2023)]{intel2023politeguard}
Intel Corporation.
\newblock Polite-Guard: A dataset for politeness classification.
\newblock HuggingFace: \texttt{Intel/polite-guard}, 2023.

\bibitem[Koo and Mae(2016)]{koo2016icc}
Terry K. Koo and Mae Yun Mae.
\newblock A guideline of selecting and reporting intraclass correlation
  coefficients for reliability research.
\newblock \emph{Journal of Chiropractic Medicine}, 15(2):155--163, 2016.

\bibitem[Shrout and Fleiss(1979)]{shrout1979icc}
Patrick E. Shrout and Joseph L. Fleiss.
\newblock Intraclass correlations: Uses in assessing rater reliability.
\newblock \emph{Psychological Bulletin}, 86(2):420--428, 1979.

\bibitem[Husain et al.(2019)]{husain2019codesearchnet}
Hamel Husain, Ho-Hsiang Wu, Tiferet Gazit, Miltiadis Allamanis, and Marc Brockschmidt.
\newblock CodeSearchNet challenge: Evaluating the state of semantic code search.
\newblock \emph{arXiv:1909.09436}, 2019.

\bibitem[Kornilova and Eidelman(2019)]{kornilova2019billsum}
Anastassia Kornilova and Vladimir Eidelman.
\newblock BillSum: A corpus for automatic summarization of US legislation.
\newblock In \emph{Proceedings of the 2nd Workshop on New Frontiers in Summarization}, 2019.

\bibitem[Chalkidis et al.(2022)]{chalkidis2021lexglue}
Ilias Chalkidis, Abhik Jana, Dirk Hartung, Michael Bommarito, Ion Androutsopoulos,
  Daniel Martin Katz, and Nikolaos Aletras.
\newblock LexGLUE: A benchmark dataset for legal language understanding in English.
\newblock In \emph{Proceedings of the 60th Annual Meeting of the ACL}, 2022.

\bibitem[Zhang et al.(2015)]{zhang2015text}
Xiang Zhang, Junbo Zhao, and Yann LeCun.
\newblock Character-level convolutional networks for text classification.
\newblock In \emph{Advances in Neural Information Processing Systems}, volume~28, 2015.

\bibitem[Borkan et al.(2019)]{borkan2019nuanced}
Daniel Borkan, Lucas Dixon, Jeffrey Sorensen, Nithum Thain, and Lucy Vasserman.
\newblock Nuanced metrics for measuring unintended bias with real data for text classification.
\newblock In \emph{Companion Proceedings of The 2019 World Wide Web Conference}, 2019.

\bibitem[Maas et al.(2011)]{maas2011learning}
Andrew~L. Maas, Raymond~E. Daly, Peter~T. Pham, Dan Huang, Andrew~Y. Ng, and Christopher Potts.
\newblock Learning word vectors for sentiment analysis.
\newblock In \emph{Proceedings of the 49th Annual Meeting of the ACL}, 2011.

\bibitem[Baumgartner et al.(2020)]{baumgartner2020pushshift}
Jason Baumgartner, Savvas Zannettou, Brian Keegan, Megan Squire, and Jeremy Blackburn.
\newblock The Pushshift Reddit dataset.
\newblock In \emph{Proceedings of the 14th International AAAI Conference on Web and Social Media}, 2020.

\end{thebibliography}
\end{document}